\documentclass[twoside,11pt]{article}

\usepackage{blindtext}

\usepackage[arxiv]{jmlr2e}

\usepackage{times}

\usepackage{url}
\usepackage{amsmath}
\usepackage{wrapfig}
\usepackage{microtype}
\usepackage{booktabs} 
\usepackage[hypcap=false]{caption}
\usepackage{subfigure}
\usepackage{tabularx}
\usepackage{threeparttable}
\usepackage{diagbox}
\usepackage{xcolor}
\usepackage{multirow}
\usepackage{mathrsfs}

\usepackage{enumitem,balance,mathtools}

\usepackage{lastpage}
\ShortHeadings{Implicit vs Unfolded Graph Neural Networks}{Yang, Liu, Wang, Huang, Wipf}

\setlist[itemize]{topsep=3pt,leftmargin=12pt}
\setlist[enumerate]{topsep=3pt,leftmargin=12pt}

\newcommand{\beginproof}[1]{\noindent\textbf{\textit{#1~ }}}
\newcommand{\qed}[0]{\hfill $\blacksquare$ \newline}

\newcommand{\Tinv}[0]{\text{\rotatebox[]{180}{$T$}}}

\newcommand{\subtitle}[1]{\noindent\textit{\textbf{#1}}}

\begin{document}

\title{Implicit vs Unfolded Graph Neural Networks}

\author{\name Yongyi Yang \email yongyi@umich.edu \\
    \addr 
    University of Michigan  \\
    Ann Arbor, United States
\AND
    \name Tang Liu \email cnliutang@gmail.com \\
    \addr 
    Alibaba \\
    Hangzhou, China
\AND
    \name Yangkun Wang \email espylapiza@gmail.com \\
    \addr 
    University of California, San Diego\\
    La Jolla, United States
\AND 
    \name Zengfeng Huang \email huangzf@fudan.edu.cn \\
    \addr 
    Fudan University \\
    Shanghai, China
\AND
    \name David Wipf \email davidwipf@gmail.com \\
    \addr Amazon Web Services \\
    Shanghai, China
}

\editor{}

\maketitle

\vspace{-0.6cm}
\begin{abstract}
It has been observed that message-passing graph neural networks (GNN) sometimes struggle to maintain a healthy balance between the efficient/scalable modeling of long-range dependencies across nodes while avoiding unintended consequences such oversmoothed node representations, sensitivity to spurious edges, or inadequate model interpretability.  To address these and other issues, two separate strategies have recently been proposed, namely \textit{implicit} and \textit{unfolded} GNNs (that we abbreviate to IGNN and UGNN respectively).  The former treats node representations as the fixed points of a deep equilibrium model that can efficiently facilitate arbitrary implicit propagation across the graph with a fixed memory footprint.  In contrast, the latter involves treating graph propagation as unfolded descent iterations as applied to some graph-regularized energy function.  While motivated differently, in this paper we carefully quantify explicit situations where the solutions they produce are equivalent and others where their properties sharply diverge.  This includes the analysis of convergence, representational capacity, and interpretability.  In support of this analysis, we also provide empirical head-to-head comparisons across multiple synthetic and public real-world node classification benchmarks.  These results indicate that while IGNN is substantially more memory-efficient, UGNN models support unique, integrated graph attention mechanisms and propagation rules that can achieve strong node classification accuracy across disparate regimes such as adversarially-perturbed graphs, graphs with heterophily, and graphs involving long-range dependencies.\footnote{A version of this work has been accepted to the Journal of Machine Learning Research.}
\end{abstract}

\section{Introduction}
Given graph data with node features, graph neural networks (GNNs) represent an effective way of exploiting relationships among these features to predict labeled quantities of interest, e.g., node classification \citep{wu2020comprehensive,DBLP:journals/corr/abs-1812-08434}.  In particular, each layer of a message-passing GNN (our focus) is constructed by bundling a graph propagation step with an aggregation function such that information can be shared between neighboring nodes to an extent determined by network depth.  \citep{kipf2017semi,HamiltonYL17,DBLP:journals/jcamd/KearnesMBPR16,VelickovicCCRLB18}.

\begin{figure}
    \centering
    \includegraphics[width=\linewidth]{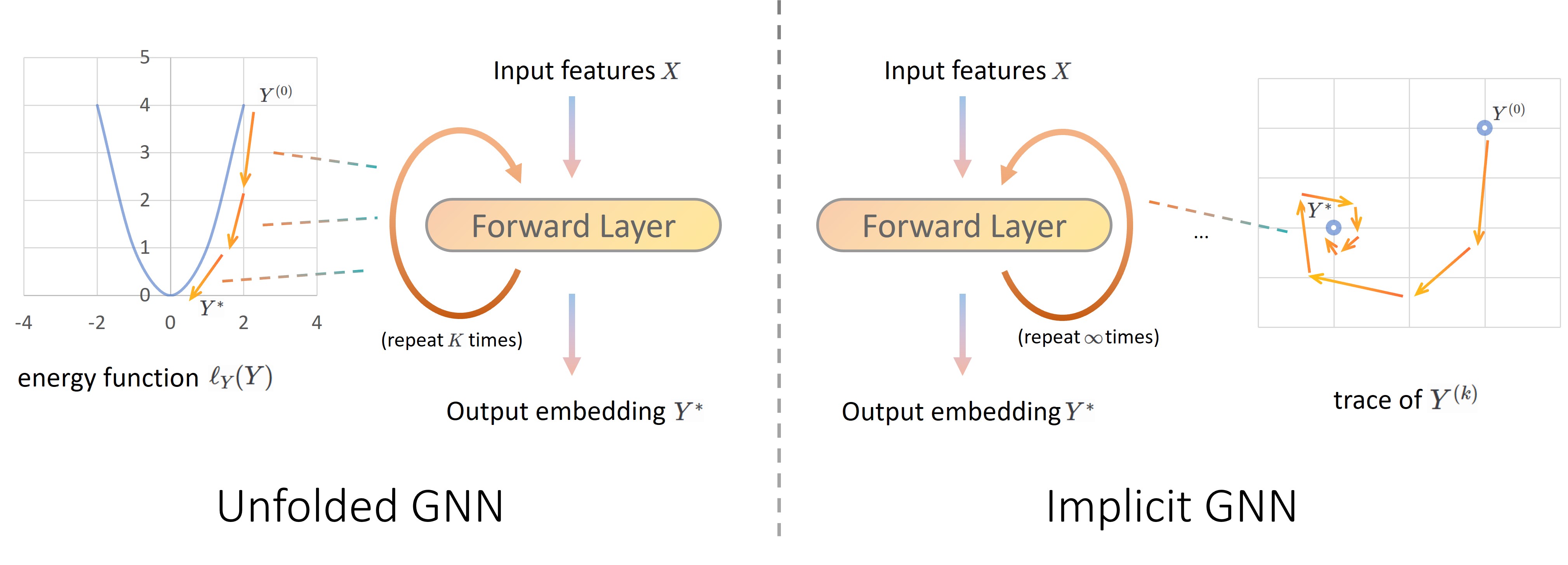}
    \caption{Illustration of UGNNs and IGNNs.  The lefthand side  displays IGNN forward pass layers, each of which reduces the \textit{distance} between the current node embedding $Y^{(k)}$ and a fixed point $Y^*$. contrast, the righthand side reveals that, given input features $X$, each UGNN forward pass layer induces node embeddings $Y$ that form a \textit{descent step} of an energy function $\ell_Y$.}
    \label{fig:intro-figure}
\end{figure}

For sparsely-labeled graphs, or graphs with entity relations reflecting long-range dependencies, it is often desirable to propagate signals arbitrary distances across nodes, but without expensive memory requirements during training, oversmoothing effects \citep{DBLP:conf/iclr/OonoS20,DBLP:conf/aaai/LiHW18}, or disproportionate sensitivity to bad/spurious edges \citep{DBLP:conf/nips/ZhuYZHAK20,DBLP:journals/corr/abs-2009-13566,DBLP:conf/ijcai/ZugnerAG19,DBLP:conf/iclr/ZugnerG19}.  Moreover, to the extent possible we would like such propagation to follow interpretable patterns that can be adjusted in predictable ways to problem-specific contexts (e.g., graphs possessing heterophily).  To address these issues, at least in part, two distinct strategies have recently been proposed that will form the basis of our study.\footnote{There are of course many more than just two approaches for tackling these types of issues, e.g., random-walk GNNs, cooperative GNNs, graph rewiring, graph Transformers \citep{finkelshtein2023cooperative,karhadkar2022fosr,tonshoff2021walking,topping2021understanding,wu2024simplifying}; however, a broader treatment of such methodology remains outside of our primary scope herein.}  First, the framework of implicit deep learning \citep{bai2019deep,el2020implicit} has been applied to producing supervised node embeddings that satisfy an equilibrium criteria instantiated through a graph-dependent fixed-point equation.  The resulting so-called \textit{implicit} GNN (IGNN) pipeline \citep{GuC0SG20}, as well as related antecedents \citep{DaiKDSS18,gallicchio2020fast} and descendants \citep{MIGNN,fu2023implicit,eignn}, mimic the behavior of a GNN model with arbitrary depth for handling long-range dependencies, but is nonetheless trainable via a fixed memory budget and robust against oversmoothing effects. See Figure \ref{fig:intro-figure} for a high-level conceptual depiction of IGNNs.

Secondly, \textit{unfolded} GNN (UGNN) architectures are formed via graph propagation layers patterned after the unfolded descent iterations of some graph-regularized energy function \citep{chen2021graph,di2023understanding,liu2021elastic,ma2020unified,pan2021a,lazygnn,yang2021,thatpaper,zhu2021interpreting}; Figure \ref{fig:intro-figure} contains an illustration. In this way, the node embeddings at each UGNN layer can be viewed as increasingly refined approximations to an interpretable energy minimizer, which can be nested within a bilevel optimization framework \citep{wang2016learning} for supervised training akin to IGNN.  More broadly, similar unfolding strategies have previously been adopted for designing a variety of new deep architectures or interpreting existing ones \citep{hamburger,gregor2010learning,Hao2017sparse,hershey2014deep,Sprechmann15,velivckovic2019neural,yang2022transformers}. 






As such, IGNN and UGNN are closely related in the sense that their node embeddings are intentionally aligned with a meta-criterion: either a fixed-point for IGNN or an energy minimizer for UGNN, where in both cases a useful inductive bias is introduced to address similar desiderata.  And yet despite these commonalities, there has thus far been no systematic examination of the meaningful similarities and differences.  We take key steps in this direction via the following workflow.  First, after introducing the basic setup and notation in Section \ref{sec:basic_setup}, we overview the IGNN framework in Section \ref{sec:ignn_intro}, including its attractive reliance on memory-efficient implicit differentiation and existing convergence results.  Next, Section \ref{sec:ugnn_intro} introduces a general-purpose UGNN framework that encompasses a variety of existing unfolded models as special cases, including models with graph attention, broad graph propagation operators, and nonlinear activation functions.  Sections \ref{sec:convergence}--\ref{sec:shared_weights} provide comparative analysis of the relative strengths and weaknesses of IGNN and UGNN models with respect to convergence guarantees and model expressiveness. We conclude with empirical comparisons in Section \ref{sec:experiments} and scalability considerations in Section \ref{sec:time-and-memory}, while proofs and supplementary details are deferred to the Appendix.  Overall, our contributions can be summarized as follows:
\begin{itemize}
\item Although the original conceptions are unique, we consider a sufficiently broad design space of IGNN and UGNN models such that practically-relevant, interpretable regions of exact overlap can be established, analyzed, and contrasted with key areas of nonconformity.  
    
\item Within this framework, we compare the relative ability of IGNNs and UGNNs to converge to optimal (or locally-optimal) solutions per various model-specific design criteria. Among other things, this analysis reveals that IGNNs enjoy an advantage in that (unlike UGNNs) their implicit layers do not unavoidably induce symmetric propagation weights.  In contrast, we show that UGNNs are more flexible by accommodating a broader class of robust nonlinear graph propagation operators while still guaranteeing at least local convergence.  This capability is relevant to handling graphs with certain types of heterophily (meaning nodes sharing an edge may not share similar labels and/or features) and adversarial perturbations.

\item We investigate the consequences of symmetric (layer-tied) UGNN propagation weights (an issue that has been raised but remains unresolved in the literature).  In particular, we prove that with linear activations, a UGNN can reproduce any IGNN representation, and define sufficient conditions for equivalency in broader regimes.  We also show that UGNN layers with symmetric, layer-tied weights can exactly mimic arbitrary graph convolutional network (GCN) models \citep{kipf2017semi} characterized by asymmetric, layer-specific weights without introducing any additional parameters or significant complexity.  Collectively, these results suggest that the weight symmetry enforced by UGNN models may not be overly restrictive in practice.


\item Empirically, we provide comprehensive, head-to-head comparisons between equivalently-sized IGNN and UGNN models while evaluating computational complexity and memory costs. These results also serve to complement our analytical findings while showcasing node classification regimes whereby IGNN or UGNN can achieve near-SOTA performance.  We also demonstrate how graph heterophily can manifest in fundamentally different ways that UGNN models in particular are naturally equipped to handle.

\end{itemize}




\section{Basic Setup} \label{sec:basic_setup}
Consider a graph ${\mathcal G} = \{{\mathcal V},{\mathcal E}\}$, with $n = |{\mathcal V}|$ nodes and $m=|{\mathcal E}|$ edges.  We define $L \in \mathbb{R}^{n\times n}$ as the Laplacian of ${\mathcal G}$, meaning $L = D-A = B^\top B$, where $D$ and $A$ are degree and adjacency matrices respectively, while $B \in \mathbb{R}^{m\times n}$ is an incidence matrix.  We also let $\widetilde{A} = A + I$ (i.e., $A$ with self loops) and denote $\widetilde{D}$ as the corresponding degree matrix.

Both IGNNs and UGNNs incorporate graph structure via optimized embeddings $Y^* \in \mathbb{R}^{n\times d}$ that are a function of some adjustable weights ${\mathcal W}$, i.e., $Y^* \equiv Y^*({\mathcal W})$ (for simplicity of notation, we will frequently omit including this dependency on ${\mathcal W}$), where by design $\partial Y^*({\mathcal W})/ \partial {\mathcal W}$ is computable, either implicitly (IGNN) or explicitly (UGNN). We may then insert these embeddings within an application-specific meta-loss given by
\vspace*{-0.0cm}
\begin{equation} \label{eq:meta_loss}
\ell_{\theta}\left(\theta,{\mathcal W} \right) \triangleq \sum_{i=1}^{n'} {\mathcal D}\big( g\left[  {\boldsymbol{y}}_i^*({\mathcal W}); \theta \right], {\boldsymbol{t}}_i  \big),
\end{equation}
\vspace*{-0.0cm}
where $g : \mathbb{R}^d \rightarrow \mathbb{R}^c$ is some differentiable node-wise function with parameters $\theta$ and $c$-dimensional output tasked with predicting ground-truth node-wise targets ${\boldsymbol{t}}_i \in \mathbb{R}^c$.  Additionally, ${\boldsymbol{y}}_i^*({\mathcal W})$ is the $i$-th row of $Y^*({\mathcal W})$,  $n' < n$ is the number of labeled nodes (we assume w.l.o.g.~that the first $n'$ nodes are labeled), 
and ${\mathcal D}$ is a discriminator function, e.g., cross-entropy for classification, squared error for regression, etc. Given that $Y^*({\mathcal W})$ is differentiable by construction, we can optimize $\ell_{\theta}\left(\theta,{\mathcal W} \right)$ via gradient descent to obtain our final predictive model.

At a conceptual level then, the only difference between IGNNs and UGNNs is in how the corresponding optimal embeddings $Y^*({\mathcal W})$ are motivated and computed.  We introduce the specifics of each option in the following two sections.

\section{Overview of Implicit GNNs} \label{sec:ignn_intro}
IGNN models are predicated on the fixed-point update
\begin{equation} \label{eq:IGNN_iteration}
    Y^{(k+1)} = \sigma \left[P Y^{(k)} W_p + f\left(X ; W_x  \right)  \right],
\end{equation}
where $Y^{(k)}$ are node embeddings after the $k$-th iteration, $\sigma$ is a nonlinear activation function, ${\mathcal W} = \{W_p,W_x \}$ is a set of two weight matrices, and $P \in \mathbb{R}^{n\times n}$ is an arbitrary graph propagation operator, e.g., $P = A$; other variants are also obtainable using, for example, graph rewiring \citep{gasteiger2019diffusion}.  Additionally, $f : \mathbb{R}^{n\times d_0}\rightarrow \mathbb{R}^{n\times d}$ is a base predictor function that converts the initial $d_0$-dimensional node features $X \in \mathbb{R}^{n\times d_0}$ into $d$-dimensional candidate embeddings, e.g., $f\left(X;W_x\right) = X W_x$ or $f\left(X;W_x\right) = \text{MLP}\left[X; W_x \right]$.  


Now assume that $\sigma$ is a differentiable component-wise non-expansive (CONE) mapping as specified in \citep{GuC0SG20}.  This condition stipulates that $\sigma$ applies the same function individually to each element of its input, and that this function satisfies $\|x-y\| \geq \|\sigma(x)-\sigma(y)\|$ for all $\{x,y\} \in \mathbb{R}^2$ (with some abuse of notation, we will overload the definition of $\sigma$ when the meaning is clear from context).  Furthermore, we assume that the weights $W_p$ are such that $\lambda_{\text{pf}}(|P \otimes W_p|) < 1$, where $|\cdot|$ denotes the element-wise absolute value and $\lambda_{\text{pf}}$ refers to the Peron-Frobenius eigenvalue.\footnote{The Peron-Frobenius (PF) eigenvalue is a real, non-negative eigenvalue that exists for all square, non-negative matrices and produces the largest modulus.}  

Under these conditions, it has been shown in \citep{GuC0SG20} that $\lim_{k\rightarrow \infty} Y^{(k)} = Y^*$, where $Y^*$ satisfies the fixed-point equation $Y^* = \sigma \left[P Y^* W_p + f\left(X ; W_x  \right)  \right]$.  Therefore, as an IGNN forward pass we can iterate (\ref{eq:IGNN_iteration}) $K$ times, where $K$ is sufficiently large, to compute node embeddings $Y^{(K)} \approx Y^*$ for use within the meta-loss from (\ref{eq:meta_loss}).  For the IGNN backward pass, implicit differentiation methods \citep{bai2019deep,el2020implicit}, carefully tailored for handling graph-based models \citep{GuC0SG20}, can but used to compute gradients of $Y^*$ with respect to $W_p$ and $W_x$.  Critically, this implicit technique does \textit{not} require storing each intermediate representation $\{ Y^{(k)} \}_{k=1}^K$ from the forward pass, and hence is quite memory efficient even if $K$ is arbitrarily large.  As $K$ can be viewed as quantifying the extent of signal propagation across the graph, IGNNs can naturally capture long-range dependencies with a $K$-\textit{independent} memory budget unlike more traditional techniques.

In follow-up work from \citep{eignn} a special case of (\ref{eq:IGNN_iteration}) is considered whereby $\sigma$ is an identity mapping and $W_p$ is constrained to be a symmetric matrix $W_p^s$.  These simplifications facilitate a more stream-lined implementation,
referred to as \textit{efficient infinite-depth graph neural networks} (EIGNN) in \citep{eignn}, while maintaining accurate predictions on various node classification tasks.  As we will demonstrate in the next section, the EIGNN fixed point minimizes an explicit energy function allowing this approach to be simultaneously interpreted as a special case of UGNN as well.  Additionally, while EIGNN is designed to to economize IGNNs by removing the nonlinear activation, more recent work has targeted increasing IGNN expressivity \citep{MIGNN} by relaxing the requirement that IGNN weights satisfy the Peron-Frobenius eigenvalue constraint from above in certain settings using monotone operator theory; hence the name MIGNN.  And in a related vein, it has also been proposed to instantiate IGNNs (and the propagation $P$ in particular) using a parameterized graph Laplacian operator that may be more suitable for handling potential oversmoothing effects \citep{fu2023implicit}.


\section{A General-Purpose Unfolded GNN Framework} \label{sec:ugnn_intro}

In this section we will first introduce a general-purpose energy function that, when reduced with simplifying assumptions, decomposes into various interpretable special cases that have been proposed in the literature, both as the basis for new UGNN models as well as motivation for canonical GNN architectures.  Later, we will describe the generic proximal gradient descent iterations designed to optimize this energy. By construction, these iterations unfold in one-to-one correspondence with the UGNN model layers of each respective architecture under consideration.

\subsection{Flexible Energy Function Design} \label{sec:flexible_energy}

Unlike IGNN models, the starting point of UGNNs is an energy function.  In order to accommodate a broad variety of existing graph-regularized objectives and ultimately GNN architectures as special cases, we propose the rather general form
\begin{equation} \label{eq:general_unfolded_objective}
\ell_Y(Y; {\mathcal W}, f, \rho, \widetilde{B},\phi) \triangleq \left\|Y - f\left(X ; W_x  \right) \right\|_{W_f}^2 + \sum_{i=1}^m \rho\left( \left[  \widetilde{B} Y W_p Y^\top \widetilde{B}^\top \right]_{ii} \right) + \sum_{i=1}^n \phi\left({\boldsymbol{y}}_{i} ; W_\phi \right),
\end{equation}
where ${\mathcal W} = \{W_x,W_f,W_p,W_\phi \}$ is a set of four weight matrices,  $\rho: \mathbb{R} \rightarrow \mathbb{R}$ is a differentiable function, and $\widetilde{B} \in \mathbb{R}^{m\times n}$ is a function of $A$ that can be viewed as a generalized incidence matrix.  Furthermore, $\phi : \mathbb{R}^d \rightarrow \mathbb{R}$ is an arbitrary function (possibly non-smooth) of the node embeddings that is bounded from below, and the weighted norm $\|\cdot \|_W$ is defined such that $\|X \|^2_W = \text{tr}[X^\top W X]$.  The above loss is composed of three terms: (i) A quadratic penalty on deviations of the embeddings $Y$ from the node-wise base model $f\left(X ; W_x  \right)$, (ii) a (possibly) non-convex graph-aware penalty that favors smoothness across edges (see illustrative special cases below), and (iii) an arbitrary function of each embedding that is independent of both the graph structure $A$ and the original node features $X$.  

As a representative special case, if $\widetilde{B} = B$, then the second term from (\ref{eq:general_unfolded_objective}) satisfies
\begin{equation} \label{eq:penalty_simplification}
\sum_{i=1}^m \rho\left( \left[  \widetilde{B} Y W_p Y^\top \widetilde{B}^\top \right]_{ii} \right) = \sum_{\{i,j\} \in {\mathcal E}} \rho\left( \left\| {\boldsymbol{y}}_{i} - {\boldsymbol{y}}_{j} \right\|_{W_p}^2 \right),
\end{equation}
which penalizes deviations between the embeddings of two nodes sharing an edge.  As has been noted in our prior work \citep{yang2021}, if $\rho$ is chosen to be concave and non-decreasing on $[0,\infty)$, the resulting regularization favors robustness to bad edges, and could be useful for handling heterophily graphs or adversarial attacks.  Intuitively, this robustness occurs because the penalized differences between the embeddings of two nodes sharing a spurious edge no longer accumulate quadratically, a growth-rate characterized by a well-known sensitivity to outliers \citep{west1984outlier}.\footnote{In the present context, this would imply that spurious edges (e.g.,  that were errantly included in the graph) or which link nodes with neither label nor features in common, could dominate the resulting loss leading to poor performance on downstream tasks.  Meanwhile, concave regularization grows slowly once the embeddings are already sufficiently separated, muting the impact of outliers.}

Simplifying further, if in addition to the assumption from (\ref{eq:penalty_simplification}), $\phi$ is set to zero, $\rho$ is a (non-robust) identity mapping, $ W_f = I$, and $W_p = \lambda I$ with $\lambda > 0$, then (\ref{eq:general_unfolded_objective}) collapses to the more familiar loss
\begin{equation} \label{eq:basic_twirls_objective}
\ell_Y(Y;W_x,f) \triangleq  \left\|Y - f\left(X ; W_x  \right) \right\|_{{\mathcal F}}^2 + \lambda \text{tr}\left[Y^\top L Y  \right],
\end{equation}
as originally proposed in \citep{zhou2004learning} to smooth base predictions from $f\left(X ; W_x  \right)$ using a quadratic, graph-aware regularization factor $\text{tr}\left[Y^\top L Y  \right] = \sum_{\{i,j\} \in {\mathcal E}} \left\| {\boldsymbol{y}}_{i} - {\boldsymbol{y}}_{j} \right\|^2$.  This construction has also formed the basis of a wide variety of work linking different GNN architectures \citep{ma2020unified,pan2021a,lazygnn,thatpaper,zhu2021interpreting}; more details to follow in Section \ref{sec:ugnn_descent}.

Similarly, if we allow for broader choices of $\widetilde{B} \neq B$ with $\widetilde{L} \triangleq \widetilde{B}^\top \widetilde{B} = \pi(A)$ for some function $\pi : \mathbb{S}^n \rightarrow \mathbb{S}^n$ over the space of $n$-dimensional PSD matrices $\mathbb{S}^n$, then  (\ref{eq:basic_twirls_objective}) can be generalized by swapping $L$ for this $\widetilde{L}$ as proposed in \citep{ioannidis2018,fu2023implicit}.  Candidates for $\widetilde{L}$ include normalized graph Laplacians \citep{von2007tutorial} and various diffusion kernels designed according to application-specific requirements, e.g., graph signal denoising or graph rewiring \citep{gasteiger2019diffusion,klicpera2019diffusion}.  Moreover, if we reintroduce $W_p \neq I$, then it can be shown that the EIGNN fixed point (assuming suitable hyperparameter settings) minimizes the resulting generalized version of (\ref{eq:basic_twirls_objective}); see Appendix for specifics of the derivation as the original EIGNN algorithm \citep{eignn} is not motivated in this way.




\subsection{Descent Iterations That Form UGNN Layers} \label{sec:ugnn_descent}

We now derive descent iterations for the general loss from the previous section (and later more interpretable special cases) that will be mapped to UGNN layers.  For this purpose, let $U^{(k)}$ denote a single gradient descent step along (\ref{eq:general_unfolded_objective}), excluding the possibly non-differentiable $\phi$-dependent term, and evaluated at some candidate point $Y^{(k)}$.  We may compute such an abridged gradient step as\footnote{Note that here w.l.o.g.~we have absorbed an extra $W_f + W_f^\top$ factor into $f(X;W)$.} 
\begin{equation} \label{eq:basic_grad_step}
U^{(k)} = Y^{(k)} - \alpha \left[ \widetilde{B}^\top \Gamma^{(k)} \widetilde{B} Y^{(k)} \left(W_p + W_p^\top \right)  + Y^{(k)}\left(W_f + W_f^\top \right) - f\left(X ; W  \right) \right],
\end{equation}
where $\alpha$ is the step size and $\Gamma^{(k)}$ is a diagonal matrix with $i$-th diagonal element given by
\begin{equation} \label{eq:gamma_update}
\gamma_e^{(k)} ~=~  \left. \frac{\partial \rho\left( z \right) }{\partial z}\right|_{z = \text{diag}\left[  \widetilde{B} Y^{(k)} W_p \left(Y^{(k)}\right)^\top \widetilde{B}^\top \right]_e},
\end{equation}
and $e \in \{1,\ldots,m\}$ is an index for the $e$-th edge in ${\mathcal E}$. We then have the following:
\begin{lemma} \label{lem:descent}
If $\rho$ has Lipschitz continuous gradients, then the proximal gradient update
\begin{equation} \label{eq:prox_step}
    Y^{(k+1)} = \text{\normalfont prox}_\phi\left( U^{(k)}\right)  \triangleq \arg\min_{Y} \left[ \frac{1}{2\alpha}\| U^{(k)} - Y \|_{{\mathcal F}}^2 + \sum_i \phi\left({\boldsymbol{y}}_i ; W_\phi \right) \right],
\end{equation}
is guaranteed to satisfy~ $\ell_Y(Y^{(k+1)}; {\mathcal W}, f, \rho, \widetilde{B},\phi) \leq \ell_Y(Y^{(k)}; {\mathcal W}, f, \rho, \widetilde{B},\phi)$~ for any $\alpha \in \left(0,1/{\mathcal L} \right]$, where ${\mathcal L}$ is the Lipschitz constant for gradients of (\ref{eq:general_unfolded_objective}) w.r.t.~$Y$, excluding the non-smooth $\phi$ term.
\end{lemma}  
The function $\text{prox}_\phi : \mathbb{R}^d \rightarrow \mathbb{R}^d$ is known as the \textit{proximal operator}\footnote{If (\ref{eq:prox_step}) happens to have multiple solutions, then the proximal operator selects one of them so as to remain a proper deterministic function of its argument, which is relevant for forming later connections with IGNN.} associated with $\phi$ (separably extended across all $n$ nodes in the graph), and the proof follows from basic properties of gradient descent \citep{Bertsekas1999} and proximal methods \citep{combettes2011proximal}; see Appendix.



Generally speaking, the mapping $Y^{(k)} \mapsto U^{(k)}$ from (\ref{eq:basic_grad_step}) can be viewed as a (possibly nonlinear) graph filter, while $\text{prox}_\phi$ from (\ref{eq:prox_step}) serves as an activation function applied to the embedding of each node.  Collectively then, $Y^{(k+1)} = \text{prox}_\phi\left( U^{(k)}\right)$ provides a flexible template for UGNN layers that naturally reduces to common GNN architectures per various design choices.  And analogous to IGNN, we can execute $K$ UGNN steps to approximate some $Y^*$, which could be a fixed-point, stationary point, or global optimum; more on this in Sections \ref{sec:ugnn_fixed_point} and \ref{sec:convergence} below.

For example, consider the selection $\sum_i \phi\left({\boldsymbol{y}}_{i} ; W_\phi \right) = \sum_{i,j} {\mathcal I}_{\infty}[y_{ij} < 0]$, where $y_{ij}$ is the $(i,j)$-th element of $Y$ and ${\mathcal I}_{\infty}$ is an indicator function that assigns an infinite penalty to any $y_{ij} < 0$.  The proximal operator then becomes $\text{prox}_\phi (U) = \text{ReLU}(U)$, i.e., a ReLU function that element-wise sets negative values to zero.  If we add this  $\phi$-dependent term to (\ref{eq:basic_twirls_objective}) (as a special case of  (\ref{eq:general_unfolded_objective})), the resulting update becomes
\begin{equation} \label{eq:example_prop1}
Y^{(k+1)} = \text{prox}_\phi\left( U^{(k)}\right) = \text{ReLU}\left( Y^{(k)} - \alpha\left[ \left( \lambda L  + I \right) Y^{(k)} - f\left(X ; W  \right) \right] \right).
\end{equation}

And when combined with observations from \citep{ma2020unified,pan2021a,thatpaper,zhu2021interpreting},\footnote{While these works do not consider the treatment of nonlinear activations, they do nicely introduce how the embedded linear gradient step $U^{(k)}$ relates to GCN filters.} if we initialize with $Y^{(0)} = f\left(X ; W  \right) = X W$ and apply simple reparameterizations, the first iteration reduces to the canonical GCN layer $Y^{(1)} = \text{ReLU}\left[\widetilde{D}^{-1/2} \widetilde{A} \widetilde{D}^{-1/2} X W \right]$ from \citep{kipf2017semi}.  Subsequent iterations are no longer equivalent, although the inherent differences (e.g., the skip connections from the input layer) are useful for avoiding oversmoothing effects that can at times hamper deep GCN models \citep{DBLP:conf/iclr/OonoS20,DBLP:conf/aaai/LiHW18,DBLP:conf/iclr/RongHXH20}.



Additionally, the widely-used APPNP architecture \citep{klicpera2019predict} is also a special case of (\ref{eq:example_prop1}) when the ReLU operator is removed (i.e., $\phi$ is zero), $\alpha=\frac{1}{1+\lambda}$, and $L$ is changed to the symmetric normalized Laplacian \citep{von2007tutorial}.  And as a final representative example, if we adopt a nonlinear choice for $\rho$, then $L$ in (\ref{eq:example_prop1}) will be replaced by $L^{(k)} \triangleq B^\top \Gamma^{(k)} B$, where $\Gamma^{(k)}$ rescales graph edges.  As has been discussed in \citep{yang2021}, this instantiates a form of graph attention, whereby the attention weights $\Gamma^{(k)}$ produced by a concave, non-decreasing $\rho$ add robustness to spurious  edges. To see why this happens, note that if $\rho$ is concave and non-decreasing, then each $\gamma_e^{(k)}$ computed via (\ref{eq:gamma_update}) will necessarily be a \textit{non-increasing} function of $z$.  Furthermore, taking $W_p = I$  and $\widetilde{B} = B$ for simplicity of exposition, then $z = \text{diag}\left[  \widetilde{B} Y^{(k)} W_p \left(Y^{(k)}\right)^\top \widetilde{B}^\top \right]_e = \|{\boldsymbol{y}}_i - {\boldsymbol{y}}_j\|^2_2$ for some $(i,j) \in {\mathcal E}$.  Hence if the difference between ${\boldsymbol{y}}_i$ and ${\boldsymbol{y}}_j$ becomes \textit{large}, as may be expected across a spurious edge, the corresponding attention weight $\gamma_e^{(k)}$ will necessarily become \textit{small} as desired to avoid dominating the loss.



\subsection{UGNN Fixed Points and Connections with IGNN} \label{sec:ugnn_fixed_point}
While UGNN layers are formed from the descent steps of a graph-regularized energy, for a more direct comparison with IGNN, it is illuminating to consider the situation where the update $Y^{(k+1)} = \text{prox}_\phi\left( U^{(k)}\right)$ is iterated with $k$ becoming sufficiently large.   In this case, we may ideally reach a fixed point $Y^*$ that satisfies
\begin{eqnarray} \label{eq:ugnn_fixed_point}
Y^* & = & \text{prox}_\phi \left( Y^* - \alpha \left[ \widetilde{B}^\top \Gamma^* \widetilde{B} Y^* \left(W_p + W_p^\top \right)  + Y^*\left(W_f + W_f^\top \right) - f\left(X ; W  \right) \right] \right) \nonumber \\
& = & \text{prox}_\phi \left( Y^*\left[I - \alpha  W_f^s -\alpha W_p^s \right]  
+ \alpha \left [I-\widetilde{B}^\top \Gamma^* \widetilde{B}\right] Y^* W_p^s + \alpha f\left(X ; W  \right)  \right),
\end{eqnarray}
where $W_p^s \triangleq W_p + W_p^\top$ and $W_f^s \triangleq W_f + W_f^\top$ are symmetric weight matrices, and $\Gamma^*$ is defined analogously to (\ref{eq:gamma_update}).  It now follows that if $W_f^s = I- W_p^s, \alpha = 1$ and $\Gamma^* = I$ (i.e., $\rho$ is an identity mapping), and we define the propagation operator as $P = I- \widetilde{B}^\top \widetilde{B} \equiv I- \widetilde{L}$, then (\ref{eq:ugnn_fixed_point}) reduces to
\begin{equation}\label{eq:ugnn_fixed_point_simple}
Y^* =  \text{prox}_\phi \left[  
 P Y^* W_p^s + f\left(X ; W  \right)  \right].
\end{equation}
\textit{This expression is exactly equivalent to the IGNN fixed point from Section \ref{sec:ignn_intro} when we set $\sigma$ to the proximal operator of $\phi$ and we restrict $W_p$ to be symmetric.}  For direct head-to-head comparisons then, it is instructive to consider what CONE activation functions $\sigma$ can actually be expressed as the proximal operator of some penalty $\phi$.  In this regard we have the following:
\begin{lemma} \label{lem:proximal}
A continuous CONE function $\sigma: \mathbb R \to \mathbb R$ can be expressed as the proximal operator of some function $\phi$ iff $\sigma$ is also non-decreasing.
\end{lemma}
The proof follows from the analysis in \citep{gribonval2020characterization}.  This result implies that, at least with respect to allowances for decreasing activation functions $\sigma$, IGNN is more flexible than UGNN.



However, the relative flexibility of IGNN vs UGNN fixed points becomes much more nuanced once we take into account convergence considerations.  For example, the proximal gradient iterations from Section \ref{sec:ugnn_descent} do \textit{not} require that $\text{prox}_\phi$ is non-expansive or CONE to guarantee descent, although it thus far remains ambiguous how this relates to obtainable fixed points of UGNN.  To this end, in Section \ref{sec:convergence} we will evaluate when such fixed points exist, what properties they have, convergence conditions for reaching them, and importantly for present purposes, how UGNN fixed points relate to their IGNN counterparts.

\subsection{Potential for Broader Fixed-Point Equivalency Regimes}

Before we tackle formal convergence considerations, a natural question arises regarding the extent to which IGNN and UGNN fixed-points are capable of alignment when we grant greater flexibility to the former.  In particular, if for the moment we relax the restriction that IGNN iterations must strictly adhere to (\ref{eq:IGNN_iteration}), we could potentially treat \textit{any} UGNN fixed-point as the fixed-point of a richer class of IGNN models merely by definition.  We would therefore obtain a trivial equivalency, and in this sense, IGNNs could be interpreted as uniformly more expressive than UGNNs. 

The unresolved issue though, at least for the purposes of the IGNN/UGNN definitions we have adopted for this paper, is that to actually qualify as an IGNN there must exist a stable/scalable procedure for \textit{implicitly} computing the gradients needed for back-propagation and model training.  In other words, we require implicit differentiation conditioned on a given fixed point in order to achieve the signature $K$-independent memory budget of IGNNs as reviewed in Section \ref{sec:ignn_intro}.  Without this, we are just swapping names and there is diminished reason to differentiate IGNNs and UGNNs in the first place.  However, to our knowledge established implicit differentiation schemes applicable to producing GNN-like architectures equipped with graph propagation are all limited to handling fixed-points induced by (\ref{eq:IGNN_iteration}), with different variants tied to varying choices for $\sigma$, $P$, $W_p$ and $f$.  Generalizing further is non-trivial, necessitating architecture-specific accommodations; see Section D of \citep{GuC0SG20} for details of the nuance involved and \citep{MIGNN} for follow-up effort demonstrating the difficulties of extending further.  In fact, even without complex graph structure linking all training instances, achieving stable implicit differentiation is known to be challenging.  Hence we adopt  (\ref{eq:IGNN_iteration}) for characterizing the design space of IGNN models, and leave broader alternatives, should they arise, to future work.



\section{Convergence Comparisons} \label{sec:convergence}
In this section we first detail situations whereby convergence to a unique fixed point, that also may at times correspond with a unique UGNN global minimum, can be established.  In these situations IGNN facilitates stronger guarantees in terms of broader choices for $W_p$ and $\sigma$.  Later, we examine alternative scenarios whereby UGNN has a distinct advantage with respect to convergence guarantees to potentially local solutions.  We remark that prior work examining UGNN iterations has largely been limited to certain special cases of (\ref{eq:general_unfolded_objective}), and critically, only established a non-increasing cost function (see for example \citep{yang2021}), \textit{not} actual convergence to a meaningful solution set; the latter requires confirmation of additional non-trivial criteria as we will tackle herein.  As such, an ancillary benefit of this section is to ground UGNN models to more comprehensive, elucidating convergence properties.

\subsection{Convergence to Unique Global Solutions}\label{sec:global-convergence}
To facilitate the most direct, head-to-head comparison between IGNN and UGNN, we consider a restricted form of the UGNN energy from (\ref{eq:general_unfolded_objective}).  In particular, let 
\begin{equation} \label{eq:unfolded_objective_special_case2}
\ell_Y(Y; {\mathcal W}, f, \phi) \triangleq \left\|Y - f\left(X ; W_x  \right) \right\|_{W_f}^2 + \text{tr}\left[Y^\top \widetilde{L} Y W_p \right] + \sum_{i=1}^n \phi\left({\boldsymbol{y}}_{i} \right),
\end{equation}
which is obtainable from (\ref{eq:general_unfolded_objective}) when $\rho$ is an identity mapping and $\phi$ is assumed to have no trainable parameters.  We also define $\Sigma \triangleq I\otimes W_f^s + \widetilde{L}\otimes W_p^s$, with minimum and maximum eigenvalues given by $\lambda_{\text{min}}(\Sigma)$ and $\lambda_{\text{max}}(\Sigma)$ respectively.


\begin{theorem} \label{thm:global_convergence}
 If $\text{\normalfont prox}_\phi$ is non-expansive and $\lambda_{\text{min}}(\Sigma) > 0$, then (\ref{eq:unfolded_objective_special_case2}) has a unique global minimum.  Additionally, starting from any initialization $Y^{(0)}$,  the iterations $Y^{(k+1)} = \text{prox}_\phi\left[U^{(k)}\right]$ applied to (\ref{eq:unfolded_objective_special_case2}) with $\alpha \in (0,2/\lambda_{\text{max}}(\Sigma) )$ are guaranteed to converge to this global minimum as $k \rightarrow \infty$.
\end{theorem}

\begin{corollary} \label{cor:global_convergence1}
For any  non-decreasing CONE function $\sigma$ and symmetric $W_p \equiv W_p^s$ satisfying $\|W_p^s \|_2 < \|P \|_2^{-1}$, there will exist a function $\phi$ such that $\sigma = \text{\normalfont prox}_\phi$ and the fixed point of (\ref{eq:ugnn_fixed_point_simple}) is the unique global minimum of (\ref{eq:unfolded_objective_special_case2}) when $W_f$ is chosen such that $W_f^s = I - W_p^s$ and $P = I - \widetilde{L}$. 
\end{corollary}
\begin{lemma}

 \label{cor:global_convergence2}
For any non-expansive $\sigma$ (not necessarily element-wise) and $W_p$ (not necessarily symmetric) satisfying $\|W_p \|_2 < \|P \|_2^{-1}$, iterating (\ref{eq:IGNN_iteration}) will converge to a unique fixed point.\footnote{An analogous result has been shown in \citep{GuC0SG20}, but restricted to CONE mappings (not arbitrary non-expansive mappings) and with a dependency on the less familiar Peron-Frobenius eigenvalue; see Section \ref{sec:ignn_intro}.}
\end{lemma}
Hence from Theorem \ref{thm:global_convergence} and Corollary \ref{cor:global_convergence1}, if we are willing to accept symmetric graph propagation weights $W_p^s$ (and a non-decreasing $\sigma$), UGNN enjoys the same convergence guarantee as IGNN, but with the added benefit of an interpretable underlying energy function.  In contrast, when $W_p$ is \textit{not} symmetric as in Lemma \ref{cor:global_convergence2}, we can only guarantee a unique IGNN fixed point, but we are no longer able to establish an association with a specific UGNN energy.  In fact, it does not seem likely that any familiar/interpretable functional form can even possibly exist to underlie such a fixed point when $W_p$ is not symmetric.  This is largely because of the following simple result:
\begin{lemma}\label{lem:no-asym-integrate}
There does not exist any second-order smooth function $h : \mathbb{R}^{n\times d} \rightarrow \mathbb{R}$ such that $\partial h(Y; W)/ \partial Y = Y W $ for all (asymmetric)  matrices $W \in \mathbb{R}^{d \times d}$ with $d > 1$ and $n\geq 1$.
\end{lemma}
And by continuity arguments, a similar result will apply to many non-smooth functions.  Consequently, with asymmetric weights there is no obvious way to associate fixed points with stationary points as we have done for UGNN.  Interestingly, while a wide variety of work on unfolding models (including outside of the graph domain) has led to architectures with symmetric or even more restrictive PSD weight constraints \citep{amos2017optnet,di2023understanding,frecon2022bregman,xie2023optimization,yang2022transformers}, the apparent inevitability of these constraints remains underexplored.

\subsection{Broader Convergence Regimes}
As we move to alternative energy functions with nonlinear dependencies on the graph, e.g., $\rho$ not equal to identity, meaningful (albeit possibly weaker) convergence properties for UGNN can still be established.  In particular, we consider convergence to local minima, or more generally, stationary points of the underlying objective.  However, because (\ref{eq:general_unfolded_objective}) may be non-convex and non-smooth, we must precisely define an applicable notion of stationarity.  

While various possibilities exist, for present purposes we define $Y^*$ as a stationary point of (\ref{eq:general_unfolded_objective}) if ${{\boldsymbol{0}}} \in \partial_{ {\mathcal F} } \left[ \ell_Y(Y^*; {\mathcal W}, f, \rho, \widetilde{B},\phi) \right]$, where $\partial_{ {\mathcal F} }$ denotes the Fr\'{e}chet subdifferential.  The latter generalizes the standard subdifferential as defined for convex functions, to non-convex cases.  More formally, the Fr\'{e}chet subdifferential \citep{li2020understanding} of some function  $h(Y)$ is defined as the set
\begin{equation}
\partial_{ {\mathcal F} } \left[h(Y)\right] = \left\{S :  h(Y) \geq h(Z) + \text{tr}\left[ S^\top\left(Y - Z \right) \right] + o\left( \left\| Y- Z \right\|_{{\mathcal F}} \right) ~~\forall Y\right\},
\end{equation}
which is equivalent to the gradient when $h$ is differentiable and the regular subdifferential when $h$ is convex.  Loosely speaking, this definition specifies a set of affine functions that approximately lower-bound $h(Y)$ within a restricted region that is sufficiently close to $Y$ as specified by the $o\left( \left\| Y- Z \right\|_{{\mathcal F}} \right)$ term; however, as we move away from $Y$ the affine bound need not hold, which allows for a meaningful, non-empty subdifferential in broader non-convex regimes.  Based on this definition, we have the following:
\begin{theorem} \label{thm:local_convergence}
Assume that $W_p$ and $W_f$ are PSD, $\phi$ is continuous with a non-empty Fr\'{e}chet subdifferential for all $Y$\footnote{For minor technical reasons, we also assume $\phi$ is such that $\lim_{\|Y\|\to\infty}\ell_Y(Y; {\mathcal W}, f, \rho, \widetilde{B},\phi) = \infty$.} and $\rho$ is a concave, non-decreasing function with Lipschitz-continuous gradients.  Additionally, starting from any initialization $Y^{(0)}$, let $ \left\{Y^{(k)} \right\}_{k=0}^\infty$ denote a sequence generated by $Y^{(k+1)} = \text{\normalfont prox}_\phi\left( U^{(k)}\right)$ with step size parameter $\alpha \in \left(0,1/{\mathcal L} \right]$.  Then all accumulation points of $\left\{Y^{(k)} \right\}_{k=0}^\infty$ are stationary points of (\ref{eq:general_unfolded_objective}).  Furthermore, $\lim_{k\rightarrow \infty} \ell_Y(Y^{(k)}; {\mathcal W}, f, \rho, \widetilde{B},\phi) = \ell_Y(Y^*; {\mathcal W}, f, \rho, \widetilde{B},\phi)$ for some stationary point $Y^*$.
\end{theorem}
The proof is based on Zangwill's global convergence theorem \citep{Luenberger84} and additional results from \citep{sriperumbudur2009convergence}; see Appendix for further details.
\begin{corollary}\label{coro:local-but-actually-global-convergence}
If in addition to the conditions from Theorem \ref{thm:local_convergence}, $\phi$ is non-expansive and $\rho$ is chosen such that the composite function $\rho\left[ \left( \cdot \right)^2\right]$ is convex, then $\lim_{k\rightarrow \infty} \ell_Y(Y^{(k)}; {\mathcal W}, f, \rho, \widetilde{B},\phi) = \ell_Y(Y^*; {\mathcal W}, f, \rho, \widetilde{B},\phi)$, where $Y^*$ is a global minimizer of (\ref{eq:general_unfolded_objective}).
\end{corollary}
These results both apply to situations where the $k$-dependent UGNN graph propagation operator $P^{(k)} \triangleq I - \widetilde{B}^\top \Gamma^{(k)} \widetilde{B}$ is nonlinear by virtue of the $Y^{(k)}$-dependency of $\Gamma^{(k)}$, i.e., $P^{(k)} Y^{(k)} W_p^s \neq P Y^{(k)} W_p^s$ for any fixed $P$.  In contrast, it is unknown (and difficult to determine) if general nonlinear alternatives to $P Y W_p$ in (\ref{eq:IGNN_iteration}) will converge.

\subsection{Recap of Relative IGNN and UGNN Flexibility}

In terms of model expressiveness, the advantage of the IGNN framework is two-fold: (i) it allows for asymmetric weights $W_p$ while still providing a strong convergence guarantee, and (ii) it allows for decreasing activation functions.  However, the latter is likely much less relevant in practice, as most deep models adopt some form of non-decreasing activation anyway, e.g., ReLU, etc.

In contrast, UGNN models are more flexible than IGNN in their accommodation of: (i) nonlinear graph propagation through the graph attention mechanism that results from a nonlinear $\rho$ as mentioned in Section \ref{sec:ugnn_descent}, and (ii) expansive proximal operators.  While the latter may seem like a mere technicality, expansive proximal operators actually undergird a variety of popular sparsity shrinkage penalties, which have been proposed for integration with GNN models \citep{zheng2021framelets}.  For example, the $\ell_0$ norm and  related non-convex regularization factors \citep{chen2017strong,fan2001variable} are expansive and can be useful for favoring parsimonious node embeddings.  And overall, with only mild assumptions, UGNN at least guarantees cost function descent across a broad class of models per Lemma \ref{lem:descent}, with even convergence to stationary points possible in relevant situations from Theorem \ref{thm:local_convergence} that IGNN cannot emulate.



\section{How Limiting Are Symmetric (Layer-Tied) Propagation Weights?} \label{sec:shared_weights}


Previously we observed that the primary advantage IGNN has over UGNN, at least in terms of model expressiveness, is that IGNN places no restriction that the propagation weight matrix $W_p$ need be symmetric, although both models ultimately involve an architecture with layer-tied weights unlike typical message-passing models such as GCN.  To better understand the implications of this distinction, we will now explore the expressiveness of GNN models with symmetric, layer-tied weights.  We remark that resolution of these issues is of growing relevance, as evidenced by a number of recent papers.  

For example, \citep{di2023understanding} explores what can be categorized as UGNN models and the emergence of symmetric weights; however, there is no direct analysis of the inevitability of symmetric weights, the expressiveness relative to GNNs with asymmetric weights, nor consideration of the limitations of layer-tied weights.  Additionally, relative to our general UGNN framework emanating from (\ref{eq:general_unfolded_objective}), their approach excludes $\rho$ and non-linear activations.  Critically, without the latter, our Theorem \ref{thm:equivalency-finite} presented below is not possible.  In a related vein, \citep{luo2022unifying} addresses the expressiveness of weight-tied GNN layers relative to GCN models, although these results are a special case of ours, and symmetric weights are not considered.\footnote{We remark that while \citep{di2023understanding} and \citep{luo2022unifying} represent valuable, complementary contributions that have been independently introduced by others, our work was both posted to arxiv and submitted to JMLR well in advance, and hence should not be viewed as derivative of them. [\underline{Note}: This footnote is to provide context specifically for reviewing purposes and can be removed later.]}  More broadly, symmetric weights naturally emerge when deriving unfolding architectures analogous to common deep feed-forward models such as MLPs, ResNets, or Transformers \citep{amos2017optnet,frecon2022bregman,xie2023optimization,yang2022transformers}.  However, even these works do not quantify the expressiveness relative to unconstrained weights as we investigate here.



\subsection{Fixed-Point Equivalency with Symmetric and Asymmetric Weights}\label{sec:fix-point-equivalency}

In this section we examine situations whereby UGNN models are able to reproduce, up to some inconsequential transform, the fixed points of an IGNN, even when the latter includes an asymmetric propagation weight matrix $W_p$.  However, because it is challenging to analyze general situations with arbitrary activation functions, we first consider the case where $\sigma$ is an identity mapping.  We then have the following:


\begin{theorem}\label{theo:equivalency-linear}
For any $W_p \in \mathbb R^{d \times d}$, $W_x \in \mathbb R^{d_0 \times d}$, $X \in \mathbb R^{n\times d_0}$, and $P$ that admit a unique IGNN fixed point $Y^* = PY^*W_p + f(X;W_x)$, there exists a $Y' \in \mathbb R^{n\times d'}$, $\tilde W_x \in \mathbb R^{d\times d'}$, right-invertible transform $T \in \mathbb R^{d\times d'}$, and symmetric $W_p^s \in \mathbb R^{d'\times d'}$, such that  
\begin{equation}
Y' T = P Y' T W_p^s + f(X;W_x) \tilde W_x, ~~~\text{with}~~\|Y'-Y^*\|_{\mathcal F} < \epsilon, ~\forall \epsilon > 0.
\end{equation}
\end{theorem}
This result implies that a UGNN model with $\phi = 0$ can produce node-wise embeddings $Y = Y'T$ capable of matching any IGNN fixed-point with arbitrary precision up to transforms $T$ and $\tilde W_x$.  And given that $T$ can be absorbed into the meta-loss output layer $g$ from (\ref{eq:meta_loss}) (which is often a linear layer anyway), and $\tilde W_x$ can be absorbed into $f$, the distinction introduced by these transforms is negligible.


Proceeding further, if we allow for nonlinear activation functions $\sigma$, we may then consider a more general, parameterized family of penalty functions $\phi({\boldsymbol{y}};W_\phi)$ such that the resulting proximal operator increases our chances of finding a UGNN fixed point that aligns with IGNN.  However, some care must be taken to control UGNN model capacity to reduce the possibility of trivial, degenerate alignments.  To this end, we consider proximal operators in the set 
\begin{equation}
\mathcal S_\sigma = \{\mathrm{prox}_\phi: {\boldsymbol{y}} \mapsto G\sigma(C {\boldsymbol{y}}) | \text{with } \{ G,C \} \text{ chosen such that } \mathrm{prox}_\phi \text{ is proximal operator.}\},
\end{equation}
where the matrices $G$ and $C$ can be aggregated into $W_\phi$.  It is then possible to derive sufficient conditions under which UGNN has optimal solutions equivalent to IGNN fixed points (it remains an open question if a necessary condition can be similarly derived). See Appendix for details.

\subsection{UGNN Capacity to Match Canonical GCN Architectures}


The previous section considered the alignment of fixed points, which are obtainable after executing potentially infinite graph propagation steps, from models with  and without symmetric propagation weights.  Somewhat differently, this section investigates analogous issues in the context of the possible embeddings obtainable after $k$ steps of graph propagation.  Specifically, we evaluate the expressiveness of a canonical GCN model with arbitrary, layer-independent weights versus a UGNN model structured so as to effectively maintain an equivalent capacity (up to a right-invertible linear transformation as discussed above).



\begin{theorem}\label{thm:equivalency-finite} Let $Y_{\mathrm{GCN}}^{(k+1)} = \sigma\left( P Y_{\mathrm{GCN}}^{(k)}W_p^{(k)} + \beta Y_{\mathrm{GCN}}^{(k)}\right)$ denote the $k$-th layer of a GCN, where $W_p^{(k)}$ is a layer-dependent weight matrix, $\sigma = \mathrm{prox}_\phi$~ for some function $\phi$, and $\beta \in \{0,1\}$ determines whether or not a residual connection is included.  Then with $\alpha = 1$, $\rho$ an identity mapping, and $f(X;W_x)$ set to zero, there will always exist a right-invertible $T$, initialization $Y^{(0)}$, and symmetric weights $W_p^s$ and $W_f^s$ such that the $k$-th iteration step computed via (\ref{eq:prox_step}) satisfies $Y^{(k+1)}T = Y^{(k+1)}_{\mathrm{GCN}}$.

\end{theorem}



Given the close association between GCNs and certain UGNN models, Theorem \ref{thm:equivalency-finite}  suggests that, at least in principle, symmetric layer-tied weights may not be prohibitively restrictive in the finite layer regime.  And as can be observed from the proof, this equivalency result is possible using a matching parameter budget instantiated through an expanded hidden dimension but constrained via block-sparse weight matrices $W_p^s$ and $W_f^s$.  Of course in practice when we extrapolate outside of regimes where the results of this section strictly hold, it remains to be seen if symmetric weight constraints have a significant adverse impact.  We explore such considerations next.


\section{Empirical Accuracy Comparisons} \label{sec:experiments}


In this section we evaluate the predictions of various comparable IGNN and UGNN models across experimental settings designed to complement our theoretical findings and showcase potential strengths and weaknesses that may arise in practice.  To this end, we first conduct a series of quasi-synthetic tests designed  to explore the interplay between the type of graph propagation weights (symmetric as with UGNN vs asymmetric as with IGNN), the graph propagation path length (finite as with UGNN vs approximately infinite as with IGNN), and model expressiveness.
We then turn to real-world benchmarks to examine relative capabilities handling long-range dependencies, graph heterophily, and adversarial attacks, all of which represent domains within which either IGNNs, UGNNs, or both have been motivated in prior work.  Under these different application scenarios, we also evaluate against popular methods specifically designed for the particular task at hand.   As for IGNN/UGNN model architectures, we assume the following:

\begin{itemize}
    \item \textbf{IGNN baselines}:  We use (\ref{eq:IGNN_iteration}) as the grounding for IGNN baselines with the following notable specializations.  First, when $\sigma$ is an identity, we refer to the resulting model as EIGNN following \citep{eignn}, and when $\sigma$ is a ReLU activation with relaxed weight constraints on $W_p$ from \citep{MIGNN}, the model is denoted MIGNN (because of its origins from monotone operator theory).  We also remark that beyond these candidates, there is as of yet no more expressive IGNN parameterization that we are aware of that comes equipped with an implementation suitable for stable training and implicit differentiation across large graphs.

    \item \textbf{UGNN baselines}:  For UGNN we limit our exploration to various special cases of the general framework introduced in Section \ref{sec:ugnn_intro} and instantiated via the composite update described by 
    (\ref{eq:basic_grad_step}), (\ref{eq:gamma_update}), and (\ref{eq:prox_step}).  To recap the role of these steps, when conditioned on fixed graph attention (\ref{eq:basic_grad_step}) computes an  affine filter that includes a graph propagation operator (as with IGNNs), (\ref{eq:gamma_update}) serves as a form of nonlinear graph attention operator (unique to currently-available UGNNs), and (\ref{eq:prox_step}) produces the nonlinear activation function for each layer (analogous to $\sigma$ for IGNNs).  While UGNNs also accommodate expansive proximal operators beyond $\sigma$, we do not consider this flexibility herein.
\end{itemize}

\noindent Further specification of dataset-specific architectural choices and hyperparameters will be provided in the forthcoming subsections, while full experimental details are deferred to the Appendix.  All models and testing were implemented using the Deep Graph Library (DGL) \citep{wang2019dgl}.  We also note that portions of these experiments have appeared previously in our conference paper \citep{yang2021}, which presents a useful UGNN-related modeling framework and supporting DGL-based code that we further augmented for our purposes herein.

\subsection{Exploring the Practical Impact of Symmetric Weights}\label{sec:label-recovering}

From our analysis in Section \ref{sec:shared_weights} it is reasonable to speculate that UGNN symmetric weights are not a significant limitation relative to IGNN when it comes to model expressiveness, potentially even in regimes that deviate from the stated technical conditions, or when UGNN has limited propagation steps as may occur in practical implementations.  To explore this hypothesis empirically, we design the following quasi-synthetic experiment: First, we train separate IGNN and UGNN models on the ogbn-arxiv dataset \citep{HuFZDRLCL20}, where the architectures are equivalent with the exception of $W_p$ and the number of propagation steps (see Appendix for all network and training details). Additionally, because UGNN requires symmetric propagation weights, a matching parameter count can be achieved by simply expanding the UGNN hidden dimension.  Once trained, we then generate predictions from each model and treat them as new, ground-truth labels.  We next train three additional models separately on both sets of synthetic labels:

\begin{enumerate}[leftmargin=1cm]
\item An IGNN with architecture equivalent to the original used for generating labels,
\item An analogous UGNN, but with limited propagation steps consistent with practical use cases,
\item A hybrid model that can equivalently be viewed as either an IGNN with finite propagation steps, or a UGNN where the restriction on symmetric weights has been relaxed.
\end{enumerate}

\noindent For all three models above the number of parameters is a small fraction of the number of labels to mitigate overfitting issues.

\begin{table}[htbp]
    \centering
    \begin{tabular}{c|cc|c}
    \toprule
    \diagbox{Gen}{Rec} 
          & IGNN & UGNN & Hybrid  \\ \midrule
        IGNN & 95.7 $\pm$ 0.9 & 89.8 $\pm$ 1.9 & 91.6 $\pm$ 1.2 \\
        UGNN &91.0 $\pm$ 1.5  & 94.5 $\pm$ 0.8 & 92.3 $\pm$ 0.7 \\
        \bottomrule
    \end{tabular}
    \caption{Accuracy recovering quasi-synthetic labels.}\label{tab:label-recovering}
\end{table}


Prediction accuracy results are presented in Table \ref{tab:label-recovering}, where the rows indicate the label generating models, and the columns denote the recovery models.  We first consider the leftmost two columns labeled IGNN and UGNN.  As would be expected, the highest recovery accuracy (95.7 and 94.5) occurs when the generating and recovery models are matched such that perfect recovery is theoretically possible by design.  But even in the mismatched cases (91.0 and 89.8) the performance is still quite strong, suggesting that a UGNN with symmetric weights and limited propagation steps can reasonably approximate an IGNN, and vice versa.  

To further probe the source of the modest gap that nonetheless does exist, we next examine the rightmost hybrid column of Table \ref{tab:label-recovering}.  Here we observe that the hybrid model performs somewhat worse recovering IGNN data (91.6 vs 95.7), indicating that truncated propagation steps, potentially more so than symmetric weights,  reduces performance when the generating model involves true long-range propagation as with IGNN.  Not surprisingly, we also notice that when trained on UGNN data, the hybrid model outperforms IGNN (92.3 vs 91.0) since here there is no benefit to modeling long-range propagation beyond the limit of the truncated UGNN propagation steps.

In general though, the fact that the performance of all models varies within a range of a few percentage points, even under these idealized conditions where the data-generating model has maximal advantage, suggests a significant overlap of model expressiveness as might be expected per the analysis in Section \ref{sec:shared_weights}.  Hence UGNN models with symmetric weights and finite propagation steps are likely adequate in many practical scenarios.



\subsection{Results Using Long-Dependency Benchmarks}\label{sec:long-range-result}

\begin{figure}[htbp]
    \centering
    \includegraphics[scale=0.5]{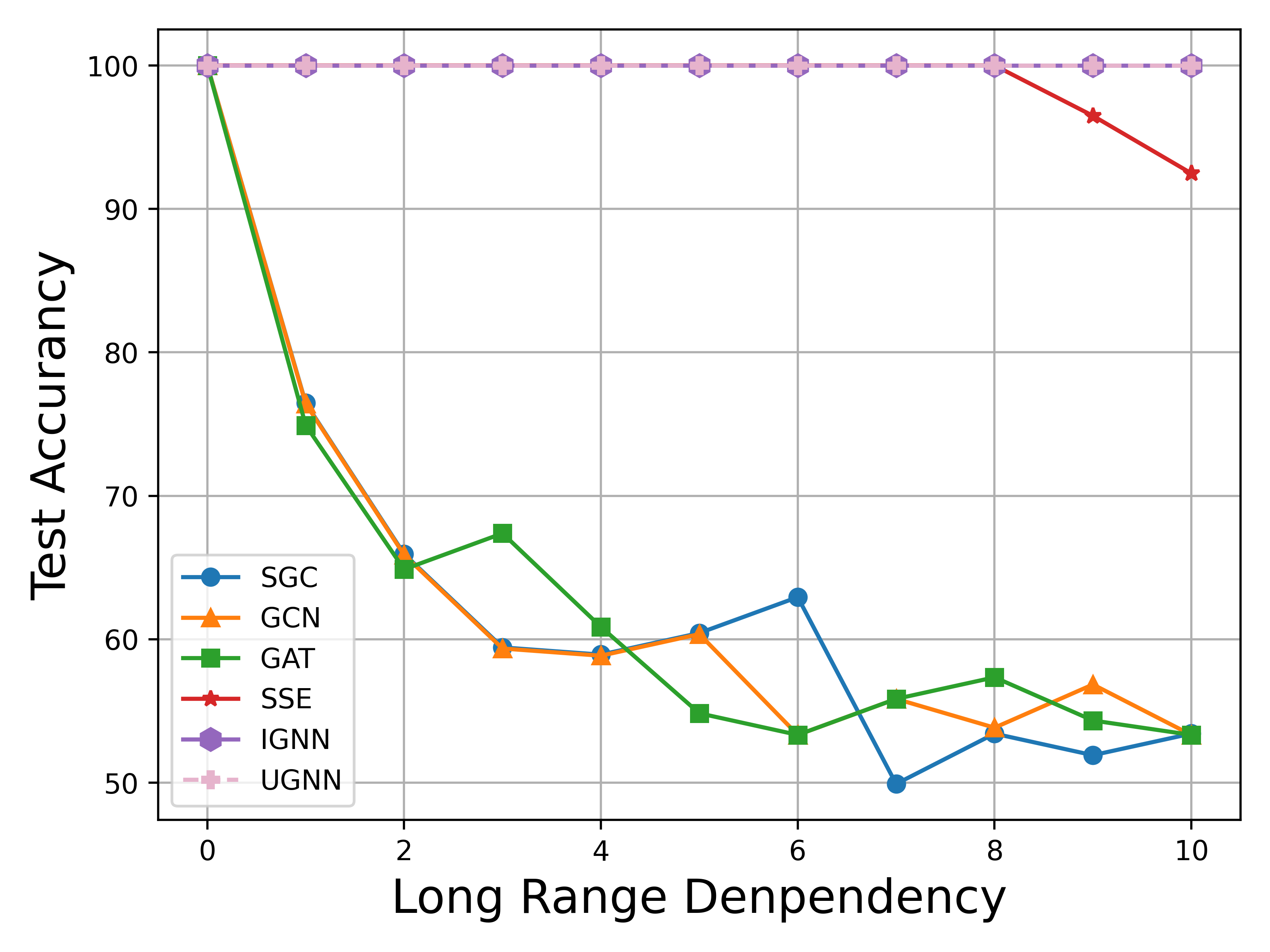}
    \caption{Accuracy comparisons on long-range Chains dataset.}
    \label{fig:chains}
\end{figure}

We now turn to experiments involving existing graph benchmarks purported to involve long-range dependencies, a scenario that has been previously used to motivate both UGNN and IGNN models.  The modeling goal here is to capture these long-range effects by stably introducing a sufficient number of propagation layers without the risk of oversmoothing, which might otherwise degrade performance.  We begin with the synthetic Chains node classification dataset \citep{GuC0SG20,eignn} explicitly tailored to evaluate GNN models w.r.t.~long-range dependencies.  Results of IGNN and UGNN models, as well as baselines SGC \citep{DBLP:conf/icml/ChenWHDL20}, GCN \citep{kipf2017semi}, GAT \citep{VelickovicCCRLB18},  and SSE \citep{DaiKDSS18} are shown in Figure \ref{fig:chains}.  Baseline results are from \citep{GuC0SG20} without error bars; however, in our own
testing we observe the results to be quite stable across trials such that error bars are not necessary for conveying the core message.  Note that SSE, like IGNN and to some extent UGNNs, is explicitly designed for capturing long-range dependencies.  While the Chains dataset has been shown in the past to differentiate IGNN and SSE from traditional GNN architectures \citep{GuC0SG20}, we observe here that UGNN is equally capable of matching performance, and hence a more challenging benchmark is needed.

For this purpose we next turn to the Amazon Co-Purchase dataset, a node classification benchmark often used for evaluating long-range dependencies, in part because of sparse labels relative to graph size \citep{MIGNN,DaiKDSS18,GuC0SG20}.\footnote{An alternative is the Long-Range Graph Benchmark \citep{dwivedi2023benchmarking}; unfortunately though, this benchmark is exclusively composed of tiny graphs for solving graph-level prediction problems whereby scalable UGNN and IGNN node-level embeddings are less relevant. Yet another option is PPI \citep{graphsage}; however, multiple existing methods can already achieve nearly 100\% accuracy \citep{GuC0SG20,eignn}, so it is less impactful for differentiation.}  We adopt the test setup from \citep{DaiKDSS18,GuC0SG20,MIGNN}, and compare performance using different label ratios.  We report results from five baselines, namely SGC, GCN, SSE, IGNN, and MIGNN, where the latter is a more expressive version of IGNN. Regardless, UGNN is still able to outperform all of them by a non-trivial margin as shown in Figures \ref{fig:amazon-micro} and \ref{fig:amazon-macro}.  This likely indicates that while some long-range capability is needed to outperform the canonical GNNs, the unbounded extent of IGNNs provides diminishing marginal returns beyond the UGNN, which has greater flexibility in other relevant aspects.  We also remark that, to the degree that MIGNN is more expressive than prior IGNNs, this does not translate into improved classification performance, possibly because the core architecture is still the same as discussed in Section \ref{sec:ignn_intro}.

\vspace*{0.5cm}
\begin{minipage}{\linewidth}
\begin{minipage}[t]{0.49\linewidth}
    \centering
    \includegraphics[scale=0.45]{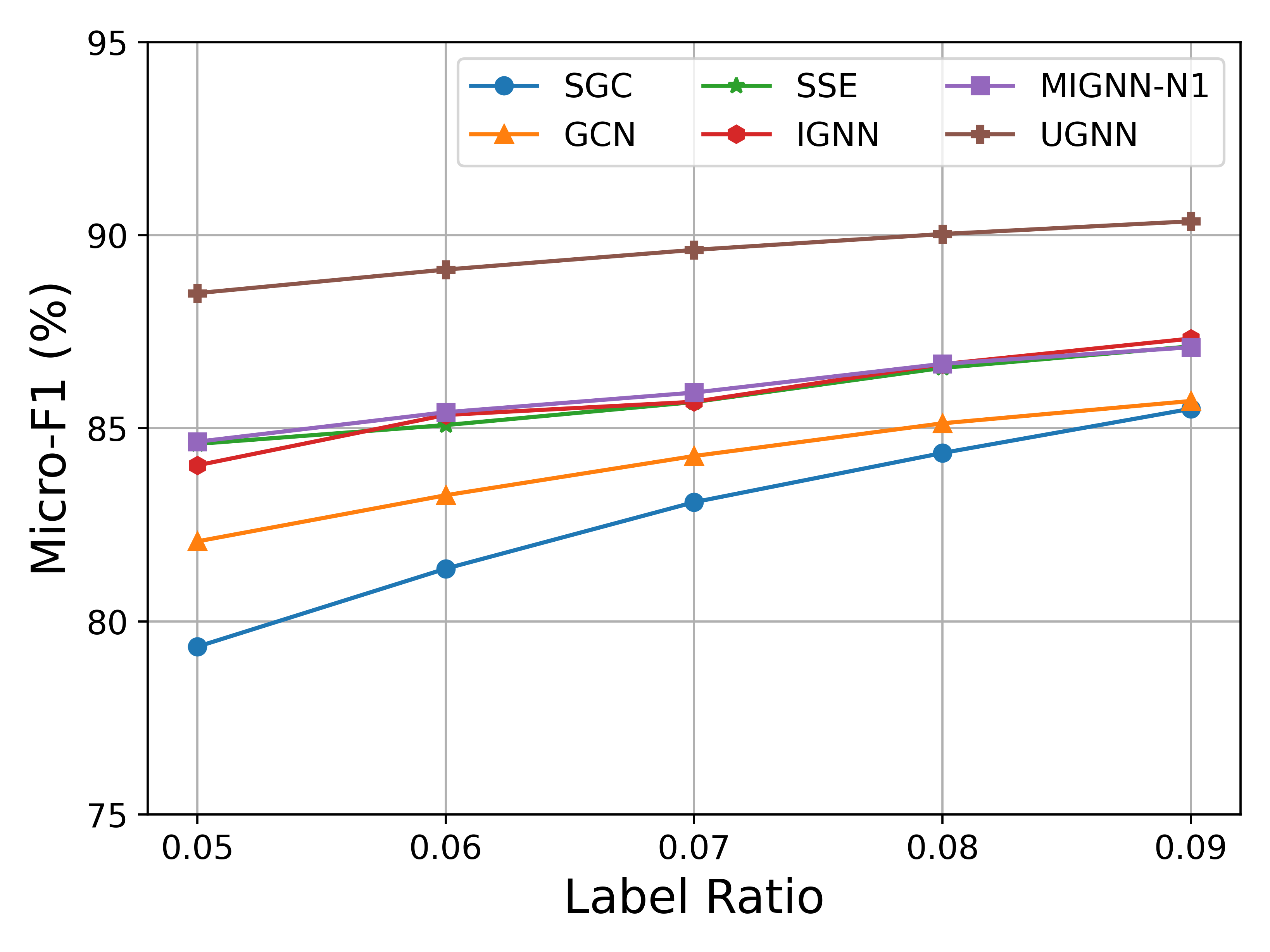}
    \captionof{figure}{Micro-F1 on Amazon Co-Purchase.}
    \label{fig:amazon-micro}
\end{minipage}
\begin{minipage}[t]{0.49\linewidth}
    \centering
    \includegraphics[scale=0.45]{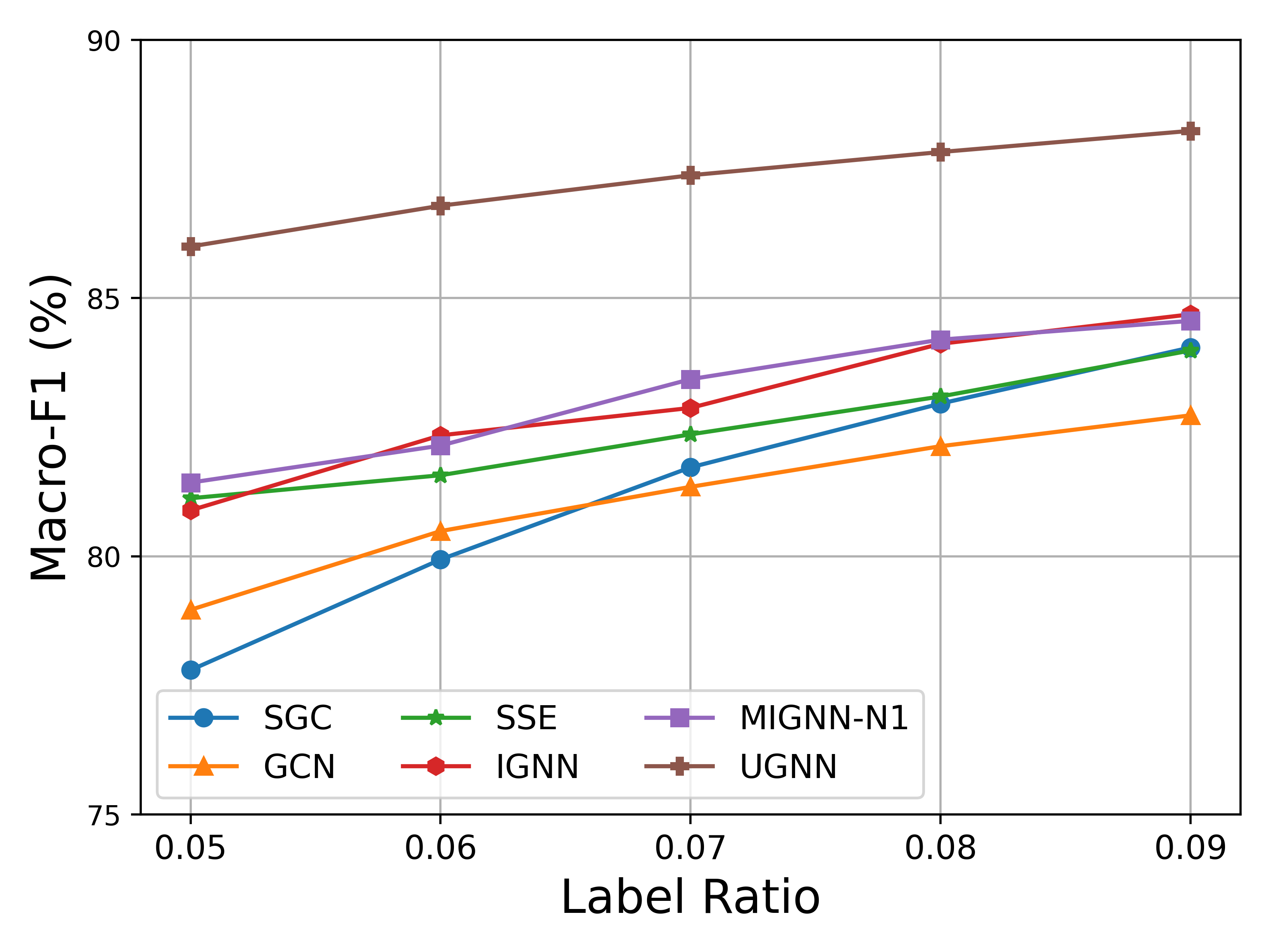}
     \captionof{figure}{Macro-F1 on Amazon Co-Purchase.}
    \label{fig:amazon-macro}
\end{minipage}
\end{minipage}



\subsection{Results on Heterophily Graphs}
\label{sec:hetero-result}

A \textit{homophily} graph occurs when nearby nodes share similar labels and/or features, and message-passing GNNs are generally well-equipped to handle them.  In contrast, \textit{heterophily} graphs exhibit an opposing phenomena, namely, nodes sharing an edge have a reduced likelihood of having the same label.  These complementary notions are often quantified via the homophily ratio 
\begin{equation}
{\mathcal H} \triangleq \frac{|\{(u,v)\in\mathcal{E} ~:~{\boldsymbol{t}}_u={\boldsymbol{t}}_v\}|}{|\mathcal{E}|}
\end{equation}
from \citep{DBLP:conf/nips/ZhuYZHAK20}, where ${\boldsymbol{t}}_u$ and ${\boldsymbol{t}}_v$ are the target labels of nodes $u$ and $v$  respectively. ${\mathcal H}$ quantifies the tendency of nodes to be connected with other nodes from the same class. Graphs with ${\mathcal H} \approx 1$ exhibit strong homophily, while conversely, those with ${\mathcal H} \approx 0$ show strong heterophily, indicating that many edges are connecting nodes with different labels.

With this in mind, it has been suggested that GNNs with more sensitivity to long-range dependencies might be well-suited for handling heterophily graphs \citep{eignn,PeiWCLY20}, since nodes sharing the same label may now be concentrated many hops away, i.e., a form of long-range association.  But we conjecture that the utility of such long-range GNNs will depend on attributes of the heterophilic graph not directly measured by the homophily ratio.  More concretely, the implicit assumption is that if we increase the receptive field of a given node, the percentage of nodes with the same label will be higher, improving model performance.  

\subsubsection{A Closer Look at Heterophily and Long-Range Dependencies} \label{sec:ASD_intro}

\begin{wrapfigure}[22]{r}{0.5\textwidth}\vspace{-4em}
  \begin{center}
    \includegraphics[scale=0.5]{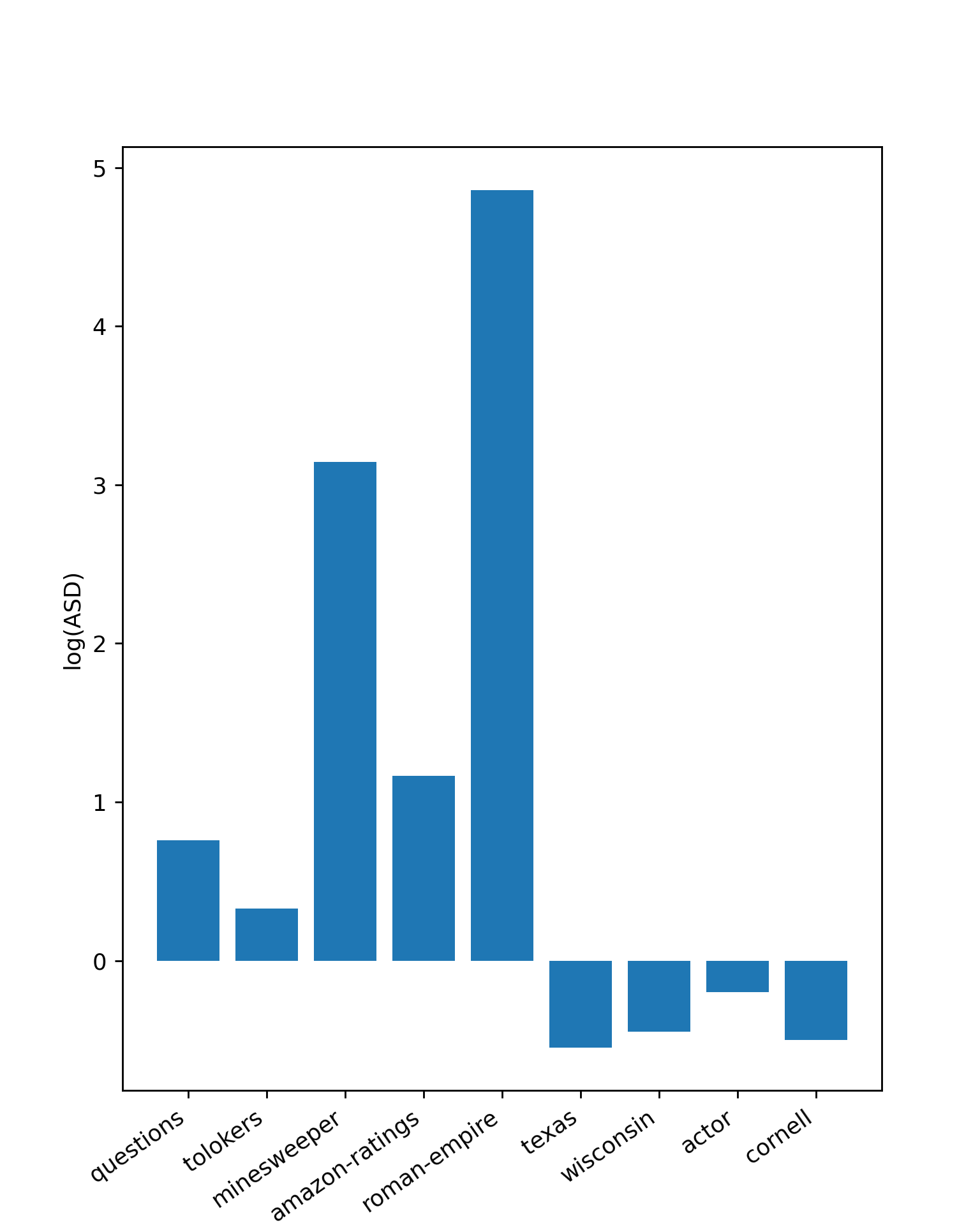}
  \end{center}
  \caption{Average same-label distance (ASD) on heterophily benchmarks.}
  \label{fig:avg-dist}
  \vspace*{-0.3cm}
\end{wrapfigure}

To kick the tires on the above assumption, we first compute an additional metric across a variety of heterophily graphs.  We name this metric the \textit{average same-label distance} (ASD), with formal definition given by
\begin{equation} \label{eq:ASD}
\mathrm{ASD} \triangleq \frac{1}{c}\mathbb E_{u \in \mathcal V} \mathbb E_{v \in \mathcal L_u} \mathrm{dist}(u,v),
\end{equation}
where ${\mathcal L}_u$ denotes the set of nodes with the same label as node $u$, $c$ is the number of classes as introduced previously, and $\mathrm{dist}(u,v)$ counts the shortest path length between nodes $u$ and $v$.  To reduce computational complexity, we compute the expectation in (\ref{eq:ASD}) by randomly sampling 1000 nodes as $u$ and then averaging.  The resulting log-scale ASD values are presented in Figure \ref{fig:avg-dist} across four classic heterophily datasets examined in \citep{PeiWCLY20},  Actor, Cornell, Texas, and Wisconsin, and the more recent heterophily benchmarks Roman, Amazon, Minesweeper, Tolokers, and Questions from \citep{new-hetero}. 

\vspace*{-0cm}

Interestingly, there is a clear demarcation between the new heterophily datasets and the classical ones: The new datasets unanimously possess a larger $\log(\mathrm{ASD} )$ (all $\log(\mathrm{ASD} ) > 1$) while the classical ones are smaller (all $\log(\mathrm{ASD} ) < 1$).  We will now empirically explore the impact of this distinction on performance.  To streamline this treatment, we reference the datasets from \citep{new-hetero} with $\log(\mathrm{ASD} ) > 1$ as $\mathrm{ASD}_{\mathrm{sup}}$, while the datasets from \citep{PeiWCLY20} with $\log(\mathrm{ASD} ) < 1$ are referenced as $\mathrm{ASD}_{\mathrm{inf}}$.

\subsubsection{How Different Manifestations of Heterophily Impact Performance}

Intuition suggests that incorporating long-range dependencies will be more effective on the $\mathrm{ASD}_{\mathrm{sup}}$ datasets, since a larger effective receptive field, as occurs with greater numbers of propagation steps over the graph, can potentially lead to aggregation over more nodes with the same label.  In contrast, for $\mathrm{ASD}_{\mathrm{inf}}$ heterophily graphs,  same-label nodes may actually be relatively close together, but mixed in with appreciable concentrations of mismatched-labeled nodes.  Under these conditions long-range propagation may be less effective and instead, treating edges between mismatched nodes as outliers a preferable alternative.  For UGNN models, such treatment can be operationalized by choosing $\rho$ as a concave non-decreasing function as espoused in Section \ref{sec:flexible_energy}.  And as alluded to in Section \ref{sec:ugnn_descent}, this leads to an emergent form of graph attention that is capable of selectively pruning bad edges.

\begin{table*}[htbp]
\begin{center}
\begin{sc}
\begin{tabular}{l|cccc|c}
\toprule
\textbf{Dataset}         & \textbf{Texas}          & \textbf{Wisconsin}      & \textbf{Actor}          & \textbf{Cornell}   & \textbf{Avg.}     \\
\midrule
Hom.~Ratio (${\mathcal H}$)             & 0.11                    & 0.21                    & 0.22                    & 0.3        &             \\
\midrule
GCN                    & 59.46$\pm$5.25          & 59.80$\pm$6.99          & 30.26$\pm$0.79          & 57.03$\pm$4.67     & 51.64   \\
GAT                    & 58.38$\pm$4.45          & 55.29$\pm$8.71          & 26.28$\pm$1.73          & 58.92$\pm$3.32     &   49.72  \\
GraphSAGE              & 82.43$\pm$6.14          & 81.18$\pm$5.56          & 34.23$\pm$0.99          & 75.95$\pm$5.01    &  68.45    \\
Geom-GCN               & 67.57                   & 64.12                   & 31.63                   & 60.81         &    56.03      \\
$\text{H}_2\text{GCN}$ & \textbf{84.86$\pm$6.77} & \textbf{86.67$\pm$4.69}          & \textbf{35.86$\pm$1.03}          & \textbf{82.16$\pm$4.80}      &  \textbf{72.39}  \\
MLP                    & 81.89$\pm$4.78          & 85.29$\pm$3.61          & 35.76$\pm$0.98          & 81.08$\pm$6.37      &  71.01  \\
\midrule
IGNN      & 83.1$\pm$5.6 & 79.2$\pm$6.0 & 35.5$\pm$2.9 & 76.0$\pm$6.2  & 68.45 \\
EIGNN      & 80.54$\pm$5.67 & 83.72$\pm$3.46  & 34.74$\pm$1.19 &  81.08$\pm$5.70   & 70.02 \\
UGNN(w/o $\rho$)                   & 81.62$\pm$5.51          & 82.75$\pm$7.83          & 37.10$\pm$1.07          & 83.51$\pm$7.30     &   71.25   \\
UGNN(w/ $\rho$)              & \textbf{84.59$\pm$3.83}          & \textbf{86.67$\pm$4.19} & \textbf{37.43$\pm$1.50} &  \textbf{86.76$\pm$5.05} & \textbf{73.86} \\
\bottomrule
\end{tabular}
\end{sc}
\captionof{table}{Node classification accuracy on $\mathrm{ASD}_{\mathrm{inf}}$ heterophily graphs.  Boldface indicates the best performing model within each block.  When equipped with a concave $\rho$, a simple/generic UGNN model outperforms IGNN models and is competitive with strong baselines explicitly designed for heterophily graphs.\vspace{-2em}}
\label{table:heterophily-results}
\end{center}
\hfill
\end{table*}

To test this hypothesis, we first train both IGNN and UGNN models on the four $\mathrm{ASD}_{\mathrm{inf}}$ datasets from Section \ref{sec:ASD_intro}, using the data splits, processed node features, and labels provided by \citep{PeiWCLY20}.  Notably, we include UGNN models both \textit{with} and \textit{without} a concave $\rho$ function (by ``without" here we just mean that $\rho$ is an inconsequential identity mapping). Node classification accuracy results are presented in Table \ref{table:heterophily-results}, where for context, we also include results from the additional GNN architectures GCN \citep{kipf2017semi}, GAT \citep{VelickovicCCRLB18}, GraphSAGE \citep{HamiltonYL17}, SOTA heterophily methods GEOM-GCN \citep{PeiWCLY20} and $\text{H}_2\text{GCN}$ \citep{DBLP:conf/nips/ZhuYZHAK20}, as well as a baseline MLP as reported in \citep{DBLP:conf/nips/ZhuYZHAK20}.  Two things stand out regarding these results.  First, the IGNN, EIGNN, and UGNN (without $\rho$) models, are comparable or worse than the MLP, subject to some dataset-to-dataset variability, indicating that merely introducing long-range dependencies alone may not necessarily offer a clear-cut advantage over ignoring the graph structure altogether.  In contrast, UGNN (with $\rho$) is uniformly better than the other IGNN/UGNN models, as well as the MLP.  This observation is consistent with the notion that heterophily of the $\mathrm{ASD}_{\mathrm{inf}}$ variety can be mitigated using robust/concave penalization that mutes the impact of edges connecting nodes with mismatched labels.  This strategy also allows this UGNN model to even outperform strong GNN architectures explicitly designed for heterophily graphs, e.g., $\text{H}_2\text{GCN}$.

To provide complementary evidence for our hypothesis, we next train IGNN and UGNN models on the five $\mathrm{ASD}_{\mathrm{sup}}$ benchmarks.  With these datasets heterophily manifests more as a longer average distance between nodes sharing a label, rather than  spurious edges connected to any given node that can be adequately treated as outliers.  As such, we might expect that the effectiveness of $\rho$ will be marginal relative to long-range graph propagations that are tantamount to a large receptive field. Results are shown in Table \ref{table:heterophily-results-new}, along with a set of recent heterophily GNN models tested in \citep{new-hetero} for reference.  These models include H$_2$GCN \citep{h2gcn}, GPR-GCN \citep{gprgcn}, FSGNN \citep{fsgcn}, FAGNCN \citep{fagcn}, GBK-GNN \citep{gbkgnn} and JacobiConv \citep{jacobgnn}.  We also include an MLP as well as GCN and GAT baselines for reference.\footnote{For GCN and GAT we ran the experiments ourselves using the standard implementations from DGL.  While \citep{new-hetero} also includes results associated with a number of popular baseline GNN architectures (e.g., GCN, GAT, GraphSAGE, GT, and variants thereof etc.), upon inspection of the associated public codebases, we find that the actual implementations involve additional tweaks/enhancements which deviate from the original models.  These and other types of modifications may be effective on specific tasks/datasets, and are interesting to consider; however, they can be applied across a wide variety of GNN architectures and are well outside of our analysis-driven priorities here.}


From these results, we observe that all UGNN/IGNN-based models significantly  outperform the MLP, suggesting that same labels spread across a wider receptive field can in fact be exploited within the $\mathrm{ASD}_{\mathrm{sup}}$ datasets.  However, the UGNN with concave $\rho$ no longer offers an advantage, in that IGNN and UGNN (regardless of $\rho$) perform similarly.  Additionally, both IGNN and UGNN are competitive with the best performing heterophily-motivated GNNs, e.g., FSGNN and GBK-GNN.

\begin{table*}[htbp]
\begin{center}\begin{sc}
\begin{footnotesize}
\begin{tabular}{l|ccccc|c}
\toprule
\textbf{Dataset}         & \textbf{Roman}          & \textbf{Amazon}      & \textbf{Minesweeper}          & \textbf{Tolokers}  & \textbf{Questions} & \textbf{Avg.} \\
\midrule
GCN & 53.31 $\pm$ 1.55 & 49.86 $\pm$ 0.21 & 77.49 $\pm$ 0.60 & 76.53 $\pm$ 0.11 &  76.61 $\pm$ 0.41 & 66.76
\\ GAT & 71.31 $\pm$ 0.55 & 50.52 $\pm$ 0.52 & 88.86 $\pm$ 0.51 & 75.72 $\pm$ 6.19 & 74.34 $\pm$ 0.38 & 72.15
\\ 
H$_2$GCN & 60.11 $\pm$ 0.52 & 36.47 $\pm$ 0.23 & 89.71 $\pm$ 0.31 & 73.35 $\pm$ 1.01 & 63.59 $\pm$ 1.46 & 64.65 \\ 
GPR-GNN & 64.85 $\pm$ 0.27 & 44.88 $\pm$ 0.34 & 86.24 $\pm$ 0.61 & 72.94 $\pm$ 0.97 & 55.48 $\pm$ 0.91 & 64.88 \\ 
FSGNN & \textbf{79.92 $\pm$ 0.56} & \textbf{52.74 $\pm$ 0.83} & 90.08 $\pm$ 0.70 & \textbf{82.76 $\pm$ 0.61} & \textbf{78.86 $\pm$ 0.92} & \textbf{76.87} \\ 
FAGCN & 65.22 $\pm$ 0.56 & 44.12 $\pm$ 0.30 & 88.17 $\pm$ 0.73 & 77.75 $\pm$ 1.05 & 77.24 $\pm$ 1.26 & 70.50 \\ 
GBK-GNN & 74.57 $\pm$ 0.47 & 45.98 $\pm$ 0.71 & \textbf{90.85} $\pm$ 0.58 & 81.01 $\pm$ 0.67 & 74.47 $\pm$ 0.86 & 73.38 \\ 
JacobiConv & 71.14 $\pm$ 0.42 & 43.55 $\pm$ 0.48 & 89.66 $\pm$ 0.40 & 68.66 $\pm$ 0.65 & 73.88 $\pm$ 1.16 & 69.38 \\ 
MLP & 65.88 $\pm$ 0.38 & 45.90 $\pm$ 0.52 & 50.89 $\pm$ 1.39 & 72.95 $\pm$ 1.06 & 70.34 $\pm$ 0.76 & 61.19 \\
\midrule
IGNN & \textbf{84.04 $\pm$ 0.85} & 51.79 $\pm$ 0.46 & 90.73 $\pm$ 1.45 & 82.06 $\pm$ 0.72 & 75.44 $\pm$ 0.45 & 76.81
\\ EIGNN & 75.72 $\pm$ 0.29 & 45.27 $\pm$ 0.78 & 88.17 $\pm$ 0.66 & 79.90 $\pm$ 0.76 & 69.13 $\pm$ 0.65 & 71.64
\\  
UGNN(w/o $\rho$) & 82.75 $\pm$ 1.71 & 50.99 $\pm$ 0.37 & \textbf{91.41 $\pm$ 0.20} & \textbf{85.62 $\pm$ 0.17} & \textbf{75.83 $\pm$ 0.40} & \textbf{77.32}
\\ UGNN(w/ $\rho$) & 83.54 $\pm$ 0.47 & \textbf{53.38 $\pm$ 0.35} & 87.58 $\pm$ 0.08 & 84.83 $\pm$ 0.25 & 75.21 $\pm$ 0.21 & 76.91 \\ 
\bottomrule
\end{tabular}
\end{footnotesize}\end{sc}
\captionof{table}{Node classification accuracy on $\mathrm{ASD}_{\mathrm{sup}}$ heterophily graphs.  Boldface indicates the best performing model within each block.
\vspace{-2em}}
\label{table:heterophily-results-new}
\end{center}
\hfill
\end{table*}

\vspace{-1em}
\subsection{Adversarial Attack Results}\label{sec:adversarial-attack-results}

As demonstrated in the previous section, when heterophily implicitly manifests as graphs with spurious edges, UGNN models with a concave $\rho$ provide a unique advantage over IGNN models.  To further explore this advantage, we turn to an application scenario whereby spurious edges are explicit, namely, certain forms of adversarial attacks.  To this end, we compare the models using graph data corrupted via the the Mettack algorithm  \citep{DBLP:conf/iclr/ZugnerG19}.  Mettack operates by perturbing graph edges with the aim of maximally reducing the node classification accuracy of a surrogate GNN model that is amenable to adversarial attack optimization. And the design is such that this reduction is generally transferable to other GNN models trained with the same perturbed graph.  In terms of experimental design, we follow the exact same non-targeted attack  scenario from \citep{ZhangZ20}, setting the edge perturbation rate to 20\%, adopting the ``Meta-Self' training strategy, and a GCN as the surrogate model.  



\begin{table}[ht]
 \begin{center} \begin{small} \begin{sc}
\begin{tabular}{l|cc|c}
\toprule
\textbf{Model} & \textbf{ATK-Cora} & \textbf{ATK-Citeseer} & \textbf{Avg.}
\\ \midrule
    surrogate (GCN) & 57.38 $\pm$ 1.42 & 60.42 $\pm$ 1.48 & 58.90 \\ \midrule
    GNNGuard & \textbf{70.46 $\pm$ 1.03} & 65.20 $\pm$ 1.84  & \textbf{67.83} \\
    GNN-Jaccard & 64.51 $\pm$ 1.35 & 63.38 $\pm$ 1.31  & 63.95 \\
    GNN-SVD & 66.45 $\pm$ 0.76 & \textbf{65.34 $\pm$ 1.00}  & 65.90 \\ \midrule
    EIGNN & 65.16 $\pm$ 0.78 &  68.36 $\pm$ 1.01  & 66.76 \\
    IGNN & 61.47 $\pm$ 2.30 & 64.70 $\pm$ 1.26  & 63.09 \\
    UGNN(w/o $\rho$) & 62.89 $\pm$ 1.59 & 63.83 $\pm$ 1.95  & 63.36 \\
    UGNN(w/ $\rho$) & \textbf{70.23 $\pm$ 1.09} &  \textbf{70.63 $\pm$ 0.93}  & \textbf{70.43} \\
\bottomrule
\end{tabular} \end{sc} \end{small} \end{center} \vskip -0.1in
\caption{Node classification accuracy under Adversarial attacks. Boldface indicates the best performing model within each block.}
\label{result-attack}
\end{table}

As baselines for reference, we use three strong defense models, namely GNNGuard \citep{ZhangZ20}, GNN-Jaccard \citep{Wu0TDLZ19} and GNN-SVD \citep{EntezariADP20}. Results on Cora and Citeseer data are reported in Table \ref{result-attack}.  As in Table \ref{table:heterophily-results}, we again observe that UGNN with $\rho$ outperforms the IGNN models as well as UGNN without $\rho$.  And somewhat surprisingly, UGNN with $\rho$ also performs comparably or better than strong existing models that were meticulously designed to defend against adversarial attacks.  These result further demonstrate the UGNN advantage in terms of robustness relative to presently-available IGNN models.

\vspace{-0.5em}
\section{Scalability Considerations}\label{sec:time-and-memory}

In this section we provide complexity analysis and supporting system efficiency empirical comparisons for full-graph training with UGNN and IGNN models.  We then conclude with a discussion of sampling-based and lazy training methods that can be applied to further speed up UGNN models.

\vspace{-1em}
\subsection{Complexity Analysis}

\paragraph{Time Complexity:} The time complexity of UGNN is $O(md K +Nnd^2)$, where $m$ is the number of edges, $n$ is the number of nodes, $K$ is the number of propagation steps, $N$ is the total number of MLP layers and $d$ is the largest hidden size. As a useful reference point, the time complexity of a standard $K$-layer GCN or GraphSAGE model is $O(md K +Knd^2)$. Moreover, if the UGNN MLP layers are all after propagation (i.e., no parameters before propagation), the time complexity can be reduced to $O(Nnd^2)$ by precomputing the propagation (i.e., the same as an MLP). 

For IGNN (as well as MIGNN), the time complexity depends on the number of iterations needed to drive (\ref{eq:IGNN_iteration}) to an approximate fixed point, which relates to certain properties of the weight matrix $W_p$. However, the official IGNN implementation\footnote{https://github.com/SwiftieH/IGNN} executes propagation steps until either acceptable error rate or an upper threshold is reached. If the number of iteration steps used to approximate the fixed point is $K$, then the IGNN time complexity is $O(mdK + Nnd^2)$, which is essentially the same as UGNN (and other message passing-based GNNs), although generally the value of $K$ will be much higher. That being said, the computational complexity of the IGNN backward pass alone is constant with respect to $K$, although this does not change the overall complexity order.

For EIGNN, based on its linearity, the overall/composite graph propagation matrix can actually be pre-computed in $O(n^3)$ time. In this way, it does not need to be recomputed during training, which reduces the training time complexity by a factor of $O(Nnd^2)$. However, for large-sized graphs often encountered in practice, an $O(n^3)$ computation  can be a prohibitive overhead, which would also need to be recomputed any time a propagation-related hyper-parameter is modified.  


\paragraph{Memory Complexity:}  Absent sampling-based training methods to be discussed further in Section \ref{sec:additional_scaling} below, the memory complexity of UGNN is $O(m d (K + N))$; a comparable $K$-layer GCN or GraphSAGE model would require $O(m d K)$.  Meanwhile the IGNN memory complexity is only $O(md)$, which is constant with respect to $K$ (a key advantage).

\subsection{Empirical Comparisons}

In terms of time and memory consumption in practice, we compare UGNN and IGNN on the Amazon Co-Purchase benchmark, which has been advocated in \citep{GuC0SG20} as a suitable data source for testing IGNN.  Results are shown in Figures \ref{fig:mem} and \ref{fig:time} based on executing 100 steps of training and evaluation on a single Tesla T4 GPU.  Clearly, IGNN maintains a huge advantage in terms of memory consumption, while UGNN has a faster runtime provided the number of propagation steps is not too large.  Of course these results are based on full-graph training, so it is worth considering additional scalability measures as described next.



\begin{minipage}{\linewidth}
\begin{minipage}[t]{0.49\linewidth}
    \centering
    \includegraphics[scale=0.4]{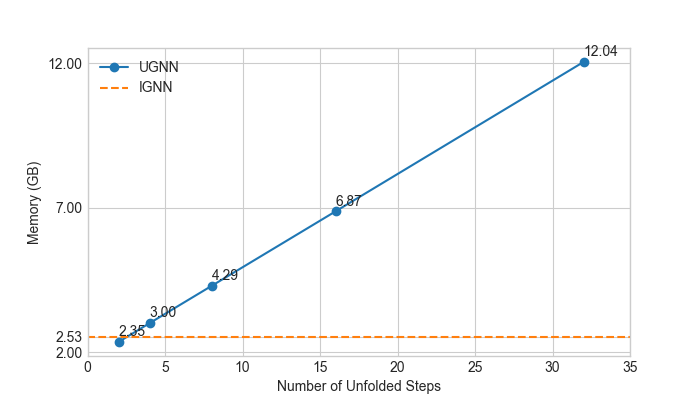}
    \captionof{figure}{Memory cost}
    \label{fig:mem}
\end{minipage}
\begin{minipage}[t]{0.49\linewidth}
    \centering
    \includegraphics[scale=0.4]{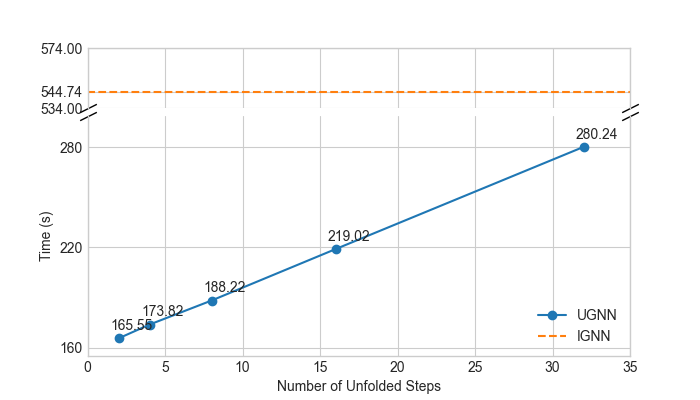}
     \captionof{figure}{Running time}
    \label{fig:time}
\end{minipage}
\end{minipage}

\subsection{Directions for Additional Scalability} \label{sec:additional_scaling}

We are presently unaware of existing work to further scale IGNN models to handle huge real-world graphs, for example, using sampling-based methods \citep{fastgcn,clustergcn,graphsage,pinsage,ladies} or historical embedding caches \citep{vrgcn,gas} as have been applied to typical message-passing GNNs. While in principle such methods may be applicable, the stability of implicit differentiation would need to be maintained, and present IGNN work thus far has largely focused on full graph training. In contrast, there exist two recent orthogonal directions explicitly designed for scaling UGNNs.

First, lazy training methods, combined with sampling and a historical embedding cache, have been developed such that UGNN  complexity is largely decoupled from model depth and far more efficient than full-graph training \citep{lazygnn}.  As a second approach, UGNN models have been combined with offline sampling \citep{shadowgnn,ibmb} such that training can proceed on the very largest publicly-available benchmarks \citep{jiang2023musegnn}.  This includes the 377GB dataset MAG240M from \citep{ogb-lsc} and the 1.15TB dataset IGB-full from \citep{igb}, where UGNN models akin to (\ref{eq:example_prop1}), but integrated with offline sampling, achieve SOTA performance.  As opposed to full-graph training, the memory requirement for this approach reduces to $O(n_s d (K + N))$, where $n_s$ is the size of the offline sampled subgraphs.  A comparable GCN or GraphSAGE model would utilize $O(n_s d K)$.


\vspace{-0.5em}
\section{Conclusions}
This work has closely examined the relationship between IGNNs and UGNNs,  shedding light on their similarities and differences.  In this regard, representative contributions and take-home messages are as follows: 
\begin{itemize}

\item Existing IGNN models broadly coalesce around the propagation update from (\ref{eq:IGNN_iteration}), with notable differences tied to varying choices for $\sigma$, $P$, $W_p$, and $f$.  As for UGNN, we have introduced a flexible foundation via the general energy from (\ref{eq:general_unfolded_objective}) and the attendant descent steps given by (\ref{eq:basic_grad_step}), (\ref{eq:gamma_update}), and (\ref{eq:prox_step}).  Collectively, these subsume a wide variety of existing UGNN models, while also motivating previously-unused variants with convergence guarantees and broader coverage of IGNN capabilities.

\item IGNNs should in principle have an advantage when long-range propagation and full-graph training are dominant factors leading to high accuracy.  However, the practical significance of such an advantage is presently limited for two reasons: (i) Many node-classification benchmarks in common use today do not actually require long-range propagation, and relatively shallow GNNs already perform well; (ii) Even in instances where long-range propagation may be helpful (e.g., certain heterophily graphs, the Amazon Co-Purchase benchmark, etc.), the dataset scales are such that UGNNs can easily be applied as well while maintaining greater flexibility for matching or exceeding the accuracy.  Of course these observations are subject to change as new application scenarios emerge, and IGNNs remain an elegant paradigm to consider for future use cases.

\item We have demonstrated that UGNN symmetric graph propagation weights are essentially unavoidable, while IGNNs have no such restriction.  And yet our complementary analytical and empirical results suggest that these symmetric UGNN  weights are not a significant hindrance relative to IGNNs, at least in terms of model expressiveness and subsequent node classification accuracy.  In contrast, UGNN nonlinear graph propagation operators, as instantiated via a concave $\rho$ and the attention weights from (\ref{eq:gamma_update}), can be advantageous relative to existing IGNNs when unreliable edges are present, e.g., as evidenced by the comparisons in Tables \ref{table:heterophily-results} and \ref{result-attack} involving certain types of heterophily graphs and adversarial attacks respectively.  

\item Regarding heterophily in particular, a by-product of our analysis is the differentiation of two heterophily regimes, characterized by distinct distributions of same-label nodes as quantified by the ASD metric in Figure \ref{fig:avg-dist}.  Within the $\mathrm{ASD}_{\mathrm{sup}}$ regime, both UGNN and IGNN models excel by exploiting long-range dependencies; however, in the $\mathrm{ASD}_{\mathrm{inf}}$ regime only UGNNs with a concave $\rho$ are competitive.

\item IGNN has a major advantage in memory costs for full-graph training, since backpropagation does not require storing intermediate node representations across all propagation steps.  However, UGNNs have an advantage in terms of computational complexity if the number of propagation steps can be truncated without compromising accuracy.  Furthermore,  UGNN memory and computation savings are possible with sampling-based methods or lazy training.

\item UGNN model energy functions in particular provide a natural entry point for inducing versatile GNN architectures that yield predictable results.  Natural candidates for extending beyond (\ref{eq:general_unfolded_objective}) include the handling of heterogeneous graphs using an energy whose minimization mimics knowledge graph embedding computations \citep{ahn2022descent}, or weaving negative samples within an energy function targeting link prediction tasks \citep{wang2024efficient}.  Future work could also consider exploiting UGNN models for the sake of explainability, e.g., disentangling the importance of input features versus network effects in arriving at node-level predictions.



\end{itemize}

\noindent Overall, both models benefit from the inductive biases of their respective design criteria, which frequently overlap but retain important areas of distinction that we have elucidated herein.





\newpage
\bibliography{references}

\begin{thebibliography}{95}
\providecommand{\natexlab}[1]{#1}
\providecommand{\url}[1]{\texttt{#1}}
\expandafter\ifx\csname urlstyle\endcsname\relax
  \providecommand{\doi}[1]{doi: #1}\else
  \providecommand{\doi}{doi: \begingroup \urlstyle{rm}\Url}\fi

\bibitem[Ahn et~al.(2022)Ahn, Yang, Gan, Moon, and Wipf]{ahn2022descent}
Hongjoon Ahn, Yongyi Yang, Quan Gan, Taesup Moon, and David Wipf.
\newblock Descent steps of a relation-aware energy produce heterogeneous graph neural networks.
\newblock \emph{Advances in Neural Information Processing Systems}, 35, 2022.

\bibitem[Amos and Kolter(2017)]{amos2017optnet}
Brandon Amos and J~Zico Kolter.
\newblock Optnet: Differentiable optimization as a layer in neural networks.
\newblock In \emph{International Conference on Machine Learning}, pages 136--145. PMLR, 2017.

\bibitem[Bai et~al.(2019)Bai, Kolter, and Koltun]{bai2019deep}
Shaojie Bai, J~Zico Kolter, and Vladlen Koltun.
\newblock Deep equilibrium models.
\newblock \emph{arXiv preprint arXiv:1909.01377}, 2019.

\bibitem[Baker et~al.(2023)Baker, Wang, Hauck, and Wang]{MIGNN}
Justin Baker, Qingsong Wang, Cory~D. Hauck, and Bao Wang.
\newblock Implicit graph neural networks: {A} monotone operator viewpoint.
\newblock In \emph{International Conference on Machine Learning}, 2023.

\bibitem[Bertsekas(1999)]{Bertsekas1999}
Dimitri Bertsekas.
\newblock \emph{Nonlinear Programming}.
\newblock Athena Scientific, 2nd edition, 1999.

\bibitem[Bo et~al.(2021)Bo, Wang, Shi, and Shen]{fagcn}
Deyu Bo, Xiao Wang, Chuan Shi, and Huawei Shen.
\newblock Beyond low-frequency information in graph convolutional networks.
\newblock In \emph{Proceedings of the AAAI Conference on Artificial Intelligence}, 2021.

\bibitem[Chen et~al.(2017{\natexlab{a}})Chen, Zhu, and Song]{vrgcn}
Jianfei Chen, Jun Zhu, and Le~Song.
\newblock Stochastic training of graph convolutional networks with variance reduction.
\newblock \emph{arXiv preprint arXiv:1710.10568}, 2017{\natexlab{a}}.

\bibitem[Chen et~al.(2018)Chen, Ma, and Xiao]{fastgcn}
Jie Chen, Tengfei Ma, and Cao Xiao.
\newblock {FastGCN}: fast learning with graph convolutional networks via importance sampling.
\newblock \emph{arXiv preprint arXiv:1801.10247}, 2018.

\bibitem[Chen et~al.(2020)Chen, Wei, Huang, Ding, and Li]{DBLP:conf/icml/ChenWHDL20}
Ming Chen, Zhewei Wei, Zengfeng Huang, Bolin Ding, and Yaliang Li.
\newblock Simple and deep graph convolutional networks.
\newblock In \emph{Proceedings of the 37th International Conference on Machine Learning, {ICML}}, volume 119, pages 1725--1735, 2020.

\bibitem[Chen and Eldar(2021)]{chen2021graph}
Siheng Chen and Yonina~C Eldar.
\newblock Graph signal denoising via unrolling networks.
\newblock In \emph{ICASSP 2021-2021 IEEE International Conference on Acoustics, Speech and Signal Processing (ICASSP)}, pages 5290--5294, 2021.

\bibitem[Chen et~al.(2017{\natexlab{b}})Chen, Ge, Wang, Wang, Ye, and Yin]{chen2017strong}
Yichen Chen, Dongdong Ge, Mengdi Wang, Zizhuo Wang, Yinyu Ye, and Hao Yin.
\newblock Strong {NP}-hardness for sparse optimization with concave penalty functions.
\newblock In \emph{International Confernece on Machine Learning}, 2017{\natexlab{b}}.

\bibitem[Chiang et~al.(2019)Chiang, Liu, Si, Li, Bengio, and Hsieh]{clustergcn}
Wei-Lin Chiang, Xuanqing Liu, Si~Si, Yang Li, Samy Bengio, and Cho-Jui Hsieh.
\newblock {Cluster-GCN}: An efficient algorithm for training deep and large graph convolutional networks.
\newblock In \emph{Proceedings of the 25th ACM SIGKDD international conference on knowledge discovery \& data mining}, pages 257--266, 2019.

\bibitem[Chien et~al.(2020)Chien, Peng, Li, and Milenkovic]{gprgcn}
Eli Chien, Jianhao Peng, Pan Li, and Olgica Milenkovic.
\newblock Adaptive universal generalized pagerank graph neural network.
\newblock \emph{arXiv preprint arXiv:2006.07988}, 2020.

\bibitem[Combettes and Pesquet(2011)]{combettes2011proximal}
Patrick~L Combettes and Jean-Christophe Pesquet.
\newblock Proximal splitting methods in signal processing.
\newblock In \emph{Fixed-point algorithms for inverse problems in science and engineering}, pages 185--212. Springer, 2011.

\bibitem[Dai et~al.(2018)Dai, Kozareva, Dai, Smola, and Song]{DaiKDSS18}
Hanjun Dai, Zornitsa Kozareva, Bo~Dai, Alex Smola, and Le~Song.
\newblock Learning steady-states of iterative algorithms over graphs.
\newblock In \emph{International Conference on Machine Learning}, pages 1106--1114, 2018.

\bibitem[Di~Giovanni et~al.(2023)Di~Giovanni, Rowbottom, Chamberlain, Markovich, and Bronstein]{di2023understanding}
Francesco Di~Giovanni, James Rowbottom, Benjamin~Paul Chamberlain, Thomas Markovich, and Michael~M Bronstein.
\newblock Understanding convolution on graphs via energies.
\newblock \emph{Transactions on Machine Learning Research}, 2023.

\bibitem[Du et~al.(2022)Du, Shi, Fu, Ma, Liu, Han, and Zhang]{gbkgnn}
Lun Du, Xiaozhou Shi, Qiang Fu, Xiaojun Ma, Hengyu Liu, Shi Han, and Dongmei Zhang.
\newblock Gbk-gnn: Gated bi-kernel graph neural networks for modeling both homophily and heterophily.
\newblock In \emph{Proceedings of the ACM Web Conference 2022}, pages 1550--1558, 2022.

\bibitem[Dwivedi et~al.(2023)Dwivedi, Joshi, Luu, Laurent, Bengio, and Bresson]{dwivedi2023benchmarking}
Vijay~Prakash Dwivedi, Chaitanya~K Joshi, Anh~Tuan Luu, Thomas Laurent, Yoshua Bengio, and Xavier Bresson.
\newblock Benchmarking graph neural networks.
\newblock \emph{Journal of Machine Learning Research}, 24\penalty0 (43):\penalty0 1--48, 2023.

\bibitem[El~Ghaoui et~al.(2020)El~Ghaoui, Gu, Travacca, Askari, and Tsai]{el2020implicit}
Laurent El~Ghaoui, Fangda Gu, Bertrand Travacca, Armin Askari, and Alicia Tsai.
\newblock Implicit deep learning.
\newblock \emph{ar{X}iv preprint ar{X}iv:1908.06315}, 2020.

\bibitem[Entezari et~al.(2020)Entezari, Al-Sayouri, Darvishzadeh, and Papalexakis]{EntezariADP20}
Negin Entezari, Saba~A Al-Sayouri, Amirali Darvishzadeh, and Evangelos~E Papalexakis.
\newblock All you need is low (rank) defending against adversarial attacks on graphs.
\newblock In \emph{Proceedings of the 13th International Conference on Web Search and Data Mining}, pages 169--177, 2020.

\bibitem[Fan and Li(2001)]{fan2001variable}
Jianqing Fan and Runze Li.
\newblock Variable selection via nonconcave penalized likelihood and its oracle properties.
\newblock \emph{JASTA}, 96\penalty0 (456):\penalty0 1348--1360, 2001.

\bibitem[Fey et~al.(2021)Fey, Lenssen, Weichert, and Leskovec]{gas}
Matthias Fey, Jan~E Lenssen, Frank Weichert, and Jure Leskovec.
\newblock {GNNAutoScale}: Scalable and expressive graph neural networks via historical embeddings.
\newblock In \emph{International Conference on Machine Learning}, pages 3294--3304. PMLR, 2021.

\bibitem[Finkelshtein et~al.(2024)Finkelshtein, Huang, Bronstein, and Ceylan]{finkelshtein2023cooperative}
Ben Finkelshtein, Xingyue Huang, Michael Bronstein, and Ismail~Ilkan Ceylan.
\newblock Cooperative graph neural networks.
\newblock \emph{International Conference on Machine Learning}, 2024.

\bibitem[Frecon et~al.(2022)Frecon, Gasso, Pontil, and Salzo]{frecon2022bregman}
Jordan Frecon, Gilles Gasso, Massimiliano Pontil, and Saverio Salzo.
\newblock Bregman neural networks.
\newblock In \emph{International Conference on Machine Learning}, pages 6779--6792. PMLR, 2022.

\bibitem[Fu et~al.(2023)Fu, Dupty, Dong, and Sun]{fu2023implicit}
Guoji Fu, Mohammed~Haroon Dupty, Yanfei Dong, and Lee~Wee Sun.
\newblock Implicit graph neural diffusion based on constrained dirichlet energy minimization.
\newblock \emph{arXiv preprint arXiv:2308.03306}, 2023.

\bibitem[Gallicchio and Micheli(2020)]{gallicchio2020fast}
Claudio Gallicchio and Alessio Micheli.
\newblock Fast and deep graph neural networks.
\newblock In \emph{Proceedings of the AAAI Conference on Artificial Intelligence}, 2020.

\bibitem[Gasteiger et~al.(2019)Gasteiger, Wei{\ss}enberger, and G{\"u}nnemann]{gasteiger2019diffusion}
Johannes Gasteiger, Stefan Wei{\ss}enberger, and Stephan G{\"u}nnemann.
\newblock Diffusion improves graph learning.
\newblock \emph{Advances in neural information processing systems}, 32, 2019.

\bibitem[Gasteiger et~al.(2022)Gasteiger, Qian, and G{\"u}nnemann]{ibmb}
Johannes Gasteiger, Chendi Qian, and Stephan G{\"u}nnemann.
\newblock Influence-based mini-batching for graph neural networks.
\newblock In \emph{Learning on Graphs Conference}, pages 9--1. PMLR, 2022.

\bibitem[Geng et~al.(2021)Geng, Guo, Chen, Li, Wei, and Lin]{hamburger}
Zhengyang Geng, Meng{-}Hao Guo, Hongxu Chen, Xia Li, Ke~Wei, and Zhouchen Lin.
\newblock Is attention better than matrix decomposition?
\newblock In \emph{9th International Conference on Learning Representations, {ICLR} 2021, Virtual Event, Austria, May 3-7, 2021}. OpenReview.net, 2021.

\bibitem[Gregor and LeCun(2010)]{gregor2010learning}
Karol Gregor and Yann LeCun.
\newblock Learning fast approximations of sparse coding.
\newblock In \emph{International Conference on Machine Learning}, 2010.

\bibitem[Gribonval and Nikolova(2020)]{gribonval2020characterization}
R{\'e}mi Gribonval and Mila Nikolova.
\newblock A characterization of proximity operators.
\newblock \emph{Journal of Mathematical Imaging and Vision}, 62\penalty0 (6):\penalty0 773--789, 2020.

\bibitem[Gu et~al.(2020)Gu, Chang, Zhu, Sojoudi, and Ghaoui]{GuC0SG20}
Fangda Gu, Heng Chang, Wenwu Zhu, Somayeh Sojoudi, and Laurent~El Ghaoui.
\newblock Implicit graph neural networks.
\newblock In \emph{Advances in Neural Information Processing Systems}, 2020.

\bibitem[Hamilton et~al.(2017{\natexlab{a}})Hamilton, Ying, and Leskovec]{graphsage}
Will Hamilton, Zhitao Ying, and Jure Leskovec.
\newblock Inductive representation learning on large graphs.
\newblock \emph{Advances in Neural Information Processing Systems}, 30, 2017{\natexlab{a}}.

\bibitem[Hamilton et~al.(2017{\natexlab{b}})Hamilton, Ying, and Leskovec]{HamiltonYL17}
William~L Hamilton, Rex Ying, and Jure Leskovec.
\newblock Inductive representation learning on large graphs.
\newblock In \emph{Proceedings of the 31st International Conference on Neural Information Processing Systems}, pages 1025--1035, 2017{\natexlab{b}}.

\bibitem[He et~al.(2017)He, Xin, Ikehata, and Wipf]{Hao2017sparse}
Hao He, Bo~Xin, Satoshi Ikehata, and David Wipf.
\newblock From bayesian sparsity to gated recurrent nets.
\newblock In \emph{Advances in Neural Information Processing Systems}, 2017.

\bibitem[Hershey et~al.(2014)Hershey, Roux, and Weninger]{hershey2014deep}
John Hershey, Jonathan~Le Roux, and Felix Weninger.
\newblock Deep unfolding: {M}odel-based inspiration of novel deep architectures.
\newblock \emph{ar{X}iv preprint ar{X}iv:1409.2574}, 2014.

\bibitem[Hu et~al.(2020)Hu, Fey, Zitnik, Dong, Ren, Liu, Catasta, and Leskovec]{HuFZDRLCL20}
Weihua Hu, Matthias Fey, Marinka Zitnik, Yuxiao Dong, Hongyu Ren, Bowen Liu, Michele Catasta, and Jure Leskovec.
\newblock Open graph benchmark: Datasets for machine learning on graphs.
\newblock \emph{Advances in neural information processing systems}, 33:\penalty0 22118--22133, 2020.

\bibitem[Hu et~al.(2021)Hu, Fey, Ren, Nakata, Dong, and Leskovec]{ogb-lsc}
Weihua Hu, Matthias Fey, Hongyu Ren, Maho Nakata, Yuxiao Dong, and Jure Leskovec.
\newblock {OGB-LSC}: A large-scale challenge for machine learning on graphs.
\newblock \emph{arXiv:2103.09430}, 2021.

\bibitem[Ioannidis et~al.(2018)Ioannidis, Ma, Nikolakopoulos, Giannakis, and Romero]{ioannidis2018}
Vassilis~N Ioannidis, Meng Ma, Athanasios~N Nikolakopoulos, Georgios~B Giannakis, and Daniel Romero.
\newblock Kernel-based inference of functions on graphs.
\newblock In D.~Comminiello and J.~Principe, editors, \emph{Adaptive Learning Methods for Nonlinear System Modeling}. Elsevier, 2018.

\bibitem[Jiang et~al.(2023)Jiang, Liu, Yan, Cai, Wang, and Wipf]{jiang2023musegnn}
Haitian Jiang, Renjie Liu, Xiao Yan, Zhenkun Cai, Minjie Wang, and David Wipf.
\newblock Muse{GNN}: {I}nterpretable and convergent graph neural network layers at scale.
\newblock \emph{arXiv preprint arXiv:2310.12457}, 2023.

\bibitem[Karhadkar et~al.(2023)Karhadkar, Banerjee, and Mont{\'u}far]{karhadkar2022fosr}
Kedar Karhadkar, Pradeep~Kr Banerjee, and Guido Mont{\'u}far.
\newblock {F}o{SR}: First-order spectral rewiring for addressing oversquashing in gnns.
\newblock \emph{International Conference on Learning Representations}, 2023.

\bibitem[Kearnes et~al.(2016)Kearnes, McCloskey, Berndl, Pande, and Riley]{DBLP:journals/jcamd/KearnesMBPR16}
Steven~M. Kearnes, Kevin McCloskey, Marc Berndl, Vijay~S. Pande, and Patrick Riley.
\newblock Molecular graph convolutions: moving beyond fingerprints.
\newblock \emph{J. Comput. Aided Mol. Des.}, 30\penalty0 (8):\penalty0 595--608, 2016.

\bibitem[Khatua et~al.(2023)Khatua, Mailthody, Taleka, Ma, Song, and Hwu]{igb}
Arpandeep Khatua, Vikram~Sharma Mailthody, Bhagyashree Taleka, Tengfei Ma, Xiang Song, and Wen-mei Hwu.
\newblock Igb: Addressing the gaps in labeling, features, heterogeneity, and size of public graph datasets for deep learning research.
\newblock \emph{arXiv preprint arXiv:2302.13522}, 2023.

\bibitem[Kinderlehrer and Stampacchia(1980)]{kinderlehrer1980introduction}
D.~Kinderlehrer and G.~Stampacchia.
\newblock \emph{An Introduction to Variational Inequalities and Their Applications}.
\newblock Classics in Applied Mathematics. Society for Industrial and Applied Mathematics (SIAM, 3600 Market Street, Floor 6, Philadelphia, PA 19104), 1980.
\newblock ISBN 9780898719451.

\bibitem[Kipf and Welling(2017)]{kipf2017semi}
Thomas Kipf and Max Welling.
\newblock Semi-supervised classification with graph convolutional networks.
\newblock In \emph{International Conference on Learning Representations}, 2017.

\bibitem[Klicpera et~al.(2019{\natexlab{a}})Klicpera, Bojchevski, and G{\"u}nnemann]{klicpera2019predict}
Johannes Klicpera, Aleksandar Bojchevski, and Stephan G{\"u}nnemann.
\newblock Predict then propagate: {G}raph neural networks meet personalized pagerank.
\newblock In \emph{International Conference on Learning Representations}, 2019{\natexlab{a}}.

\bibitem[Klicpera et~al.(2019{\natexlab{b}})Klicpera, Wei{\ss}enberger, and G{\"u}nnemann]{klicpera2019diffusion}
Johannes Klicpera, Stefan Wei{\ss}enberger, and Stephan G{\"u}nnemann.
\newblock Diffusion improves graph learning.
\newblock In \emph{Advances in Neural Information Processing Systems}, 2019{\natexlab{b}}.

\bibitem[Li et~al.(2020{\natexlab{a}})Li, So, and Ma]{li2020understanding}
Jiajin Li, Anthony Man-Cho So, and Wing-Kin Ma.
\newblock Understanding notions of stationarity in nonsmooth optimization: A guided tour of various constructions of subdifferential for nonsmooth functions.
\newblock \emph{IEEE Signal Processing Magazine}, 37\penalty0 (5):\penalty0 18--31, 2020{\natexlab{a}}.

\bibitem[Li et~al.(2018)Li, Han, and Wu]{DBLP:conf/aaai/LiHW18}
Qimai Li, Zhichao Han, and Xiao-Ming Wu.
\newblock Deeper insights into graph convolutional networks for semi-supervised learning.
\newblock In \emph{Proceedings of the AAAI Conference on Artificial Intelligence}, 2018.

\bibitem[Li et~al.(2020{\natexlab{b}})Li, Jin, Xu, and Tang]{abs-2005-06149}
Yaxin Li, Wei Jin, Han Xu, and Jiliang Tang.
\newblock Deeprobust: A pytorch library for adversarial attacks and defenses.
\newblock \emph{arXiv preprint arXiv:2005.06149}, 2020{\natexlab{b}}.

\bibitem[Liu et~al.(2021{\natexlab{a}})Liu, Kawaguchi, Hooi, Wang, and Xiao]{eignn}
Juncheng Liu, Kenji Kawaguchi, Bryan Hooi, Yiwei Wang, and Xiaokui Xiao.
\newblock Eignn: Efficient infinite-depth graph neural networks.
\newblock In \emph{Proceedings of the 31st International Conference on Neural Information Processing Systems}, 2021{\natexlab{a}}.

\bibitem[Liu et~al.(2021{\natexlab{b}})Liu, Jin, Ma, Li, Liu, Wang, Yan, and Tang]{liu2021elastic}
Xiaorui Liu, Wei Jin, Yao Ma, Yaxin Li, Hua Liu, Yiqi Wang, Ming Yan, and Jiliang Tang.
\newblock Elastic graph neural networks.
\newblock In \emph{International Conference on Machine Learning}, 2021{\natexlab{b}}.

\bibitem[Luenberger(1984)]{Luenberger84}
D.G. Luenberger.
\newblock \emph{Linear and Nonlinear Programming}.
\newblock Addison--Wesley, Reading, Massachusetts, second edition, 1984.

\bibitem[Luo et~al.(2022)Luo, Duan, Luo, and Chen]{luo2022unifying}
Yi~Luo, Guiduo Duan, Guangchun Luo, and Aiguo Chen.
\newblock Unifying label-inputted graph neural networks with deep equilibrium models.
\newblock \emph{arXiv preprint arXiv:2211.10629v1}, 2022.

\bibitem[Ma et~al.(2020)Ma, Liu, Zhao, Liu, Tang, and Shah]{ma2020unified}
Yao Ma, Xiaorui Liu, Tong Zhao, Yozen Liu, Jiliang Tang, and Neil Shah.
\newblock A unified view on graph neural networks as graph signal denoising.
\newblock \emph{arXiv preprint arXiv:2010.01777}, 2020.

\bibitem[Maurya et~al.(2022)Maurya, Liu, and Murata]{fsgcn}
Sunil~Kumar Maurya, Xin Liu, and Tsuyoshi Murata.
\newblock Simplifying approach to node classification in graph neural networks.
\newblock \emph{Journal of Computational Science}, 62:\penalty0 101695, 2022.

\bibitem[Nesterov(2003)]{nesterov2003introductory}
Yurii Nesterov.
\newblock \emph{Introductory lectures on convex optimization: A basic course}, volume~87.
\newblock Springer Science \& Business Media, 2003.

\bibitem[Oono and Suzuki(2020)]{DBLP:conf/iclr/OonoS20}
Kenta Oono and Taiji Suzuki.
\newblock Graph neural networks exponentially lose expressive power for node classification.
\newblock In \emph{8th International Conference on Learning Representations, {ICLR}}, 2020.

\bibitem[Pan et~al.(2021)Pan, Song, and Huang]{pan2021a}
Xuran Pan, Shiji Song, and Gao Huang.
\newblock A unified framework for convolution-based graph neural networks, 2021.
\newblock URL \url{https://openreview.net/forum?id=zUMD--Fb9Bt}.

\bibitem[Pei et~al.(2019)Pei, Wei, Chang, Lei, and Yang]{PeiWCLY20}
Hongbin Pei, Bingzhe Wei, Kevin Chen-Chuan Chang, Yu~Lei, and Bo~Yang.
\newblock Geom-gcn: Geometric graph convolutional networks.
\newblock In \emph{International Conference on Learning Representations}, 2019.

\bibitem[Platonov et~al.(2023)Platonov, Kuznedelev, Diskin, Babenko, and Prokhorenkova]{new-hetero}
Oleg Platonov, Denis Kuznedelev, Michael Diskin, Artem Babenko, and Liudmila Prokhorenkova.
\newblock A critical look at the evaluation of gnns under heterophily: Are we really making progress?
\newblock In \emph{International Conference on Learning Representations}, 2023.

\bibitem[Rong et~al.(2020)Rong, Huang, Xu, and Huang]{DBLP:conf/iclr/RongHXH20}
Yu~Rong, Wenbing Huang, Tingyang Xu, and Junzhou Huang.
\newblock Dropedge: Towards deep graph convolutional networks on node classification.
\newblock In \emph{8th International Conference on Learning Representations, {ICLR}}, 2020.

\bibitem[Sprechmann et~al.(2015)Sprechmann, Bronstein, and Sapiro]{Sprechmann15}
Pablo Sprechmann, Alex Bronstein, and Guillermo Sapiro.
\newblock Learning efficient sparse and low rank models.
\newblock \emph{IEEE Trans. Pattern Analysis and Machine Intelligence}, 37\penalty0 (9), 2015.

\bibitem[Sriperumbudur and Lanckriet(2009)]{sriperumbudur2009convergence}
Bharath Sriperumbudur and Gert Lanckriet.
\newblock On the convergence of the concave-convex procedure.
\newblock In \emph{Advances in Neural Information Processing Systems}, 2009.

\bibitem[T{\"o}nshoff et~al.(2023)T{\"o}nshoff, Ritzert, Wolf, and Grohe]{tonshoff2021walking}
Jan T{\"o}nshoff, Martin Ritzert, Hinrikus Wolf, and Martin Grohe.
\newblock Walking out of the {W}eisfeiler {L}eman hierarchy: Graph learning beyond message passing.
\newblock \emph{Transactions on Machine Learning Research}, 2023.

\bibitem[Topping et~al.(2022)Topping, Di~Giovanni, Chamberlain, Dong, and Bronstein]{topping2021understanding}
Jake Topping, Francesco Di~Giovanni, Benjamin~Paul Chamberlain, Xiaowen Dong, and Michael Bronstein.
\newblock Understanding over-squashing and bottlenecks on graphs via curvature.
\newblock \emph{International Conference on Learning Representations}, 2022.

\bibitem[Velickovic et~al.(2018)Velickovic, Cucurull, Casanova, Romero, Li{\`{o}}, and Bengio]{VelickovicCCRLB18}
Petar Velickovic, Guillem Cucurull, Arantxa Casanova, Adriana Romero, Pietro Li{\`{o}}, and Yoshua Bengio.
\newblock Graph attention networks.
\newblock In \emph{6th International Conference on Learning Representations, {ICLR}}, 2018.

\bibitem[Veli{\v{c}}kovi{\'c} et~al.(2020)Veli{\v{c}}kovi{\'c}, Ying, Padovano, Hadsell, and Blundell]{velivckovic2019neural}
Petar Veli{\v{c}}kovi{\'c}, Rex Ying, Matilde Padovano, Raia Hadsell, and Charles Blundell.
\newblock Neural execution of graph algorithms.
\newblock In \emph{International Conference on Learning Representations}, 2020.

\bibitem[Von~Luxburg(2007)]{von2007tutorial}
Ulrike Von~Luxburg.
\newblock A tutorial on spectral clustering.
\newblock \emph{Statistics and computing}, 17\penalty0 (4):\penalty0 395--416, 2007.

\bibitem[Wang et~al.(2019)Wang, Zheng, Ye, Gan, Li, Song, Zhou, Ma, Yu, Gai, Xiao, He, Karypis, Li, and Zhang]{wang2019dgl}
Minjie Wang, Da~Zheng, Zihao Ye, Quan Gan, Mufei Li, Xiang Song, Jinjing Zhou, Chao Ma, Lingfan Yu, Yu~Gai, Tianjun Xiao, Tong He, George Karypis, Jinyang Li, and Zheng Zhang.
\newblock Deep graph library: A graph-centric, highly-performant package for graph neural networks.
\newblock \emph{arXiv preprint arXiv:1909.01315}, 2019.

\bibitem[Wang and Zhang(2022)]{jacobgnn}
Xiyuan Wang and Muhan Zhang.
\newblock How powerful are spectral graph neural networks.
\newblock In \emph{International Conference on Machine Learning}, pages 23341--23362. PMLR, 2022.

\bibitem[Wang et~al.(2024)Wang, Hu, Gan, Huang, Qiu, and Wipf]{wang2024efficient}
Yuxin Wang, Xiannian Hu, Quan Gan, Xuanjing Huang, Xipeng Qiu, and David Wipf.
\newblock Efficient link prediction via gnn layers induced by negative sampling.
\newblock \emph{IEEE Transactions on Knowledge and Data Engineering}, 2024.

\bibitem[Wang et~al.(2016)Wang, Ling, and Huang]{wang2016learning}
Zhangyang Wang, Qing Ling, and Thomas Huang.
\newblock Learning deep $\ell_0$ encoders.
\newblock In \emph{AAAI Conference on Artificial Intelligence}, 2016.

\bibitem[West(1984)]{west1984outlier}
Mike West.
\newblock Outlier models and prior distributions in {B}ayesian linear regression.
\newblock \emph{J. Royal Statistical Society: Series B}, 46\penalty0 (3), 1984.

\bibitem[Wu et~al.(2019)Wu, Wang, Tyshetskiy, Docherty, Lu, and Zhu]{Wu0TDLZ19}
Huijun Wu, Chen Wang, Yuriy Tyshetskiy, Andrew Docherty, Kai Lu, and Liming Zhu.
\newblock Adversarial examples for graph data: Deep insights into attack and defense.
\newblock In \emph{Proceedings of the Twenty-Eighth International Joint Conference on Artificial Intelligence, {IJCAI}}, pages 4816--4823, 2019.

\bibitem[Wu et~al.(2023)Wu, Zhao, Yang, Zhang, Nie, Jiang, Bian, and Yan]{wu2024simplifying}
Qitian Wu, Wentao Zhao, Chenxiao Yang, Hengrui Zhang, Fan Nie, Haitian Jiang, Yatao Bian, and Junchi Yan.
\newblock {SGF}ormer: {S}implifying and empowering transformers for large-graph representations.
\newblock \emph{Advances in Neural Information Processing Systems}, 37, 2023.

\bibitem[Wu et~al.(2020)Wu, Pan, Chen, Long, Zhang, and Philip]{wu2020comprehensive}
Zonghan Wu, Shirui Pan, Fengwen Chen, Guodong Long, Chengqi Zhang, and S~Yu Philip.
\newblock A comprehensive survey on graph neural networks.
\newblock \emph{IEEE transactions on neural networks and learning systems}, 32\penalty0 (1):\penalty0 4--24, 2020.

\bibitem[Xie et~al.(2023)Xie, Wang, Ling, Li, Liu, and Lin]{xie2023optimization}
Xingyu Xie, Qiuhao Wang, Zenan Ling, Xia Li, Guangcan Liu, and Zhouchen Lin.
\newblock Optimization induced equilibrium networks: An explicit optimization perspective for understanding equilibrium models.
\newblock \emph{IEEE Transactions on Pattern Analysis and Machine Intelligence}, 45\penalty0 (3):\penalty0 3604--3616, 2023.

\bibitem[Xue et~al.(2023)Xue, Han, Torkamani, Pei, and Liu]{lazygnn}
Rui Xue, Haoyu Han, MohamadAli Torkamani, Jian Pei, and Xiaorui Liu.
\newblock Lazy{GNN}: {L}arge-scale graph neural networks via lazy propagation.
\newblock In \emph{International Conference on Machine Learning}, 2023.

\bibitem[Yang et~al.(2021)Yang, Liu, Wang, Zhou, Gan, Wei, Zhang, Huang, and Wipf]{yang2021}
Yongyi Yang, Tang Liu, Yangkun Wang, Jinjing Zhou, Quan Gan, Zhewei Wei, Zheng Zhang, Zengfeng Huang, and David Wipf.
\newblock Graph neural networks inspired by classical iterative algorithms.
\newblock In \emph{International Conference on Machine Learning}, 2021.

\bibitem[Yang et~al.(2022)Yang, , Huang, and Wipf]{yang2022transformers}
Yongyi Yang, , Zengfeng Huang, and David Wipf.
\newblock Transformers from an optimization perspective.
\newblock In \emph{Advances in Neural Information Processing Systems}, volume~35, 2022.

\bibitem[Ying et~al.(2018)Ying, He, Chen, Eksombatchai, Hamilton, and Leskovec]{pinsage}
Rex Ying, Ruining He, Kaifeng Chen, Pong Eksombatchai, William~L Hamilton, and Jure Leskovec.
\newblock Graph convolutional neural networks for web-scale recommender systems.
\newblock In \emph{Proceedings of the 24th ACM SIGKDD international conference on knowledge discovery \& data mining}, pages 974--983, 2018.

\bibitem[Zeng et~al.(2021)Zeng, Zhang, Xia, Srivastava, Malevich, Kannan, Prasanna, Jin, and Chen]{shadowgnn}
Hanqing Zeng, Muhan Zhang, Yinglong Xia, Ajitesh Srivastava, Andrey Malevich, Rajgopal Kannan, Viktor Prasanna, Long Jin, and Ren Chen.
\newblock Decoupling the depth and scope of graph neural networks.
\newblock \emph{Advances in Neural Information Processing Systems}, 34, 2021.

\bibitem[Zhang et~al.(2020)Zhang, Yan, Xie, Xia, and Zhang]{thatpaper}
Hongwei Zhang, Tijin Yan, Zenjun Xie, Yuanqing Xia, and Yuan Zhang.
\newblock Revisiting graph convolutional network on semi-supervised node classification from an optimization perspective.
\newblock \emph{CoRR}, abs/2009.11469, 2020.

\bibitem[Zhang and Zitnik(2020)]{ZhangZ20}
Xiang Zhang and Marinka Zitnik.
\newblock Gnnguard: Defending graph neural networks against adversarial attacks.
\newblock \emph{Advances in Neural Information Processing Systems}, 33, 2020.

\bibitem[Zheng et~al.(2021)Zheng, Zhou, Gao, Wang, Lio, Li, and Mont{\'u}far]{zheng2021framelets}
Xuebin Zheng, Bingxin Zhou, Junbin Gao, Yu~Guang Wang, Pietro Lio, Ming Li, and Guido Mont{\'u}far.
\newblock How framelets enhance graph neural networks.
\newblock In \emph{International Conference on Machine Learning}, 2021.

\bibitem[Zhou et~al.(2004)Zhou, Bousquet, Lal, Weston, and Sch{\"o}lkopf]{zhou2004learning}
Dengyong Zhou, Olivier Bousquet, Thomas~Navin Lal, Jason Weston, and Bernhard Sch{\"o}lkopf.
\newblock Learning with local and global consistency.
\newblock \emph{Advances in Neural Information Processing Systems}, 2004.

\bibitem[Zhou et~al.(2018)Zhou, Cui, Zhang, Yang, Liu, Wang, Li, and Sun]{DBLP:journals/corr/abs-1812-08434}
Jie Zhou, Ganqu Cui, Zhengyan Zhang, Cheng Yang, Zhiyuan Liu, Lifeng Wang, Changcheng Li, and Maosong Sun.
\newblock Graph neural networks: A review of methods and applications.
\newblock \emph{arXiv preprint arXiv:1812.08434}, 2018.

\bibitem[Zhu et~al.(2020{\natexlab{a}})Zhu, Rossi, Rao, Mai, Lipka, Ahmed, and Koutra]{DBLP:journals/corr/abs-2009-13566}
Jiong Zhu, Ryan~A Rossi, Anup Rao, Tung Mai, Nedim Lipka, Nesreen~K Ahmed, and Danai Koutra.
\newblock Graph neural networks with heterophily.
\newblock \emph{arXiv preprint arXiv:2009.13566}, 2020{\natexlab{a}}.

\bibitem[Zhu et~al.(2020{\natexlab{b}})Zhu, Yan, Zhao, Heimann, Akoglu, and Koutra]{DBLP:conf/nips/ZhuYZHAK20}
Jiong Zhu, Yujun Yan, Lingxiao Zhao, Mark Heimann, Leman Akoglu, and Danai Koutra.
\newblock Beyond homophily in graph neural networks: Current limitations and effective designs.
\newblock In \emph{Advances in Neural Information Processing Systems, NeurIPS}, 2020{\natexlab{b}}.

\bibitem[Zhu et~al.(2020{\natexlab{c}})Zhu, Yan, Zhao, Heimann, Akoglu, and Koutra]{h2gcn}
Jiong Zhu, Yujun Yan, Lingxiao Zhao, Mark Heimann, Leman Akoglu, and Danai Koutra.
\newblock Beyond homophily in graph neural networks: Current limitations and effective designs.
\newblock \emph{Advances in neural information processing systems}, 33:\penalty0 7793--7804, 2020{\natexlab{c}}.

\bibitem[Zhu et~al.(2021)Zhu, Wang, Shi, Ji, and Cui]{zhu2021interpreting}
Meiqi Zhu, Xiao Wang, Chuan Shi, Houye Ji, and Peng Cui.
\newblock Interpreting and unifying graph neural networks with an optimization framework.
\newblock \emph{arXiv preprint arXiv:2101.11859}, 2021.

\bibitem[Zou et~al.(2019)Zou, Hu, Wang, Jiang, Sun, and Gu]{ladies}
Difan Zou, Ziniu Hu, Yewen Wang, Song Jiang, Yizhou Sun, and Quanquan Gu.
\newblock Layer-dependent importance sampling for training deep and large graph convolutional networks.
\newblock \emph{Advances in Neural Information Processing Systems}, 32, 2019.

\bibitem[Z{\"{u}}gner and G{\"{u}}nnemann(2019)]{DBLP:conf/iclr/ZugnerG19}
Daniel Z{\"{u}}gner and Stephan G{\"{u}}nnemann.
\newblock Adversarial attacks on graph neural networks via meta learning.
\newblock In \emph{7th International Conference on Learning Representations, {ICLR}}, 2019.

\bibitem[Z{\"{u}}gner et~al.(2019)Z{\"{u}}gner, Akbarnejad, and G{\"{u}}nnemann]{DBLP:conf/ijcai/ZugnerAG19}
Daniel Z{\"{u}}gner, Amir Akbarnejad, and Stephan G{\"{u}}nnemann.
\newblock Adversarial attacks on neural networks for graph data.
\newblock In \emph{Proceedings of the Twenty-Eighth International Joint Conference on Artificial Intelligence, {IJCAI}}, 2019.

\end{thebibliography}

\newpage 
\appendix

\section{Datasets, Experiment Details, and Model Specifications}

In this section we introduce additional experiment details as well as give a comprehensive introduction of the UGNN architectures used in the paper.

\subsection{Datasets}


\subtitle{Practical Impact of Symmetric Weights}~ We choose the popular ogbn-arxiv dataset \citep{HuFZDRLCL20} for the experiments in Section \ref{sec:label-recovering}.  Original training splits for generating subsequent synthetic labels follow the standard OGB protocol.

\subtitle{Long-Range Dependency/Sparse Label Tests}~ In Section \ref{sec:long-range-result}, we adopt the Amazon Co-Purchase dataset and Chains dataset, both of which have previously been used in \citep{GuC0SG20}  for evaluating performance involving long-range dependencies.  We use the Amazon Co-Purchase dataset provided by the IGNN repo \citep{GuC0SG20}, including the data-processing and evaluation code, in order to obtain a fair comparison. As for splitting, 10\% of nodes are selected as the test set.  Because there is no dev set, we directly report the test result of the last epoch. We also vary the fraction of training nodes from 5\% to 9\%. Additionally, because there are no node features, we learn a 128-dim feature vector for each node. All of these settings from above follow from \citep{GuC0SG20}. The Chains dataset is a synthetic dataset introduced by \citep{GuC0SG20}. We use exactly the same setting as in \citep{GuC0SG20}.

\subtitle{Heterophily Experiments}~ In Section \ref{sec:hetero-result}, we use several heterophily datasets, including Texas, Wisconsin, Actor and Cornell datasets which are introduced by \citep{PeiWCLY20}, and Roman-empire, Amazon-ratings, Minesweeper, Tolokers and Questions which are introduced by \citep{new-hetero}. We used the data split, processed node features, and labels provided by \citep{PeiWCLY20} and \citep{new-hetero} respectively.

\subtitle{Adversarial Attack Experiments}~ As mentioned in Section \ref{sec:adversarial-attack-results}, we tested on Cora and Citerseer using Mettack. We use the DeepRobust library \citep{abs-2005-06149} and apply the exact same non-targeted attack setting as in \citep{ZhangZ20}. For all the baseline results in Table \ref{result-attack}, we run the implementation in the DeepRobust library or the GNNGuard official code. Note the GCN-Jaccard results differ slightly from those reported in \citep{ZhangZ20}, likely because of updates in the DeepRobust library and the fact that \citep{ZhangZ20} only report results from a single trial (as opposed to averaged results across multiple trails as we report).



\subtitle{Summary Statistics}~ Table \ref{dataset-info} summarizes the attributes of each dataset listed above.

\begin{table}[htbp] \caption{Dataset statistics. The \textit{FEATURES} column describes the dimensionality of node features. The \textit{Amazon Co-P} row denotes the Amazon Co-Purchase dataset. Note that the Amazon Co-Purchase dataset has no node features.}
\label{dataset-info}
\begin{center} \begin{sc}
\begin{tabular}{lcccc}
\toprule
\textbf{Dataset} & \textbf{Nodes} & \textbf{Edges} & \textbf{Features} & \textbf{Classes}
\\ \midrule
    Amazon Co-P & 334,863 & 2,186,607 & - & 58 \\\midrule
    Cora & 2,708 & 5,429 & 1,433 & 7 \\
    Citeseer & 3,327 & 4,732 & 3,703 & 6 \\
    ogbn-arxiv & 169,343 & 1,166,243 & 128 & 40 \\ \midrule
    Texas & 183 & 309 & 1,703 & 5 \\
    Wisconsin & 251 & 499 & 1,703 & 5 \\
    Actor & 7,600 & 33,544 & 931 & 5 \\
    Cornell & 183 & 295 & 1,703 & 5 \\ \midrule
    Roman-Empire       & 22,662    & 32,927    & 300 & 18 \\
    Amazon-Ratings  & 32,927    & 93,050    & 300 & 5 \\
    Minesweeper & 1,000     & 39,402    & 7 & 2 \\
    Tolokers    & 11,758    & 519,000   & 10 & 2 \\ 
    Questions   & 48,921    & 153,540   & 301 & 2 \\ 
\bottomrule
\end{tabular} \end{sc} \end{center}
\end{table}

\subsection{Experiment Details}

In all experiments of IGNN and EIGNN, we optionally set $W_p^s$ to be trainable parameters\footnote{For EIGNN, following the exact formulation in \citep{eignn}, we actually use $W_p^s = \mu W$, where $W$ is trainable and is forced to be symmetric, and $\mu$ is a hyperparamter.} or fix it to $I$. For IGNN, following the original paper, we optionally set $f(X) = \mathrm{MLP}(PX)$ or $\mathrm{MLP}(X)$, while for EIGNN and UGNN we only set $f(X) = \mathrm{MLP}(X)$. 

In label recovering task in Section \ref{sec:label-recovering} and the evaluation of time and memory in Section \ref{sec:time-and-memory}, for the purpose of comparing finite vs infinite number of propagations and symmetric vs asymmetric weights, we ignore irrelevant options such as different choice of $f(X)$, hidden size, and the attention mechanism in UGNN. We always use $f(X) = XW_x$ and $g\left[{\boldsymbol{y}}_i^*(\mathcal W);\theta\right] = W_g {\boldsymbol{y}}_i^*(\mathcal W)$ where $W_g \in \mathbb R^{c\times d}$ is a learned matrix that maps the  propagated node features to the output space. We set the hidden size to $32$ in asymmetric case and $34$ in symmetric case to ensure nearly the same number of parameters (we deliberately set the hidden size to this small to ensure the models do not overfit). For generating models, we first train them using the original labels of the dataset by 500 steps. For UGNN, we fix the number of propagation steps is set to $2$, and adopt (\ref{eq:ugnn_fixed_point_simple}) of UGNN.

When projecting $W_p$ (or $\tilde W$ in UGNN) to the space that admits unique fixed point, in IGNN's original paper it uses $\|\cdot\|_{\mathrm{inf}}$ \citep{GuC0SG20}, which would break the symmetry, based on our result in Section \ref{sec:global-convergence}, in our implementation when the model weight is symmetric we project $\|\cdot\|_2$ instead of $\|\cdot\|_{\mathrm{inf}}$. In EIGNN, we use rescaling as in \citep{eignn} instead of projection.

In the direct comparison between IGNN and UGNN in Section \ref{sec:label-recovering} and the time $\&$ memory comparison in Section \ref{sec:time-and-memory}, we always use normalized adjacency $P = \hat A$ for propagation matrix. However, in other experiments with UGNN, we allow extra flexibility of the propagation matrix.

Finally, for the results in Sections \ref{sec:label-recovering}, \ref{sec:hetero-result} and  \ref{sec:adversarial-attack-results}, we run each experiment for 10 trials to obtain the standard deviation.

\subsection{Basic UGNN Architecture}

The UGNN architecture is composed of the input module $f\left(X;W\right)$, followed by the unfolded propagation layers defined by (\ref{eq:prox_step}), concluding with $g({\boldsymbol{y}};\theta)$. 

\begin{figure}[htbp]
\centering
\includegraphics[width=40mm]{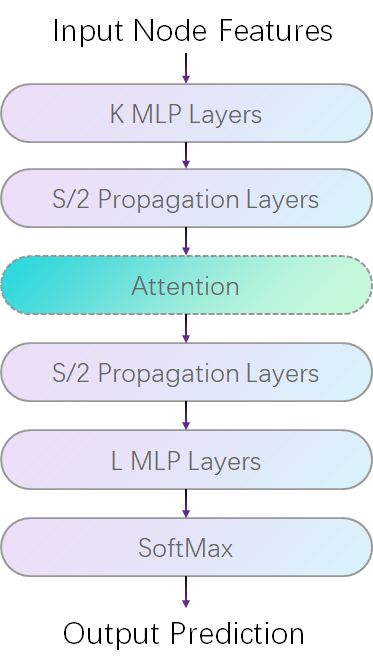}
\captionof{figure}{Model Architecture. $S$ is the total number of propagation steps. $K$ and $L$ are number of MLP layers before and after the propagation respectively. While not a requirement, in all of our experiments, either $K$ or $L$ is set to zero, meaning that the MLP exists on only one side of the propagation layers.}
\label{fig:modelstruct}
\end{figure}

We allow an optional attention layer in UGNN by specifying the $\rho$ function in (\ref{eq:general_unfolded_objective}). See Section \ref{sec:specific-attention-formula} below and \citep{yang2021} for details. The aggregate design of UGNN is depicted in in Figure \ref{fig:modelstruct}.  For simplicity, we generally adopt a single attention layer sandwiched between equal numbers of propagation layers; however, for heterophily datasets we apply an extra attention layer before propagation. Note that the attention only involves reweighting the edge weights of the graph (i.e., it does not alter the node embeddings at each layer).  



\subsection{UGNN Variations}

\subtitle{Pre-conditioning and Reparameterization} \label{sec:gcn_connection_supp}~  
If we unfold the iteration step of (\ref{eq:basic_twirls_objective}), define the reparameterized embeddings $Z = \tilde D^{1/2}Y$ and left multiply the update rule by $\tilde D^{1/2}$, we have
{\small\begin{align}Z^{(k+1)} & = (1-\alpha)Z^{(k)} + \alpha\lambda\tilde D^{-1/2}AY^{(k)} + \alpha \tilde D^{-1/2}f(X;W) \nonumber
\\ & = (1-\alpha)Z^{(k)} + \alpha\lambda\tilde D^{-1/2}A\tilde D^{-1/2}Z^{(k)} + \alpha \tilde D^{-1}Z^{(0)}. \label{eq-prop-1}
\end{align}} From here, if we choose $\alpha = \lambda = 1$, for $Z^{(1)}$ we have that 
\begin{eqnarray}
Z^{(1)} & = & \left(\tilde D^{-1/2}A\tilde D^{-1/2} + \tilde D^{-1}\right)Z^{(0)} \nonumber \\
& = & \tilde D^{-1/2}\tilde A\tilde D^{-1/2}Z^{(0)},
\end{eqnarray}
which gives the exact single-layer GCN formulation in $Z$-space with $Z^{(0)} = f(X;W)$.
\\

\subtitle{Normalized Laplacian Unfolding}~  From another perspective, if we replace $L$ in (\ref{eq:basic_twirls_objective}) with a normalized graph Laplacian, and then take gradients steps as before, there is no need to do preconditioning and reparameterizing. For example, unfolding (\ref{eq:basic_twirls_objective}) with $L$ changed to the symmetrically-normalized version $\tilde L = I - \tilde D^{-1/2}\tilde A\tilde D^{-1/2}$, we get{\small\begin{equation}Y^{(k+1)} = (1-\alpha-\alpha\lambda)Y^{(k)} + \alpha\lambda \tilde D^{-1/2}\tilde A \tilde D^{-1/2}Y^{(k)} + \alpha Y^{(0)},\label{eq-prop-2}
\end{equation}}
where we set $\tilde D = I+D$. This formula is essentially the same as (\ref{eq-prop-1}). The main difference is that there is no $\tilde D^{-1}$ in front of $X$, which indicates an emphasis on the initial features. We found this version to be helpful on ogbn-arxiv and Amazon Co-Purchase data.  Note however that all of our theoretical support from the main paper applies equally well to this normalized version, just with a redefinition of the gradient steps to include the normalized Laplacian.

\subsection{Specific Attention Formula}\label{sec:specific-attention-formula}

\begin{figure}[htbp]
\begin{center}
\begin{minipage}[ht]{0.49\linewidth}
\begin{center}
\begin{small}
\begin{sc}
\centering
\includegraphics[width=60mm]{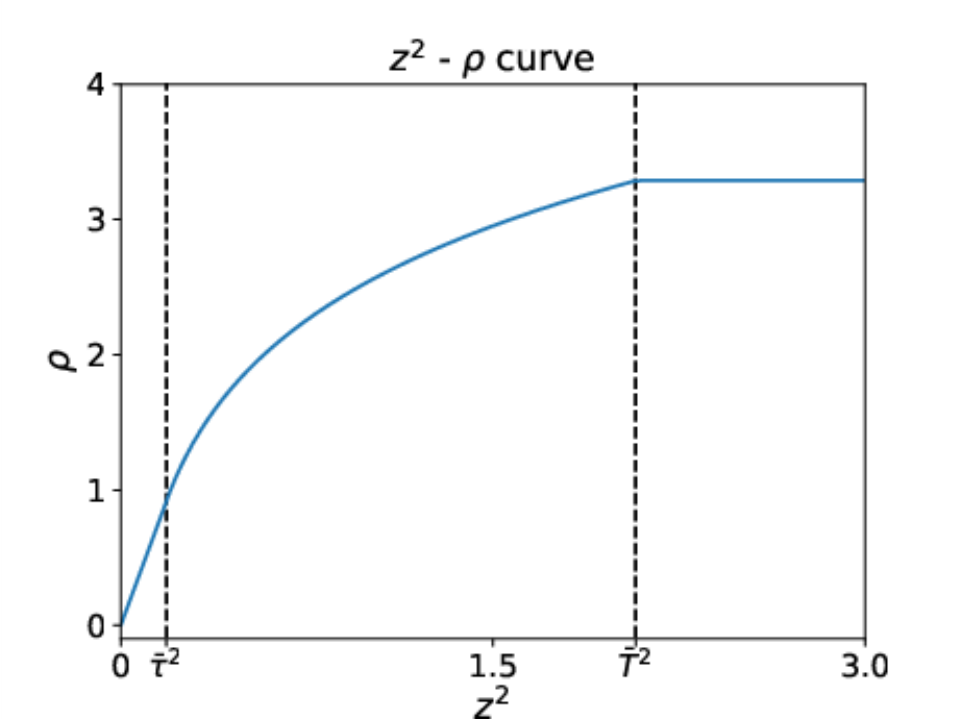}
\end{sc}
\end{small}
\end{center}
\end{minipage}
\begin{minipage}[ht]{0.49\linewidth}
\begin{center}
\begin{small}
\begin{sc}
\centering
\includegraphics[width=60mm]{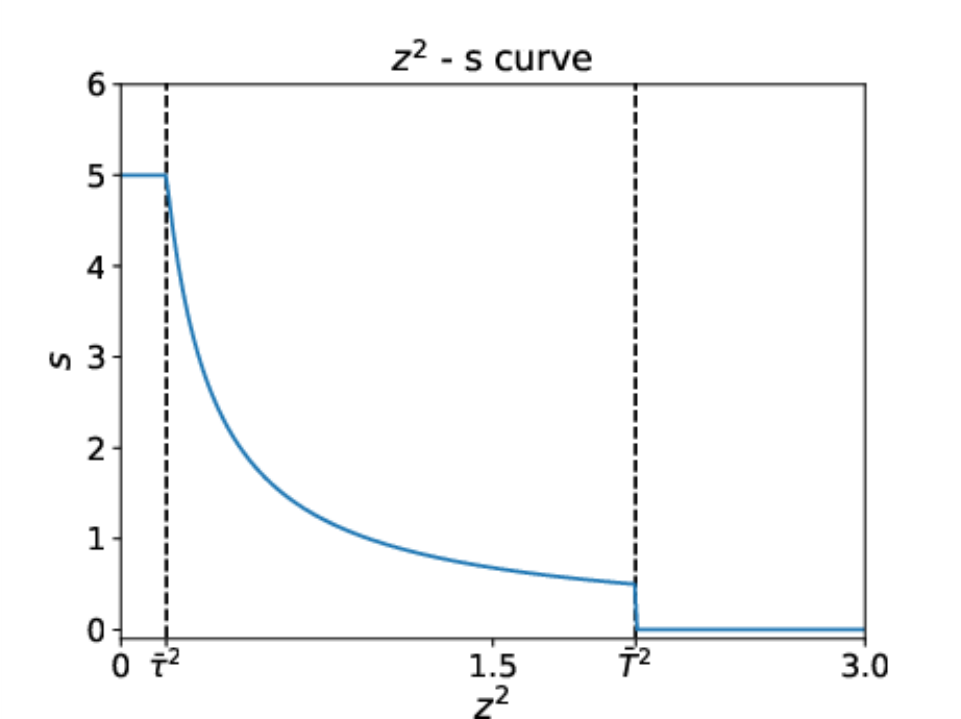}
\end{sc}
\end{small}
\end{center}
\end{minipage}

\captionof{figure}{A visualization of attention functions.}
\label{fig:attn}
\end{center}
\end{figure}

We adapt the TWIRLS model from \citep{yang2021} as the basis for UGNNs with an optional attention layer that is determined by the $\rho$ function in (\ref{eq:general_unfolded_objective}). While the attention mechanism can in principle adopt any concave, non-decreasing function $\rho$, in this work we restrict $\rho$ to a single functional form that is sufficiently flexible to effectively accommodate all experimental scenarios.  Specifically, we adopt
\begin{equation}
\rho(z^2) = \begin{cases}\bar \tau^{p-2}z^2 &\quad\text{if }z<\bar \tau \\ \tfrac{2}{p}\bar T^{p} - \rho_0 &\quad \text{if } z>\bar T\\ \tfrac{2}{p} z^{p} - \rho_0 &\quad\text{otherwise},\end{cases}\end{equation} 
where $p$, $\bar T$, and $\bar \tau$ are non-negative hyperparameters and $\rho_0 = \frac{2-p}{p}\bar \tau^p$ is a constant that ensures $\rho$ is continuous. Additionally, the gradient of $\rho$ produces the attention score function (akin to $\gamma$ in the main paper) given by
\begin{equation} s(z^2) \triangleq \frac{\partial p(z^2)}{\partial z^2}  = \begin{cases}\bar \tau^{p-2} &\quad\text{if }z<\bar \tau \\ 0 &\quad \text{if } z>\bar T\\ z^{p-2} &\quad\text{otherwise}.\end{cases}
\end{equation} 
And for convenience and visualization, we also adopt the reparameterizations $\tau = \bar \tau^{\frac{1}{2-p}}$ and $T = \bar T^\frac{1}{2-p}$, and plot $\rho(z^2)$ and $s(z^2)$ in Figure \ref{fig:attn} using $p=0.1$, $\tau=0.2$, $T=2$.                     




Overall, this flexible choice has a natural interpretation in terms of its differing behavior between the intervals $[0,\bar{\tau}]$, $(\bar{\tau},\bar{T})$, and $(\bar{T},\infty)$.  For example, in the $[0,\bar{\tau}]$ interval a quadratic penalty is applied, which leads to constant attention independent of $z$.  This is exactly like the simple case where there is no attention.  In contrast, within the $(\bar{T},\infty)$ interval $\rho$ is constant and the corresponding attention weight is set to zero (truncation), which is tantamount to edge removal.  And finally, the middle interval provides a natural transition between these two extremes, with $\rho$ becoming increasingly flat with larger $z$ values.

Additionally, many familiar special cases emerge for certain parameter selections.  For example $T=\infty$ corresponds with no explicit truncation, while $p=2$ can instantiate no attention.  And $T=\tau$ means we simply truncate those edges with large distance, effectively keeping the remaining edge attention weights at $1$ (note that there is normalization during propagation, so setting edges to a constant is equivalent to setting them to $1$).

\section{EIGNN as a Special Case of UGNN}

As claimed in the main paper, the EIGNN model can be viewed as a special case of UGNN; we provide the details here.  We begin with the general energy from (\ref{eq:general_unfolded_objective}), and then choose $\phi\left({\boldsymbol{y}};W_\phi\right) = 0$, $\rho\left(z^2\right) = z^2$, and adopt the weight parameterization $W_P^s = s^2F^\top F$, where $F$ is a trainable parameter matrix and $s$ is a rescaling factor given by
\begin{equation}
s = \left(\|F^\top F\|_\mathcal  + \epsilon_F\right)^{-\frac{1}{2}}\text{.}
\end{equation}
This produces the simplified loss
\begin{equation}\label{eq:eignn-energy}
\ell_Y(Y;W_x,f) = \|Y-f(X;W_x)\|_\mathcal F^2 + \mu \mathrm{tr}\left[F^\top Y^\top LYF\right]\text{,}
\end{equation}
where $P=I-L$ is the normalized adjacency matrix. A gradient step on (\ref{eq:eignn-energy}) initialized at $Y^{(k)}$ can then be computed as
\begin{equation}Y^{(k+1)} = \gamma s^2F^\top F Y^{(k)}W_p^s + f\left(X,W_x\right),\end{equation}
which matches the original EIGNN updates from \citep{eignn}.

\section{Notation Used in Proofs}

In following sections, we give proofs of all propositions in the main paper. We first introduce some notations used in proofs here.

For simplicity, in following sections we denote $\ell_Y(Y; {\mathcal W}, f, \rho, \widetilde{B},\phi)$ in (\ref{eq:general_unfolded_objective}) by $\ell_Y(Y)$, denote $\phi\left({\boldsymbol{y}}_{i} ; W_\phi \right)$ by $\phi({\boldsymbol{y}}_i)$ and $ \sum_{i}\phi({\boldsymbol{y}}_i)$ by $\phi(Y)$, and denote $f(X;W)$ by $f(X)$. And the smooth part of $\ell_Y$ is denoted by $\ell_Y^s(Y) = \ell_Y(Y) - \phi(Y)$. And we possibly ignore the parameter of the function (e.g. $\ell_Y$).

We denote the vectorization of a matrix $M$ by $\mathrm{vec} (M)$ and the Kronecker product of two matrices $M_1$ and $M_2$ denoted by $M_1 \otimes M_2$.

For Fr\'{e}chet subdifferential respect to $Y$, we denote it by $\partial_Y$, ignoring the subscript $_{\mathcal F}$ used in main paper.

\section{Proofs Related to UGNN Convergence}

In this section, we investigate different conditions to guarantee convergence (to stationary point) of UGNN under different assumptions of $\ell_Y$.

To start with, we first prove that as long as proximal gradient descent has a fixed point, the fixed point is a stationary point.
\begin{lemma}\label{lem:fix-is-station}
For any $Z$ such that
\begin{equation}Z \in \arg\min_Y \frac{1}{2\alpha}\left\|Y- \left(Z-\alpha \nabla \ell_Y^s(Z)\right)\right\|_{\mathcal F}^2 + \phi(Y), \end{equation}
it must hold that $0 \in \partial_Z \ell_Y(Z)$.
\end{lemma}
\beginproof{Proof}
We will use the summation property and local minimality property of Fr\'{e}chet subdifferential (see \citep{li2020understanding} for detailed descriptions). For any $Y'$ such that
\begin{align}
   &  Y' \in \arg\min_Y \frac{1}{2\alpha}\left\|Y- \left(Z-\alpha \nabla \ell_Y^s(Z)\right)\right\|_{\mathcal F}^2 + \phi(Y),
\end{align}
we have
\begin{align}
  0 \in \partial_Y \left. \left[\frac{1}{2\alpha}\left\|Y- \left(Z-\alpha \nabla \ell_Y^s(Z)\right)\right\|_{\mathcal F}^2 + \phi(Y)\right]\right|_{Y=Y'}.
\end{align}
It follows that
\begin{align}
 0\in \nabla \ell_Y^s(Z) + \partial_{Y'} \phi(Y').
\end{align}
So, if $Y'=Z$, we have \begin{equation}0 \in \nabla \ell_Y^s(Z) + \partial_Z \phi(Z) = \partial_Z\left[\ell_Y^s(Z) + \phi(Z)\right] = \partial_Z \ell_Y(Z)\end{equation}
\qed

With Lemma \ref{lem:fix-is-station}, to prove that UGNN converges to a stationary point, we only need to show it has a fixed point. 

For the convergence of UGNN, we assume $\phi$ is continuous and $\ell_Y$ is Lipschitz smooth (i.e. its gradient is Lipschitz continuous) with Lipschitz constant $\mathcal L$. And we further distinguish two cases 1) No other conditions provided 2) $\phi$ is convex and $\ell_Y$ is strongly convex. We will use different strategies to prove convergence and show that the conditions on the step-size that guarantee convergence are different in these two cases.

Note that, as mentioned in the main paper, in the first case, we also need to assume $\ell_Y(Y) \to \infty$ when $\|Y\|\to\infty$ to avoid a degenerate case that the process goes to infinite.

\subtitle{The Framework of Proving Convergence} 

To prove the convergence of the iteration of (\ref{eq:prox_step}), we will use Zangwill's convergence theorem. For clarity, we state it next.
\begin{proposition} [Zangwill's Convergence Theorem \citep{Luenberger84}]\label{fact:zangwill}
Let $A: \mathcal X \to 2^\mathcal X$ be a set-valued function, $\mathcal G \subset \mathcal X$ be a solution set (which can be any set we are interested in). For any sequence $\{{\boldsymbol{x}}_k\}_{k=1}^\infty$ such that ${\boldsymbol{x}}_{k+1} \in A({\boldsymbol{x}}_k)$, if the following conditions hold jointly:
\begin{enumerate}
    \item (boundedness) $\{{\boldsymbol{x}}_k | 1 \leq k \leq \infty \}$ is contained in some compact subset of $\mathcal X$.
    \item (descent function) there exists a continuous function $\ell$ such that \begin{enumerate}
        \item if ${\boldsymbol{x}} \not\in \mathcal G$, then $\forall {\boldsymbol{y}} \in A({\boldsymbol{x}}), \ell({\boldsymbol{y}}) < \ell({\boldsymbol{x}})$.
        \item if ${\boldsymbol{x}} \in \mathcal G$, then $\forall {\boldsymbol{y}}\in A({\boldsymbol{x}}), \ell({\boldsymbol{y}}) \leq \ell({\boldsymbol{x}})$.
    \end{enumerate}
    \item (closedness) the mapping $A$ is closed at all point of $\mathcal X \setminus \mathcal G$ i.e. $\{({\boldsymbol{x}},{\boldsymbol{y}}) | {\boldsymbol{x}} \in  \mathrm{clos}\left(\mathcal X \setminus \mathcal G\right) , {\boldsymbol{y}} \in A({\boldsymbol{x}})\}$ is a closed set, where $\mathrm{clos}$ means set closure. 
\end{enumerate}
then the limit of any convergent subsequence of $\{{\boldsymbol{x}}_k\}_{k=1}^\infty$ is a solution i.e. inside $\mathcal G$.
\end{proposition}

Given Lemma \ref{lem:fix-is-station}, we define the solution set $\mathcal G$ as set of fixed points: 
\begin{equation}\label{eq:solution-set}\mathcal G = \left\{Z \left| Z \in \arg\min_Y \frac{1}{2\alpha}\left\|Y- \left(Z-\alpha \nabla \ell_Y^s(Z)\right)\right\|_{\mathcal F}^2 + \phi(Y) \right.\right\}.\end{equation}

Then the proof is composed of three steps: \begin{enumerate}
    \item showing there exists some function that is a descent function; \label{stat:prof-local-conv-step-descent}
    \item showing the process is closed;\label{stat:prof-local-conv-step-closed}
    \item showing $\left\{Y^{(k)}\right\}_{k=1}^\infty$ are bounded. \label{stat:prof-local-conv-step-bounded}
\end{enumerate}

Thereinto, the closedness is a straightforward consequence of Lemma 6 in \citep{sriperumbudur2009convergence}. For boundedness, as long as we find some descent function $\ell$,  we consider the set $\mathcal S = \left\{Y ~|~ \ell(Y) \leq \ell\left(Y^{(0)}\right) \right\}$. Apparently $\left\{Y^{(k)}\right\}_{k=1}^\infty \subseteq \mathcal S$, so we only need to show that $\mathcal S$ is bounded. It suffices to prove  $\displaystyle \lim_{\|Y\|\to \infty}\ell(Y) = \infty$. Therefore, step \ref{stat:prof-local-conv-step-closed} is already finished and step \ref{stat:prof-local-conv-step-bounded} is nearly finished. The most tricky part is proving descent of some function.
 
\subsection{UGNN Convergence in General Case}\label{sec:ugnn-convergence-general}

Here we don't cast convexity assumption on the objective function. In this case we choose $\ell_Y$ as the candidate descent function. Note that by the assumption that  $\displaystyle \lim_{\|Y\|\to \infty}\ell_Y(Y) = \infty$ made in main paper, we have boundedness of $\mathcal S$.

To prove descent, we first prove loosely descent (i.e. Lemma \ref{lem:descent}) and then show when $Y^{(k)} \not \in \mathcal G$, it is also strict.

\beginproof{Proof of Lemma \ref{lem:descent}}~ Here we  first state a property of Lipschitz smooth function:\begin{proposition}[\citep{Bertsekas1999}]\label{fact:lipschitz}
for any $\ell({\boldsymbol{y}})$ whose gradient is $\mathcal L$-Lipschitz, it satisfies \begin{equation}\forall {\boldsymbol{y}},{\boldsymbol{z}}, \ell({\boldsymbol{y}}) \leq \ell({\boldsymbol{z}}) + \nabla \ell({\boldsymbol{z}})^\top({\boldsymbol{y}}-{\boldsymbol{z}}) + \frac{\mathcal L}{2}\|{\boldsymbol{y}}-{\boldsymbol{z}}\|_2^2. \end{equation}
\end{proposition}

From Proposition \ref{fact:lipschitz}, if function $\ell$ is $\mathcal L$-Lipschitz smooth,  then for any ${\boldsymbol{z}},{\boldsymbol{y}}$, we have
\begin{align}
     \ell ({\boldsymbol{y}}) & \leq \ell({\boldsymbol{z}}) + \nabla \ell({\boldsymbol{z}})^\top({\boldsymbol{y}}-{\boldsymbol{z}}) + \frac{\mathcal L}{2}\|{\boldsymbol{y}}-{\boldsymbol{z}}\|_2^2
    \\ & = \ell({\boldsymbol{z}}) + \nabla \ell({\boldsymbol{z}})^\top {\boldsymbol{y}} - \nabla \ell({\boldsymbol{z}})^\top {\boldsymbol{z}} + \frac{\mathcal L}{2}\|{\boldsymbol{y}}\|_2^2 + \frac{\mathcal L}{2} \|{\boldsymbol{z}}\|_2^2 - \mathcal L{\boldsymbol{y}}^\top {\boldsymbol{z}}
    \\ & = \ell({\boldsymbol{z}}) - \frac{1}{2\mathcal L}\|\nabla \ell({\boldsymbol{z}})\|_2^2 \nonumber\\
    &~~+ \frac{\mathcal L}{2} \left[\frac{1}{\mathcal L^2}\|\nabla \ell({\boldsymbol{z}})\|_2^2 +  \|{\boldsymbol{y}}\|_2^2 + \|{\boldsymbol{z}}\|_2^2 + \frac{2}{\mathcal L}\nabla\ell({\boldsymbol{z}})^\top {\boldsymbol{y}} - \frac{2}{\mathcal L}\nabla \ell({\boldsymbol{z}})^\top {\boldsymbol{z}} - 2{\boldsymbol{y}}^\top {\boldsymbol{z}} \right]
    \\ & = \ell ({\boldsymbol{z}}) - \frac{1}{2\mathcal L}\|\nabla \ell({\boldsymbol{z}})\|_2^2 + \frac{\mathcal L}{2} \left\|{\boldsymbol{y}} - \left[{\boldsymbol{z}} - \frac{1}{\mathcal L}\nabla \ell({\boldsymbol{z}})\right] \right\|_2^2. \label{eq:local-conv-step-1-prior}
\end{align}

Taking $\ell({\boldsymbol{y}}) = \ell_Y^s({\boldsymbol{y}})$, ${\boldsymbol{y}}=\mathrm{vec} (Y)$ and ${\boldsymbol{z}}=\mathrm{vec} \left(Y^{(k)}\right)$, and let \begin{equation}\beta_Y\left(Y^{(k)}\right) =\ell_Y^s\left(Y^{(k)}\right) - \frac{\alpha}{2}\left\|\nabla \ell_Y^s\left(Y^{(k)}\right)\right\|_{\mathcal F}^2,\end{equation}we can define an upper-bound of $\ell_Y$: \begin{align}\ell_Y^{(\mathrm{upp})}(Y;Y^{(k)}) & = \frac{1}{2\alpha} \left\|Y - \left[Y^{(k)} - \alpha\nabla \ell\left(Y^{(k)}\right)\right] \right\|_{\mathcal F}^2 + \sum_{i}\phi({\boldsymbol{y}}_i) + \beta_Y\left(Y^{(k)}\right)
\\ & = \frac{1}{2\alpha} \left\|Y - U^{(k)} \right\|_{\mathcal F}^2 + \sum_{i}\phi({\boldsymbol{y}}_i) + \beta_Y\left(Y^{(k)}\right), \label{eq:def-upperbound}\end{align}
and adopting (\ref{eq:local-conv-step-1-prior}) we have that:\begin{align}
\ell_Y(Y) = \ell_Y^s(Y) + \sum_{i}\phi({\boldsymbol{y}}_i)  \leq \ell_Y^{(\mathrm{upp})}.
\end{align}(Notice that since we have assumed $\alpha \leq \frac{1}{\mathcal L}$, $\ell_Y$ is also $\frac{1}{\alpha}$-Lipschitz).
It's not difficult to check $\ell_Y^{(\mathrm{upp})}\left(Y^{(k)};Y^{(k)}\right) = \ell_Y\left(Y^{(k)}\right)$. Also, since adding terms which is not related to $Y$ does not affect the $\arg\min$ result in (\ref{eq:prox_step}), we have that \begin{align}
Y^{(k+1)} & \in \arg\min_Y \frac{1}{2\alpha}\|Y-U^{(k)}\|_{\mathcal F}^2 + \sum_i \phi({\boldsymbol{y}}_i)
\\ & = \arg\min_Y \frac{1}{2\alpha}\|Y-U^{(k)}\|_{\mathcal F}^2 + \sum_i \phi({\boldsymbol{y}}_i) + \beta_Y\left(Y^{(k)}\right)
\\ & = \arg\min_Y \ell_Y^{(\mathrm{upp})}\left(Y;Y^{(k)}\right)
.\end{align}
Then we have \begin{align}\label{eq:loose-descent}\ell_Y\left(Y^{(k+1)}\right) \leq \ell_Y^{(\mathrm{upp})}\left(Y^{(k+1)};Y^{(k)}\right) \leq \ell_Y^{(\mathrm{upp})}\left(Y^{(k)};Y^{(k)}\right) = \ell_Y\left(Y^{(k)}\right), \end{align}
which proves the the lemma.
\qed

Furthermore, when $Y^{(k)} \not \in \mathcal G$, i.e.$Y^{(k)} \not \in \arg\min \ell_Y^{(\mathrm{upp})}\left(Y;Y^{(k)}\right)$, we have that \begin{align}\ell_Y^{(\mathrm{upp})}\left(Y^{(k+1)};Y^{(k)}\right) < \ell_Y^{(\mathrm{upp})}\left(Y^{(k)};Y^{(k)}\right),\end{align}
and thus (\ref{eq:loose-descent}) holds strictly, which matches the condition of strictly descent given in Proposition \ref{fact:zangwill}.

\subsection{UGNN Convergence Under Convexity}\label{ugnn-convergence-convex}

When $\ell_Y^s$ is strongly convex and $\phi$ is convex, we have already know the existence and uniqueness of stationary point (global optimum) $Y^*$, so by Lemma \ref{lem:fix-is-station} the fixed point of iteration (\ref{eq:prox_step}) must be $Y^*$, i.e. $\mathcal G = \{Y^*\}$. In this case, we choose the distance of current point to the fixed point $\ell(Y) = \left\|Y-Y^*\right\|_\mathcal F^2$ as the descent function. As long as descent is proved, the boundedness of $\ell(Y)$ is obvious. 

We prove the descent of $\ell(Y)$ by using the fact that when $\phi$ and $\ell_Y^s$ are convex, the proximal operator $\mathrm{prox}_\phi$ is non-expansive and the gradient step is contraction with small enough $\alpha$.

For proximal operator, Proposition 1 in \citep{gribonval2020characterization} has already shown that as long as $\phi$ is convex, $\mathrm{prox}_\phi$ is non-expansive, i.e. \begin{equation}\label{eq:prox-is-non-expansive}\forall {\boldsymbol{x}},{\boldsymbol{y}},\left\| \mathrm{prox}_\phi({\boldsymbol{x}})-\mathrm{prox}_\phi({\boldsymbol{y}})\right\|_2 \leq \|{\boldsymbol{x}}-{\boldsymbol{y}}\|_2.\end{equation}
Thus we only need to prove gradient descent is contraction. To prove this, we first introduce another property of convex Lipschitz smooth function:
\begin{proposition}[Theorem 2.1.5 of \citep{nesterov2003introductory}]\label{fact:lipschitz-convex-descent} For any function $\ell$ that is convex and whose gradient is $\mathcal L$-Lipschitz continuous, we have \begin{equation}\forall {\boldsymbol{x}},{\boldsymbol{y}}, \left[\nabla \ell({\boldsymbol{x}}) - \nabla\ell({\boldsymbol{y}}) \right]^\top({\boldsymbol{x}}-{\boldsymbol{y}}) \geq \frac{1}{\mathcal L}\|\nabla \ell({\boldsymbol{x}}) - \nabla \ell({\boldsymbol{y}})\|_2^2.\end{equation}
\end{proposition}
Then we prove the contraction of gradient descent:
\begin{lemma}\label{lem:gradient-is-contraction}
for any function $\ell$ that is $\mathcal L$-Lipschitz smooth satisfies that \begin{equation}\forall {\boldsymbol{x}},{\boldsymbol{y}}, \left\|\left[{\boldsymbol{x}} - \alpha \nabla \ell({\boldsymbol{x}})\right] - \left[{\boldsymbol{y}} - \alpha \nabla \ell({\boldsymbol{y}})\right]\right\|_2^2 \leq (1-\alpha^2\mathcal  L^2 - 2\alpha\mathcal L)\|{\boldsymbol{x}}-{\boldsymbol{y}}\|_2^2.
\end{equation}
\end{lemma}
\beginproof{Proof}
From Proposition \ref{fact:lipschitz-convex-descent}, we have that:
\begin{align}
& \left\|\left[{\boldsymbol{x}} - \alpha \nabla \ell({\boldsymbol{x}})\right] - \left[{\boldsymbol{y}} - \alpha \nabla \ell({\boldsymbol{y}})\right]\right\|_2^2 
\\ = & \|{\boldsymbol{x}}-{\boldsymbol{y}}\|_2^2 + \alpha^2 \|\nabla \ell({\boldsymbol{x}}) - \nabla\ell({\boldsymbol{y}})\|_2^2 - 2\alpha ({\boldsymbol{x}}-{\boldsymbol{y}})^\top \left[\nabla \ell({\boldsymbol{x}}) - \nabla\ell({\boldsymbol{y}})\right]
\\ \leq & \|{\boldsymbol{x}}-{\boldsymbol{y}}\|_2^2 + \left(\alpha^2 - \frac{2\alpha}{\mathcal L}\right) \left\|\nabla \ell({\boldsymbol{x}}) - \nabla\ell({\boldsymbol{y}})\right\|_2^2
\\ \leq & (1-\alpha^2\mathcal  L^2 - 2\alpha\mathcal L)\|{\boldsymbol{x}}-{\boldsymbol{y}}\|_2^2,
\end{align}
which finishes the proof.
\qed

Combining (\ref{eq:prox-is-non-expansive}) and Lemma \ref{lem:gradient-is-contraction}, we get following corollary:
\begin{corollary}
For $\ell_Y(Y) = \ell_Y^s(Y) + \phi(Y)$, where $\ell_Y^s$ is strongly convex and $\mathcal L$-Lipschitz smooth, $\phi$ is convex, $\alpha < \frac{2}{\mathcal L}$, we have \begin{equation}\left\|Y^{(k+1)} - Y^*\right\|_\mathcal F < \left\|Y^{(k)} - Y^*\right\|_\mathcal F\end{equation}if $Y^{(k)} \neq Y^*$.
\end{corollary}
\beginproof{Proof} Since $Y^*$ is the fixed point, we have $Y^* \in \mathrm{prox}_\phi\left[Y^* - \alpha \nabla \ell_Y^s(Y^*)\right]$, therefore by (\ref{eq:prox-is-non-expansive}) \begin{align} \left\|Y^{(k+1)} - Y^*\right\|_\mathcal F & = \left\|  \mathrm{prox}_\phi\left[Y^{(k)} - \alpha\nabla\ell_Y^s\left(Y^{(k)}\right)\right] - \mathrm{prox}_\phi\left[Y^* - \alpha \nabla \ell_Y^s(Y^*)\right]\right\|_\mathcal F ^2
\\ & \leq \left\|\left[Y^{(k)} - \alpha\nabla\ell_Y^s\left(Y^{(k)}\right) \right] - \left[Y^* - \alpha \nabla \ell_Y^s(Y^*)\right]\right\|_\mathcal F^2
\end{align}

Since $\alpha < \frac{2}{\mathcal L}$, from Lemma \ref{lem:gradient-is-contraction} we have that when $Y^{(k)} \neq Y^*$, \begin{equation}\left\|\left[Y^{(k)} - \alpha\nabla\ell_Y^s\left(Y^{(k)}\right) \right] - \left[Y^* - \alpha \nabla \ell_Y^s(Y^*)\right]\right\|_\mathcal F^2 < \left\|Y^{(k)}-Y^*\right\|.\end{equation}
\qed

Therefore, $\ell\left(Y^{(k)}\right) = \left\|Y^{(k)}-Y^*\right\|$ is strictly descent when $Y^{(k)}\neq Y^*$. Combining this and Proposition \ref{fact:zangwill} we conclude that:
\begin{theorem}\label{theorem:ugnn-convergence-convex}
    When $\ell_Y^s(Y)$ is $\mathcal L$-Lipschitz smooth and strongly convex, $\phi(Y)$ is continuous and convex, iteration (\ref{eq:prox_step}) converges to the unique global minimum of $\ell_Y$ when $\alpha < \frac{2}{\mathcal L}$.
\end{theorem} 

\subsection{Remarks}

Note that, the upper-bound of $\alpha$ is different in the two cases. Actually, if there is no penalty term $\phi$, i.e. no proximal step, in both cases the upper bound is $2 / \mathcal L$, which is a standard result on gradient descent for smooth functions that can be found in any optimization textbook. To get some intuition, considering the upper bound derived in (\ref{eq:def-upperbound}), although $\alpha = 1 / \mathcal L$ is the optimal step size to minimize $\ell^{(\mathrm{upp})}_Y$, with $\alpha$ relaxed to $\leq 2/\mathcal L$, $\ell^{(\mathrm{upp})}_Y$ still goes down: \begin{equation}\ell_Y^{(\mathrm{upp})}\left[Y^{(k)}-\alpha \nabla \ell_Y^s\left(Y^{(k)}\right);Y^{(k)}\right] \leq \ell_Y^{(\mathrm{upp})}\left(Y^{(k)};Y^{(k)}\right) = \ell_Y^s\left(Y^{(k)}\right).\end{equation}
However, this is no longer true when the proximal step is added. Thus in the latter case, we either cast a more strict bound on $\alpha$ to still guarantee descent of $\ell_Y$, or add extra assumptions that the process is non-expansive to guarantee that the output of each iteration would not be pushed further away from the fixed point despite not ensuring the descent of $\ell_Y$.


\subsection{Other Related Proofs}

Note that Lemma \ref{lem:descent} and Theorem \ref{thm:local_convergence} has been proved in section \ref{sec:ugnn-convergence-general} \footnote{The last part of Theorem \ref{thm:local_convergence} that $\displaystyle \lim_{k\to \infty} \ell_Y\left(Y^{(k)}\right) = \ell_Y\left(Y^*\right)$ follows directly from the continuity of $\ell_Y$.}. Here we provide other missing proofs from Section \ref{sec:global-convergence}.

\beginproof{Proof of Theorem \ref{thm:global_convergence}} First, we will show that $\ell_Y^s$ is strongly convex under conditions given. We have \begin{align}
\mathrm{vec} \left (\frac{\partial \ell_Y^s}{\partial Y } \right) & = \mathrm{vec} \left[ YW_f^s- f(X;W) + \tilde L Y W_p^s \right]
\\ & = \left(W_f^s \otimes I + W_p^s \otimes \tilde L\right)\mathrm{vec} (Y) + \mathrm{vec} \left[f(X;W)\right]
\\ & = \Sigma \mathrm{vec} (Y) + \mathrm{vec} \left[f(X;W)\right].
\end{align}
Therefore the Hessian of $\ell_Y^s$ is \begin{equation}\label{eq:sigma-is-hessian} \frac{\partial ^2\ell_Y^s}{\partial \mathrm{vec} (Y) ^ 2} = \Sigma.\end{equation}
We also know that when $\lambda_{\text{min}}(\Sigma) > 0$, $\ell_Y^s$ is strongly convex respect to $Y$, thus has a unique global minimum. And from Proposition 1 in \citep{gribonval2020characterization}, we know $\phi$ is convex through the non-expansiveness of $\sigma$.

Furthermore, the $\lambda_{\text{max}}(\Sigma)$ is the Lipschitz constant of $\ell^s_Y$, and from theorem \ref{theorem:ugnn-convergence-convex} we know when $\alpha < 2 / \lambda_{\text{max}}(\Sigma)$ the algorithm converges to the stationary point, which is the global optimum.
\qed

\beginproof{Proof of Corollary \ref{cor:global_convergence1}}~ Taking $W_f^s = I-W_p^s$ and $P=I-\tilde L$, we have that \begin{equation}\Sigma = I - P \otimes W_p^s. \end{equation}
By the spectral properties of Kronecker product, we have \begin{equation}\lambda_{\text{min}}(\Sigma) \geq 1- \left\|W_p^s\right\|_2 \left\|P\right\|_2 > 0\end{equation}
and \begin{equation}\lambda_{\text{max}}(\Sigma) \leq 1+ \left\|W_p^s\right\|_2 \left\|P\right\|_2 < 2.\end{equation}

The discussion in Section \ref{sec:ugnn_fixed_point} mentioned that to get (\ref{eq:ugnn_fixed_point_simple}), we need to set $\alpha = 1 < \frac{2}{\lambda_{\text{max}}(\Sigma)}$, therefore from Theorem \ref{thm:global_convergence} we conclude that (\ref{eq:ugnn_fixed_point_simple}) converges to the global minimum of (\ref{eq:unfolded_objective_special_case2}).
\qed

\beginproof{Proof of Lemma \ref{cor:global_convergence2} (Convergence of IGNN)} We first convert (\ref{eq:IGNN_iteration}) to the vector form:\begin{equation} \label{eq:IGNN-iteration_v}
\mathrm{vec} \left(Y^{(k+1)}\right) = \sigma\left[ \left(W_p \otimes P\right) \mathrm{vec} \left(Y^{(k)}\right) + \mathrm{vec} (f)\right],
\end{equation}
where $f = f(X;W_x)$, ignoring the parameters for simplicity. Let $M = W_p\otimes P$. From the spectral properties of knronecker product, we have \begin{equation}\|M\|_2 = \|W_p\|_2\|P\|_2 < 1.\end{equation}

By definition of matrix norm we have \begin{equation}\forall {\boldsymbol{x}}_1,{\boldsymbol{x}}_2, \|M({\boldsymbol{x}}_1-{\boldsymbol{x}}_2)\|_2 \leq \|M\|_2 \|{\boldsymbol{x}}_1-{\boldsymbol{x}}_2\|_2,\end{equation}
which means $M$ (as a linear transformation) is a contraction mapping on Euclidean space. We also assumed $\sigma$ is contraction mapping, and thus from Banach's fixed point theorem\citep{kinderlehrer1980introduction} we conclude that (\ref{eq:IGNN-iteration_v}), as well as (\ref{eq:IGNN_iteration}), has a unique fixed point.
\qed

\section{Proofs Related to UGNN Expressiveness}

In this section, we provide detailed proofs of propositions in Section \ref{sec:shared_weights}, as well as an analysis under cases where activation functions are non-linear.

\subsection{Proof of Lemma \ref{lem:proximal}} This lemma is a straight-forward corollary of Theorem 1 in \citep{gribonval2020characterization}. For clarity, we excerpt the theorem here:
\begin{proposition}[Theorem 1 in \citep{gribonval2020characterization}]\label{fact:characterization-proximal}

Let $\mathcal H$ be some Hilbert space (e.g. $\mathbb R^k$) and $\mathcal Y \subset \mathcal H$ be non-empty. A function $\sigma: \mathcal Y \to \mathcal H$ is the proximal operator of a function $\phi: \mathcal H \to \mathbb R\cup \{+\infty\}$ if and only if there exists a convex l.s.c.~(lower semi-continuous) function $\psi: \mathcal H \to \mathbb R\cup \{+\infty\}$ such that for each ${\boldsymbol{y}} \in \mathcal Y, \sigma({\boldsymbol{y}}) \in \partial \psi({\boldsymbol{y}})$.
\end{proposition}

Since our $\sigma$ is component-wise, we only need to consider $\mathcal Y = \mathcal  H = \mathbb R$. Furthermore, since $\sigma({\boldsymbol{x}})$ is continuous, its indefinite integration exists.  Let $\psi({\boldsymbol{x}}) = \int \sigma({\boldsymbol{x}}) \mathrm{d} {\boldsymbol{x}}$, which must be convex since its derivative is non-decreasing. Then according to Proposition \ref{fact:characterization-proximal}, there exists a function $\phi$ whose proximal operator is $\sigma$. 
Conversely, if there exists $\phi$ such that $\sigma = \mathrm{prox}_\phi$, from Proposition \ref{fact:characterization-proximal} it is the subdifferential of some convex function $\psi$, which means it is non-decreasing.

\subsection{Proof of Lemma \ref{lem:no-asym-integrate}}

We only need a counterexample to prove this. Suppose $n=1$ and $d=2$, consider $W = \begin{bmatrix}0  & 1 \\ 0 &  0 \end{bmatrix}$, $Y = \begin{bmatrix}a & b\end{bmatrix}$, then $YW = \begin{bmatrix}0 \\ a\end{bmatrix}$.

If there exists any second order smooth function $h$ such that $\frac{\partial h}{\partial a} = 0$ and $\frac{\partial h}{\partial b} = a$, we have that $$\frac{\partial^2 h}{\partial a\partial b} = 0 \neq \frac{\partial ^2 h}{\partial b\partial a} = 1 ,$$which contradicts the second order smoothness assumption of $h$.

\subsection{Proof of Theorem \ref{theo:equivalency-linear}}

In this subsection we denote $\hat X = f(X;W_x)$ for simplicity. First consider a case which allows $T$ and $W_p^s$ to have complex values. Hereinafter we denote the space of all complex-valued matrices with shape $d_1 \times d_2$ by $\mathbb C^{d_1\times d_2}$.

We recall Jordan's decomposition theorem below. Let $J(\lambda)$ be Jordan block matrix $$J(\lambda) = \begin{bmatrix}\lambda & 1 &   &  &   \\   & \lambda & 1 &  &   \\   &   & \lambda &  &   \\  &  &  & \ddots & 1 \\   &   &   &  & \lambda \end{bmatrix}.$$ 
We have the following well-known fact.
\begin{proposition}[Jordan]
For every $W \in \mathbb R^{d\times d}$, there exists a complex-valued invertible matrix $P \in \mathbb C^{d\times d}$ and complex-valued block-diagonal matrix $$\Omega = \begin{bmatrix}J(\lambda_1) &   &  &   \\   & J(\lambda_2) &   &   \\  &  & \ddots &  \\   &   &  & J(\lambda_k) \end{bmatrix}$$ 
such that $W = P\Omega P^{-1}$, where $\lambda_j \in \mathbb C$ is the $j$-th eigenvalue of $W$ and the size of $J(\lambda_j)$ is the algebraic multiplicity of $\lambda_j$.
\end{proposition}

\begin{corollary}\label{cor-diag}
If a square matrix $W \in \mathbb  R^{d\times d}$ has $d$ distinct eigenvalues, then there exists $P,\Lambda \in \mathbb C^{d\times d}$ where $P$ is invertible and $\Lambda$ is diagonal, such that $W = P\Lambda P^{-1}$.
\end{corollary}

Then, we shall show that actually on the complex domain ``almost" every matrix can be diagonalized, which means, for each matrix $W$, either it itself can be diagonalized, or there's a diagonalizable matrix $W'$ that is arbitrarily closed to $W$.

\begin{corollary}\label{cor-cont}
For every square matrix $W \in \mathbb R^{d\times d}$ and any $\epsilon > 0$, there exists a diagonalizable square matrix $W' \in \mathbb R^{d\times d}$ such that $\|W'-W\| \leq \epsilon$.
\end{corollary}
\beginproof{Proof} Let $W = R\Omega R^{-1}$ and $E = \mathrm{diag}\begin{bmatrix}\epsilon_1, & \epsilon_2, & \cdots, & \epsilon_n\end{bmatrix}$ such that the diagonal of $\Omega + E$ is distinct and $\epsilon_j^2 <  \frac{\epsilon^2}{d\|R\|^2\|R^{-1}\|^2}$. 

Since $\Omega$ is upper-triangle, its eigenvalues are the elements in its diagonal, which are distinct. Thus according to Corollary \ref{cor-diag}, $\Omega+E$ is diagonalizable, suppose $\Omega + E = Q\Lambda Q^{-1}$. Let $W' = R(\Omega+E)R^{-1} = RQ\Lambda Q^{-1}R^{-1}$, it is apparent $W'$ also diagonalizable.

Now consider the difference between $W'$ and $W$. We have $\|W'-W\|^2 = \|PEP^{-1}\|^2 \leq \|E\|^2\|R\|^{2}\|R^{-1}\|^2 < \epsilon^2$.
\qed

\noindent Note that the norm in Corollary \ref{cor-cont} can be any matrix norm.

\begin{lemma} \label{lem-cont}
The solution to the equation $PYW + \hat X = Y$ is continuous respect to $W$ as long as a unique  solution exists.

\end{lemma}
\beginproof{Proof} The solution of this equation is $\mathrm{vec} (Y) = \left(I - W^\top \otimes P\right)^{-1}\mathrm{vec}(\hat X)$, which is continuous.
\qed

\begin{theorem}\label{theo-lin-comp}
For any $W_p \in \mathbb R^{d\times d}$ that admits a unique fixed point for IGNN and $\hat X \in \mathbb R^{n \times d}$, suppose $Y\in \mathbb R^{n\times d}$ is the only solution of $PYW+\hat X=Y$, then there exists a $Y' \in \mathbb C^{n \times d}, \bar W_x \in \mathbb C^{d\times d'}$, right-invertible $\bar T \in \mathbb C^{d\times d'}$ and hermitian $\bar W_p^s\in \mathbb C^{d'\times d'}$, 
such that \begin{equation}
PY'\bar T\bar W_p^s + \hat X\bar W_x = Y'\bar T, \text{ with } \|Y'-Y\| < \epsilon, \forall \epsilon > 0, Y'\in \mathbb C^{n\times d}.\end{equation}
\end{theorem}
\beginproof{Proof}
By Corollary \ref{cor-cont} and Lemma \ref{lem-cont}, there exists diagonalizable $W' \in \mathbb R^{d\times d}$ such that $PY'W' + \hat X = Y'$ for some $Y'$ satisfies $\|Y'-Y\| < \epsilon$.

Suppose now $W' = R\Lambda R^{-1}$ where $\Lambda$ is diagonal and $R$ is invertible. We have \begin{equation}PY'R\Lambda R^{-1} + \hat X = Y', \end{equation} and right-multiply by $R$ at each side of this equation we get \begin{equation}P(Y'R)\Lambda + \hat X R = Y'R.\end{equation}
By choosing $\bar T = \bar W_x = R$ and $\bar W_p^s = \Lambda$, we proved the theorem.
\qed

Next, we consider restricting this result to the real domain. This is straight-forward since every complex linear transformation can be converted to real linear transformation by consider real and imaginary parts separately. Hereinafter, we denote the real part of an complex-valued matrix $M$ by $M_{(r)}$ and imaginary part by $M_{(i)}$.

\beginproof{Proof of Theorem \ref{theo:equivalency-linear}} Let $\bar Y$, $\bar T$, $\bar W_p^s$, $\bar W_x$ be the matrices obtained from Theorem \ref{theo-lin-comp}. Note the last three are complex-valued, $\bar T$ is invertible and $\bar W_p^s$ is hermitian. 

Consider $T = \begin{bmatrix}\bar T_{(r)} & \bar T_{(i)}\end{bmatrix}$. Since $\bar T$ is invertible, let $\bar \Tinv = \bar T^{-1}$, then we have $I = (\bar T\bar \Tinv)_{(r)} = \bar T_{(r)}\bar \Tinv_{(r)} - \bar T_{(i)}\bar \Tinv_{(i)}$. Thus $T \begin{bmatrix}\bar \Tinv_{(r)} \\ -\bar \Tinv_{(i)}\end{bmatrix} = \bar T_{(r)}\bar \Tinv_{(r)} - \bar T_{(i)}\bar \Tinv_{(i)} = I$. This means $T$ is right-invertible.

Then let $W_s^p = \begin{bmatrix}\bar W^s_{p(r)} & \bar W^s_{p(i)} \\ -\bar W^s_{p(i)} & \bar W^s_{p(r)}\end{bmatrix}$. Since $\bar W^s_p$ is hermitian, we have that $\bar W^s_{p(r)} = \bar W^{s\top}_{p(r)}$ and $\bar W^s_{p(i)} = -\bar W^{s\top}_{p(i)}$, which ensures $W^s_p$ constructed here is symmetric.
Finally, let $\tilde W_x = \begin{bmatrix}\bar W_{x(r)} & \bar W_{x(i)}\end{bmatrix}$.
Now it's easy to verify that $PY'TW_p^s + \hat X\tilde W_x = Y'T$.
\qed

\subsection{Proof of Theorem \ref{thm:equivalency-finite}}

First we will illustrate that by setting the parameters in UGNN to properly, the iteration step from (\ref{eq:prox_step}) is equivalent to a GCN layer with symmetric shared weights, with or without residual connections. We set $\rho$ to be identity (so that $\Gamma = I$), $\tilde B$ satisfies that $\tilde B^\top \tilde B = I-P$ (e.g. if $P=\hat A = \tilde D^{-1/2}\tilde A\tilde D^{-1/2}$ is normalized adjacency, then $\tilde B$ is normalized incidence matrix $\tilde B = \tilde D^{-1/2}B$).

For a GCN model without residual connections, let $W_f^s = I-W_p^s$, then we have $Y^{(k+1)} = \sigma(PY^{(k)}W_p^s)$. Consider $ Y^{(0)} =W \begin{bmatrix}Y^{(0)}_{\mathrm{GCN}} & 0 & 0 & \cdots & 0\end{bmatrix}$ where $Y^{(0)}_{\mathrm{GCN}}$ is the input of GCN, and the $i$th $0$ here denote a block matrix of all zero with the same size as $Y^{(i)}_{\mathrm{GCN}}$. When we let 
\begin{equation}\label{eq:finit-gcn-wps}
W_p^s = \begin{bmatrix}
         & W_p^{(1)} &  &  &  & 
\\    W_p^{(1)\top}  & & W_p^{(2)} & &  & 
\\   & W_p^{(2)\top}  &  & W_p^{(3)} &  & 
\\   &  & W_p^{(3)\top} &  & \ddots & 
\\   &  & & \ddots &  & W_p^{(k)}
\\   &  & & & W_p^{(k)\top} & 
\end{bmatrix},\end{equation}
it is not difficult to verify that the $k$th block of $Y^{(k)}$ is $Y^{(k)}_{\mathrm{GCN}}$. Note the $0$s in the (\ref{eq:finit-gcn-wps}) are block matrices of all zeros with proper size.

For GCN model with residual connection, let $W_f^s = I-W_p^s-W_r^s$, then $Y^{(k+1)} = \sigma(PY^{(k)}W_p^s + Y^{(k)}W_r^s)$. We take the same $W_p^s$ as before (\ref{eq:finit-gcn-wps}), and let \begin{equation}\label{eq:finit-gcn-wrs}
W_r^s = \begin{bmatrix}
         & I &  &  &  & 
\\    I  & & I & &  & 
\\   & I  &  & I &  & 
\\   &  & I &  & \ddots & 
\\   &  & & \ddots &  & I
\\   &  & & & I & 
\end{bmatrix}.\end{equation}
Since $W_r^s$ and $W_p^s$ are both symmetric, apparently $W_f^s$ is also symmetric. Also notice that since residual connection exists, the size of all the $W_p^{(i)}$s are the same, so the size of $I$ in the (\ref{eq:finit-gcn-wrs}) is the same as $W_p^{(i)}$.

\subsection{A Sufficient Condition of Equivalence of Symmetric and Asymmetric Weights}

In this section, we give a sufficient condition of fixed-point equivalence of symmetric and asymmetric weights in general case as mentioned in Section \ref{sec:fix-point-equivalency}. Here we consider general nonlinear functions $\sigma$ and penalty term $\phi$. Hereinafter we denote $\mathrm{prox}_\phi$ by $\tilde \sigma$. 

\subtitle{Constraint on $\sigma$}~ To simplify the discussion, we add a mild assumption on $\sigma$ and consider a general type of $\tilde \sigma$: We assume $\sigma({\boldsymbol{x}})$ is a proximal operator of some penalty function which is differentiable with respect to $x$. 

This assumption is consistent with practice. Firstly, commonly used non-linearities like $\mathrm{tanh}$, $\mathrm{sigmoid}$ and ReLU are all proximal operators since they are all continuous and component-wise non-decreasing (By Lemma \ref{lem:proximal}). Also, although at some point ReLU and Leaky-ReLU are not differentiable, we can approximate them by $$\text{ReLU}(x) \approx  \frac{1}{r} \log\left( \exp(rx)+1 \right)$$ and $$\text{Leaky-ReLU}(x,p) \approx  \frac{1}{r} \log\left( \exp(rx)+\exp(prx) \right)$$
with big enough $r$, which are differentiable and do not affect their practical attributes.

\subtitle{$\tilde\sigma$ Considered}~ As discussed in the main paper, it would cause some degenerated cases if we allow a general class of proximal operators. Also it would too difficult to handle in this case. Therefore, we only consider proximal operators in a special while general family $$\mathcal S = \{\tilde\sigma: {\boldsymbol{x}} \mapsto G\sigma(C{\boldsymbol{x}}) | G,C \text{ are chosen such that } \tilde\sigma \text{ is proximal operator}\}$$

Now we consider under what value of $G$ and $C$ we can ensure $\tilde\sigma$ is proximal operator. From Proposition \ref{fact:characterization-proximal}, we know $\exists \phi, \tilde \sigma = \mathrm{prox}_\phi$ iff $\tilde \sigma$ is subgradient of some l.s.c convex function. Since we have assumed $\sigma$ is continuous and differentiable, $\int \sigma({\boldsymbol{x}}) \mathrm{d} {\boldsymbol{x}}$ is twice differentiable. We know that a twice-differentiable function $\psi$ is convex iff its Hessian $H_\psi$ is positive semi-definite. Therefore the Hessian of $\sigma$ must be positive semi-definite, we denote it by $H({\boldsymbol{x}}) = \frac{\partial \sigma({\boldsymbol{x}})}{\partial {\boldsymbol{x}}}$, and so do Hessian of $\tilde\sigma({\boldsymbol{x}})$, which we denote by $\tilde H({\boldsymbol{x}}) = \frac{\partial \tilde\sigma({\boldsymbol{x}})}{\partial {\boldsymbol{x}}}$. From the chain rule we have $$\tilde H({\boldsymbol{x}}) = \frac{\partial \tilde \sigma({\boldsymbol{x}})}{\partial {\boldsymbol{x}}} = \frac{\partial G\sigma(C{\boldsymbol{x}})}{\partial {\boldsymbol{x}}} = G \frac{\partial \sigma(C{\boldsymbol{x}})}{\partial C{\boldsymbol{x}}}\frac{\partial C{\boldsymbol{x}}}{\partial {\boldsymbol{x}}}= GH(C{\boldsymbol{x}})C.$$

From the deduction above, we can conclude that:
\begin{lemma} \label{lem-tilde-sigma}
$\tilde\sigma: \mathbb R^d \to \mathbb R^d, {\boldsymbol{x}}\mapsto G\sigma(C{\boldsymbol{x}})$ is proximal operator iff $\forall {\boldsymbol{x}} \in C\mathbb R^d, GH({\boldsymbol{x}})C$ is positive semi-definite where $H({\boldsymbol{x}}) = \frac{\partial \sigma({\boldsymbol{x}})}{\partial {\boldsymbol{x}}}$.
\end{lemma}

\subtitle{The Condition of General Fixed-Point Alignment}~ 

\begin{theorem} \label{theo-fix-poi-cond}
For $\sigma$ satisfying the assumptions above, $W_p \in \mathbb  R^{d\times d}$ that admits unique fixed point for IGNN, and $W_x\in \mathbb R^{d_0\times d}$, suppose $Y$ is the only solution of $Y =  \sigma(PYW+XW_x)$, a sufficient condition of    
    \begin{gather*}\exists \phi, \text{ right invertible } T \in \mathbb R^{d\times d'}, \text{ symmetric } W_p^s \in \mathbb R^{d'\times d'} \text{ and } \tilde W_x \in \mathbb R^{d\times d} \text{ such that } \\
         \tilde\sigma( PYTW_p^s + XW_x) = YT
    \end{gather*}
is that \begin{gather*}
    \exists \text{ right-invertible } T \in \mathbb R^{d \times d'},C \in \mathbb R^{d'\times d},W_p^s \in \mathbb R^{d'\times d'} \text{ such that } TW_p^s C = W_p \\ \text{ and } \forall {\boldsymbol{x}} \in C^\top \mathbb R, T^\top H({\boldsymbol{x}})C^\top  \text{ is P.S.D,}\end{gather*}
    where $\tilde \sigma({\boldsymbol{x}}) = \mathrm{prox}_\phi({\boldsymbol{x}})$ and $H({\boldsymbol{x}}) = \frac{\partial \sigma({\boldsymbol{x}})}{\partial {\boldsymbol{x}}}$.
\end{theorem}
\beginproof{Proof}
We have assumed $\tilde\sigma \in \mathcal S$, thus $\tilde\sigma({\boldsymbol{x}}) = G\sigma(C{\boldsymbol{x}})$. We want to prove that \begin{align}YT & = \tilde\sigma(PYT W_p^s + X\tilde W_x) \\&=  \sigma(PYTW_p^sC + X\tilde W_xC)G \\ & = \sigma(PYTW_p^sC + XW_x)G & \text{ (Assume $\tilde W_xC = W_x$)}.\end{align}

A sufficient condition for this equation to hold is $T=G$ and $Y = \sigma(AYTW_p^sC + XW_x) = \sigma(AYW_p + XW_x)$. Comparing the last two terms, apparently it holds when $TW_p^sC = W_p$. And to ensure $\tilde \sigma \in \mathcal S$, from Lemma \ref{lem-tilde-sigma} we know it means $T^\top H(C^\top x)C^\top $ be positive semi-definite for all $x$.

\qed

We can verify this Theorem \ref{theo-fix-poi-cond} by linear case. If $\sigma$ is linear, then $H(Cx) = I$, thus simply taking $C = T^{-1}$ (on complex field) we have $TW_p^sT^{-1}$ can generate any real matrix $W$ and $T^\top H(C^\top x)C^\top  = T^\top T^{-\top} = I$ is positive semi-definite.

\end{document}